\documentclass[a4paper,10pt]{article}
 
\usepackage{graphicx}
\usepackage{epstopdf}
\usepackage{latexsym}
\usepackage{amsmath,amssymb}

\usepackage{algorithm}
\usepackage{algpseudocode}

\usepackage{color}

\usepackage[nocompress]{cite}

\usepackage{enumitem}
\usepackage{url}

\usepackage{booktabs}

\usepackage{multirow}

\usepackage{mdframed}
\usepackage{mathtools}



\long\def\invis#1{}

\title{Combining Graph Neural Networks and Mixed Integer Linear Programming for Molecular Inference
under the Two-Layered Model
}
\author{
Jianshen Zhu$^{1,5}$ \and
Naveed Ahmed Azam$^2$ \and
Kazuya Haraguchi$^1$ \and
Liang Zhao$^3$ \and
Tatsuya Akutsu$^4$
}
\date{
$^1$Graduate of Informatics, Kyoto University, Kyoto 606-8501, Japan \\
$^2$Department of Mathematics, Quaid-i-Azam University, Islamabad 45320, Pakistan \\
$^3$Graduate School of Advanced Integrated Studies in Human Survivability   (Shishu-Kan),  
 Kyoto University, Kyoto 606-8306, Japan  \\
 $^4$Bioinformatics Center,  Institute for Chemical Research, 
 Kyoto University, Uji 611-0011, Japan \\ 
 $^5$Department of Information Sciences, Tokyo University of Science, Noda, Chiba 278-8510, Japan \\
}

\newcommand{\filename}{2LMM\_GNN}
\newcommand{\bbC}{\mathbb{C}}
\newcommand{\bbR}{\mathbb{R}}

\newcommand{\calC}{\mathcal{C}}
\newcommand{\calG}{\mathcal{G}}
\newcommand{\calM}{\mathcal{M}}

\newcommand{\MILP}{\calM(g, x, y; \calC_1, \calC_2)}

\newcommand{\ylb}{\underline{y}^*}
\newcommand{\yub}{\overline{y}^*}

\newcommand{\sHomo}{\textsc{Homo}}
\newcommand{\sLumo}{\textsc{Lumo}}
\newcommand{\sGap}{\textsc{Gap}}
\newcommand{\smu}{\textsc{mu}}
\newcommand{\sAlpha}{\textsc{Alpha}}

\newcommand{\sCv}{\textsc{Cv}}
\newcommand{\sZPVE}{\textsc{ZPVE}}

\newcommand{\sRR}{\langle \textsc{R}^2 \rangle}

\newcommand{\GNN}{{\tt 2L-GNN}}
\newcommand{\molinfer}{{\tt mol-infer}}
\newcommand{\molinfergnn}{{\tt mol-infer-GNN}}

\newcommand{\zf}{{K_\mathcal{F}}}

\newcommand{\lf}{\mathrm{lf}}

\newcommand{\eledeg}{\mathrm{eledeg}}  
\newcommand{\eledegC}{\mathrm{eledeg}_\mathrm{C}}   
\newcommand{\eledegT}{\mathrm{eledeg}_\mathrm{T}}   
\newcommand{\eledegF}{\mathrm{eledeg}_\mathrm{F}}   
\newcommand{\eledegX}{\mathrm{eledeg}_\mathrm{X}}   

\newcommand{\vion}{\mathrm{v}_\mathrm{ion}}  

\newcommand{\ttH}{{\tt H}}  
\newcommand{\ttC}{{\tt C}}  
\newcommand{\ttO}{{\tt O}}  
\newcommand{\ttN}{{\tt N}}  
\newcommand{\ttP}{{\tt P}}  
\newcommand{\ttF}{{\tt F}}  
  
\newcommand{\ttS}{{\tt S}}

\newcommand{\oH}{\overline{{\tt H}}}  

\newcommand{\Z}{\mathbb{Z}}  
  
\newcommand{\C}{\mathbb{C}}  
\newcommand{\Co}{\mathbb{C}}  

\newcommand{\anC}{\langle \mathbb{C} \rangle}  
\newcommand{\anpsi}{\langle \psi \rangle}

\newcommand{\R}{\mathbb{R}}

\newcommand{\deghyd}{\deg^\mathrm{hyd}}

\newcommand{\thftr}{\theta_\bbC} 
\newcommand{\thC}{\theta^\mathrm{C}} 
\newcommand{\tauC}{\tau^\mathrm{C}} 
\newcommand{\deltauC}{\delta^\tau_\mathrm{C}}

\newcommand{\thT}{\theta^\mathrm{T}} 
\newcommand{\tauT}{\tau^\mathrm{T}} 
\newcommand{\deltauT}{\delta^\tau_\mathrm{T}} 
\newcommand{\thF}{\theta^\mathrm{F}} 
\newcommand{\tauF}{\tau^\mathrm{F}} 
\newcommand{\deltauF}{\delta^\tau_\mathrm{F}} 
\newcommand{\thX}{\theta^\mathrm{X}}

\newcommand{\thCTT}{\theta_\mathrm{T}^\mathrm{CT}} 
\newcommand{\thTCT}{\theta_\mathrm{T}^\mathrm{TC}} 
\newcommand{\thCTC}{\theta_\mathrm{C}^\mathrm{CT}} 
\newcommand{\thTCC}{\theta_\mathrm{C}^\mathrm{TC}} 
\newcommand{\thCFF}{\theta_\mathrm{F}^\mathrm{CF}} 
\newcommand{\thCFC}{\theta_\mathrm{C}^\mathrm{CF}} 
\newcommand{\thTFF}{\theta_\mathrm{F}^\mathrm{TF}} 
\newcommand{\thTFT}{\theta_\mathrm{T}^\mathrm{TF}}

\newcommand{\FrC}{\mathcal{F}^\mathrm{C}} 
\newcommand{\FrT}{\mathcal{F}^\mathrm{T}} 
\newcommand{\FrF}{\mathcal{F}^\mathrm{F}} 
\newcommand{\FrX}{\mathcal{F}^\mathrm{X}}

\newcommand{\Vleaf}{V_\mathrm{leaf}} 
\newcommand{\Eleaf}{E_\mathrm{leaf}} 

\newcommand{\sint}{\sigma_\mathrm{int}} 
\newcommand{\sce}{\sigma_\mathrm{ce}}

\newcommand{\Ez}{E_{(0/1)}}
\newcommand{\Ew}{E_{(\geq 1)}}
\newcommand{\Et}{E_{(\geq 2)}}
\newcommand{\Eew}{E_{(=1)}}

\newcommand{\Iz}{I_{(0/1)}}
\newcommand{\Iw}{I_{(\geq 1)}}
\newcommand{\It}{I_{(\geq 2)}}
\newcommand{\Iew}{I_{(=1)}}

\newcommand{\Gac}{\Gamma_\mathrm{ac}}

\newcommand{\typ}{\mathrm{t}}

\newcommand{\w}{w}

\newcommand{\ta}{{\tt a}}
\newcommand{\tb}{{\tt b}}
\newcommand{\Ldg}{\Lambda_{\mathrm{dg}}}

\newcommand{\fc}{\mathrm{fc}} 
\newcommand{\betar}{\beta_\mathrm{r}}

\newcommand{\val}{\mathrm{val}}

\newcommand{\inte}{\mathrm{int}}

\newcommand{\F}{\mathcal{F}}

\newcommand{\T}{\mathcal{T}}

\newcommand{\nint}{\mathrm{n}^\mathrm{int}}

\newcommand{\h}{\mathrm{ht}}

\newcommand{\ch}{\mathrm{ch}}

\newcommand{\dg}{\mathrm{dg}}

\newcommand{\na}{\mathrm{na}}
 
\newcommand{\naX}{\mathrm{na}_\mathrm{X}}

\newcommand{\naC}{\mathrm{na}_\mathrm{C}}
\newcommand{\naT}{\mathrm{na}_\mathrm{T}}
\newcommand{\naF}{\mathrm{na}_\mathrm{F}}

\newcommand{\acC}{\mathrm{ac}_\mathrm{C}}

\newcommand{\bdX}{\mathrm{bd}_\mathrm{X}}

\newcommand{\bdC}{\mathrm{bd}_\mathrm{C}}
\newcommand{\bdT}{\mathrm{bd}_\mathrm{T}}
\newcommand{\bdF}{\mathrm{bd}_\mathrm{F}}
\newcommand{\bdCT}{\mathrm{bd}_\mathrm{CT}}
\newcommand{\bdTC}{\mathrm{bd}_\mathrm{TC}}
\newcommand{\bdTF}{\mathrm{bd}_\mathrm{TF}}
\newcommand{\bdCF}{\mathrm{bd}_\mathrm{CF}}

\newcommand{\ns}{\mathrm{ns}}

\newcommand{\dgX}{\mathrm{dg}_\mathrm{X}}

\newcommand{\ec}{\mathrm{ec}}
\newcommand{\ac}{\mathrm{ac}}

\newcommand{\bl}{\mathrm{bl}}

\newcommand{\bd}{\mathrm{bd}}

\newcommand{\UB}{\mathrm{UB}}
\newcommand{\LB}{\mathrm{LB}}

\newcommand{\ex}{\mathrm{ex}}

\newcommand{\GC}{G_\mathrm{C}}

\newcommand{\mC}{m_\mathrm{C}}

\newcommand{\hC}{h^\mathrm{C}}
\newcommand{\hT}{h^\mathrm{T}} 
 
\newcommand{\hX}{h^\mathrm{X}}

\newcommand{\VF}{V_\mathrm{F}}
\newcommand{\VT}{V_\mathrm{T}}
\newcommand{\VC}{V_\mathrm{C}} 
 
\newcommand{\VX}{V_\mathrm{X}}

\newcommand{\ET}{E_\mathrm{T}}
\newcommand{\EC}{E_\mathrm{C}}

\newcommand{\EF}{E_\mathrm{F}}
\newcommand{\ECT}{E_\mathrm{CT}}
\newcommand{\ETC}{E_\mathrm{TC}}
\newcommand{\ETF}{E_\mathrm{TF}}

\newcommand{\ECF}{E_\mathrm{CF}}

\newcommand{\EX}{E_\mathrm{X}}

\newcommand{\vT}{{v^\mathrm{T}}}
\newcommand{\vC}{{v^\mathrm{C}}} 
  
\newcommand{\vX}{{v^\mathrm{X}}}

\newcommand{\vF}{{v^\mathrm{F}}}

\newcommand{\eF}{{e^\mathrm{F}}}
\newcommand{\eT}{{e^\mathrm{T}}}
\newcommand{\eC}{{e^\mathrm{C}}} 
 
\newcommand{\eX}{{e^\mathrm{X}}}

\newcommand{\eCF}{{e^\mathrm{CF}}}

\newcommand{\eCT}{{e^\mathrm{CT}}}
\newcommand{\eTC}{{e^\mathrm{TC}}}
\newcommand{\eTF}{{e^\mathrm{TF}}}

\newcommand{\tT}{{t_\mathrm{T}}}
\newcommand{\tC}{{t_\mathrm{C}}} 
\newcommand{\tF}{{t_\mathrm{F}}} 
 
\newcommand{\tX}{{t_\mathrm{X}}}

\newcommand{\IC}{{I_\mathrm{C}}}

\newcommand{\degCint}{{\deg_\mathrm{C}^\mathrm{int}}}
\newcommand{\degTint}{{\deg_\mathrm{T}^\mathrm{int}}}
\newcommand{\degFint}{{\deg_\mathrm{F}^\mathrm{int}}}
\newcommand{\degXint}{{\deg_\mathrm{X}^\mathrm{int}}}

\newcommand{\hyddeg}{\mathrm{hyddeg}}

\newcommand{\hyddegX}{\mathrm{hyddeg}^\mathrm{X}}

\newcommand{\degCex}{{\deg_\mathrm{C}^\mathrm{ex}}}
\newcommand{\degTex}{{\deg_\mathrm{T}^\mathrm{ex}}}
\newcommand{\degFex}{{\deg_\mathrm{F}^\mathrm{ex}}}
\newcommand{\degXex}{{\deg_\mathrm{X}^\mathrm{ex}}}
\newcommand{\degF}{{\deg^\mathrm{F}}}
\newcommand{\degT}{{\deg^\mathrm{T}}}
\newcommand{\degC}{{\deg^\mathrm{C}}} 
 
\newcommand{\degX}{{\deg^\mathrm{X}}}

\newcommand{\degCT}{\deg_\mathrm{CT}}
\newcommand{\degTC}{\deg_\mathrm{TC}}

\newcommand{\tldgC}{{\widetilde{\deg}_\mathrm{C}} }

\newcommand{\cF}{{c_\mathrm{F}}}

\newcommand{\kC}{{k_\mathrm{C}}}

\newcommand{\chiF}{{\chi^\mathrm{F}}} 
\newcommand{\dclrF}{\delta_{\chi}^\mathrm{F}}
\newcommand{\clrF}{\mathrm{clr}^{\mathrm{F}}}

\newcommand{\chiT}{{\chi^\mathrm{T}}}
\newcommand{\dclrT}{\delta_{\chi}^\mathrm{T}}
\newcommand{\clrT}{\mathrm{clr}^{\mathrm{T}}}

\newcommand{\tail}{\mathrm{tail}} 
\newcommand{\hd}{\mathrm{head}}

\newcommand{\dlfrF}{\delta_\mathrm{fr}^\mathrm{F}}

\newcommand{\dlfrC}{\delta_\mathrm{fr}^\mathrm{C}} 
 
\newcommand{\dlfrX}{\delta_\mathrm{fr}^\mathrm{X}}
 
\newcommand{\ddgF}{\delta_\mathrm{dg}^\mathrm{F}}
\newcommand{\ddgT}{\delta_\mathrm{dg}^\mathrm{T}}
\newcommand{\ddgC}{\delta_\mathrm{dg}^\mathrm{C}} 
 
\newcommand{\ddgX}{\delta_\mathrm{dg}^\mathrm{X}}

\newcommand{\ddgFint}{\delta_\mathrm{dg,F}^\mathrm{int}}
\newcommand{\ddgTint}{\delta_\mathrm{dg,T}^\mathrm{int}}
\newcommand{\ddgCint}{\delta_\mathrm{dg,C}^\mathrm{int}}  
\newcommand{\ddgXint}{\delta_\mathrm{dg,X}^\mathrm{int}}

\newcommand{\bF}{\beta^\mathrm{F}}
\newcommand{\bT}{\beta^\mathrm{T}}
\newcommand{\bC}{\beta^\mathrm{C}} 
\newcommand{\bX}{\beta^\mathrm{X}}
  
\newcommand{\bCT}{\beta^\mathrm{CT}}
\newcommand{\bTC}{\beta^\mathrm{TC}} 
\newcommand{\bTF}{\beta^\mathrm{TF}} 
\newcommand{\bCF}{\beta^\mathrm{CF}} 
\newcommand{\bXF}{\beta^\mathrm{XF}} 
\newcommand{\bsF}{\beta^{*\mathrm{F}}}

\newcommand{\bFex}{\beta^\mathrm{F}_\mathrm{ex}} 
\newcommand{\bTex}{\beta^\mathrm{T}_\mathrm{ex}} 
\newcommand{\bCex}{\beta^\mathrm{C}_\mathrm{ex}} 
\newcommand{\bXex}{\beta^\mathrm{X}_\mathrm{ex}}

\newcommand{\delbF}{\delta_{\beta}^\mathrm{F}}
\newcommand{\delbT}{\delta_{\beta}^\mathrm{T}}
\newcommand{\delbC}{\delta_{\beta}^\mathrm{C}}

\newcommand{\delbCT}{\delta_{\beta}^\mathrm{CT}}
\newcommand{\delbTC}{\delta_{\beta}^\mathrm{TC}}

\newcommand{\delbsF}{\delta_{\beta}^{*\mathrm{F}}} 
\newcommand{\delbX}{\delta_{\beta}^\mathrm{X}}

\newcommand{\aC}{{\alpha}^\mathrm{C}}  
 
\newcommand{\aX}{{\alpha}^\mathrm{X}}

\newcommand{\delaC}{\delta_\mathrm{\alpha}^{\mathrm{C}}}
\newcommand{\delaT}{\delta_\mathrm{\alpha}^{\mathrm{T}}}
\newcommand{\delaF}{\delta_\mathrm{\alpha}^{\mathrm{F}}}
\newcommand{\delaX}{\delta_\mathrm{\alpha}^{\mathrm{X}}}

\newcommand{\zmax}{K_{\textrm{node}}}
\newcommand{\llmax}{L}
\newcommand{\pcnv}{{K_{\bbC}}}
\newcommand{\phid}{K_{\textrm{hid}}}

\newcommand{\wll}{w_\ell}

\newcommand{\wftr}{w_{\bbC}}
\newcommand{\bias}{\textrm{bias}}
\newcommand{\Mll}{M_\ell}

\newcommand{\psifour}{\textsc{Psi4}}
\newcommand{\pyscf}{\textsc{PySCF}}

\begin{document} 
\maketitle

\begin{quote}  
{\bf Abstract}\\  
 Recently, a novel two-phase framework named \molinfer\ 
for inference of chemical compounds with prescribed abstract structures and desired property values
has been proposed.
The framework \molinfer\ is primarily based on using mixed integer linear programming (MILP)
to simulate the computational process of machine learning methods and describe 
the necessary and sufficient conditions to ensure such a chemical graph exists.
The existing approaches usually first convert the chemical compounds into handcrafted feature vectors
to construct prediction functions, 
but because of the limit on the kinds of descriptors 
originated from the need for tractability in the MILP formulation,
the learning performances on datasets of some properties are not good enough.
A lack of good learning performance can greatly lower the quality of the inferred chemical graphs,
and thus improving learning performance is of great importance.
On the other hand, graph neural networks (GNN) offer
a promising machine learning method to directly utilize the chemical graphs
as the input, and many existing GNN-based approaches 
to the molecular property prediction problem have shown that they
can enjoy better learning performances compared to
the traditional approaches that are based on feature vectors.
In this study, we develop a molecular inference framework based on \molinfer, 
namely \molinfergnn, that utilizes GNN as the learning method while keeping
the great flexibility originated from the two-layered model on the abstract structure of the chemical graph to be inferred.
We conducted computational experiments on the QM9 dataset to
show that our proposed GNN model can obtain satisfying learning performances for some properties
despite its simple structure,
and can infer small chemical graphs comprising up to 20 non-hydrogen atoms 
within reasonable computational time.



\end{quote}

\section{Introduction}\label{sec:introduction}

Designing novel molecules with predetermined structures and desired properties is 
a critical challenge across diverse research fields, 
including materials science~\cite{Morgan:2020aa} 
and drug discovery~\cite{Miyao:2016aa,Shi:2020aa}.
In recent years, significant progress has been made in molecular design using 
various machine learning techniques~\cite{Lo:2018aa, Tetko:2020aa}.
Computational molecular design, historically entrenched in chemoinformatics, 
has been studied under the name of \emph{quantitative structure activity/property relationship} 
(QSAR/QSPR)~\cite{Cherkasov:2014aa, Skvortsova:1993aa} 
and its inverse counterpart \emph{inverse quantitative structure
activity/property relationship} (inverse QSAR/QSPR)~\cite{Ikebata:2017aa, Miyao:2016aa, Rupakheti:2015aa}.
Analysis of the activities and properties of chemical compounds is crucial not only in chemistry
 but also in biology,  given their pivotal roles in various metabolic pathways.

 QSAR/QSPR aims to predict chemical activities from given chemical 
structures~\cite{Cherkasov:2014aa}. A prediction function is usually constructed from existing 
structure-activity relation data, employing 
machine learning-based
methods, including artificial neural network (ANN)-based methods~\cite{Lo:2018aa, Tetko:2020aa}.
On the other hand, inverse QSAR/QSPR seeks to infer chemical structures from given 
chemical activities~\cite{Ikebata:2017aa, Miyao:2016aa, Rupakheti:2015aa}.
In most classical approaches,
chemical structures are typically treated as an undirected graph called
\emph{chemical graphs} and encoded as a vector of real numbers called \emph{descriptors} or
\emph{feature vectors}.
A typical approach to inverse QSAR/QSPR involves first inferring feature vectors from
 given chemical activities and then reconstructing chemical graphs from these 
 feature vectors~\cite{Ikebata:2017aa, Miyao:2016aa, Rupakheti:2015aa}. 
 While these handcrafted features have been effective in many cases,
 the limitations become apparent when dealing with large-scale datasets~\cite{Gilmer:2017aa}.
 
 Recent advancements in deep learning, especially the development of \emph{graph neural networks} (GNNs),
 have provided a promising alternative to traditional feature-based methods.
 GNNs have demonstrated superior performance in capturing intricate relationships with 
 chemical graphs by directly utilizing the graph structure as input, eliminating the need for
 manually designed descriptors~\cite{Kipf:2017aa, Schutt:2017aa,Gilmer:2017aa}.
 The ability to automatically learn meaningful representations from non-Euclidean data
 has opened new avenues for molecular property prediction,
 with several GNN-based models achieving state-of-the-art results on benchmark datasets like 
 QM9~\cite{Duvenaud:2015aa,Gilmer:2017aa,Schutt:2017aa,Gasteiger:2022aa,Zhang:2023ab}.
 Also, there are several GNN-based approaches to the molecular graph generation problem~\cite{Wu:2021aa}.
These methods either generate a new graph 
in a sequential manner~\cite{Gomez-Bombarelli:2018aa, Kusner:2017aa}--adding vertices and edges step by step, 
or in a global manner~\cite{De-Cao:2022aa, Bojchevski:2018aa}--outputting an entire graph at once.
 
 However, despite the remarkable progress made with GNNs (and other deep learning-based approaches),
 most existing approaches focus mainly on the prediction task of QSAR/QSPR, and few studies have
 effectively integrated GNNs into the inverse QSAR/QSPR frameworks.
 A significant challenge is to ensure the following two important properties of the generated chemical structures,
 namely  \emph{optimality}---the quality of the solution for the inverse problem of the learning methods,
 and \emph{exactness}---whether the solution admits a valid chemical graph.
 While many deep learning-based generative models (e.g., \cite{Shino:2025aa, Cai:2022aa, Bort:2022aa, Kaneko:2023aa}) aim to create chemically plausible molecules,
 they often fail to guarantee the optimality or the exactness of the inferred solutions mathematically,
 which can be problematic in practical applications~\cite{Zhu:2023aa}.
 

\begin{figure}[t!]
\begin{center}
\includegraphics[width=.89\columnwidth]{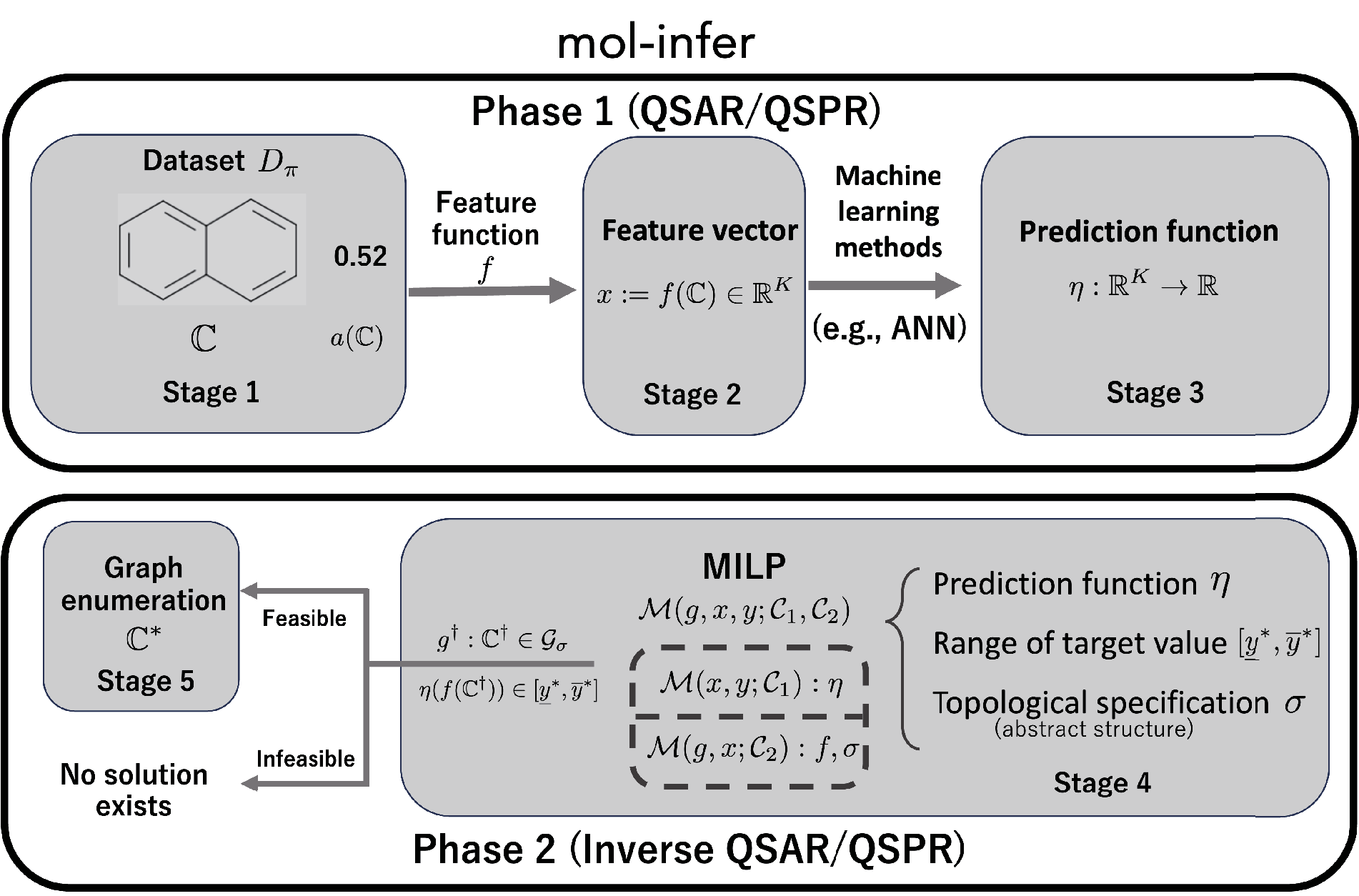}
\end{center}
\caption{An illustration of the two-phase framework \molinfer.}
\label{fig:framework}
\end{figure}

To overcome these limitations,
a novel framework called \molinfer~\cite{Azam:2020aa, Zhang:2022aa, Shi:2021aa, Zhu:2022ad}
was recently proposed 
for inferring chemical compounds with prescribed abstract structures and desired property values. 
This framework is primarily based on using the \emph{mixed integer linear programming} (MILP) formulation
to simulate the computational process of machine learning methods and also describe 
the necessary and sufficient conditions to ensure the existence of a valid chemical graph.
As a result, \molinfer\ generates chemical graphs in a global manner and 
guarantees both the optimality and exactness of the inferred chemical graph.
Figure~\ref{fig:framework} provides an illustration of \molinfer.
Simply put, \molinfer\ consists of two phases. 
Phase~1 is the QSAR/QSPR phase aiming to construct a prediction function $\eta$ between
chemical compounds and their observed property values.
Let $\calG$ denote the set of all possible chemical graphs.
First, we collect a dataset $D_\pi \subseteq \calG$ of chemical graphs consisting of
chemical graphs $\bbC$ and the observed values $a(\bbC)$ (Stage~1).
Then we use a feature function $f: \calG \to \bbR^K$ ($K$ is a positive integer) to convert chemical
graphs to a $K$-dimensional real vectors (Stage~2). 
Finally, a prediction function $\eta: \bbR^K \to \bbR$ is constructed by some machine learning methods
(Stage~3).
Phase~2 is the inverse QSAR/QSPR phase, and the target is to
infer chemical graphs with specific property values.
Given a set of rules called topological specification $\sigma$ that specifies
the desired structure of the inferred chemical graphs, and a desired range $[\ylb, \yub]$ of the target value, 
Stage~4 is designed to infer chemical graphs $\bbC^*$ that satisfy the rules $\sigma$ and
$\eta(f(\bbC^*)) \in [\ylb, \yub]$  by 
solving an MILP formulation $\MILP$ that represents:
\begin{itemize}
\item[(i)] $\mathcal{M}(x,y;\mathcal{C}_1)$: the computation process of the prediction function $\eta$; and
\item[(ii)] $\mathcal{M}(g,x;\mathcal{C}_2)$: that of the feature function $f$ and the constraints for $\bbC \in \calG_\sigma$,
\end{itemize}
where $\calG_\sigma$ denotes the set of all chemical graphs satisfying $\sigma$.
In Stage~5, dynamic programming-based 
graph enumeration algorithms~\cite{Ido:2025aa} are used to generate isomers of the inferred
chemical graphs $\bbC^*$ obtained in Stage~4.
This framework was originally proposed for only limited classes of chemical graphs;
e.g., trees~\cite{Zhang:2022aa, Azam:2021aa}, rank-1 graphs~\cite{Ito:2021aa}, 
and rank-2 graphs~\cite{Zhu:2020aa}.
The \emph{two-layered model} (2L-model) proposed by Shi~et~al.~\cite{Shi:2021aa} 
admits us to infer any chemical graph,
where users need to design an abstract structure as a part of the input,
and is now the standard model in \molinfer.
One of the superiority of \molinfer\ is that
it can suggest that
$\calG_\sigma$ does not contain  such a desired chemical graph 
when the MILP formulation $\MILP$ is infeasible,
while most existing inverse QSAR/QSPR models fail to do this. 
Several machine learning methods have been employed into \molinfer, for example,
ANNs~\cite{Shi:2021aa}, linear regression~\cite{Zhu:2022ad, Zhu:2025aa}, and 
decision trees~\cite{Tanaka:2021aa}. 
(We refer to the thesis~\cite{Zhu:2023aa} for a more comprehensive description of \molinfer.)

All of these previous approaches rely heavily on handcrafted feature vectors
because of the limit on the kinds of descriptors 
originating from the need for tractability in the MILP formulation $\mathcal{M}(g,x;\mathcal{C}_2)$.
However, 
the learning performances on datasets of some properties are limited.
For example, the median test $R^2$ score for the dataset of electric dipole moment ($\smu$) 
consisting of randomly selected 1000 molecules from the QM9 dataset 
is less than 0.7~\cite{Zhu:2025aa, Zhu:2024aa}, 
and for the dataset of autoignition temeprature annotated from 
Hazardous Substances Data Bank (HSDB)~\cite{HSDB} on Pubchem~\cite{Kim:2023aa} 
it is around 0.8~\cite{Zhu:2024aa}.
A lack of good learning performance can greatly lower the quality of chemical graphs inferred in Phase 2,
and thus developing new ways to improve learning performance is of great importance.
One noteworthy issue is that,
more complex methods can increase the accuracy of the resultant prediction functions in Phase~1.
On the other hand, 
it is generally hard to represent the computational process of such a method by MILP (e.g., kernel methods like support vector machine),
and even in the case where it is possible,
the time needed to solve the MILP formulations in Phase~2 will increase drastically as well,
and thus, it is challenging to incorporate complicated learning methods into \molinfer.

In this paper, we introduce \molinfergnn, an advanced extension of the \molinfer\ framework 
that integrates GNNs as the primary learning method while retaining the rigorous MILP formulations
for structure inference.
By directly utilizing the chemical graph as input,
the newly proposed framework overcomes the limitations of handcrafted features and enhances
the predictive power of the model.
The key contributions of this work are summarized as follows:
\begin{itemize}
\item[-] We develop a relatively simple GNN architecture, \GNN,
specifically designed for the two-layered model, 
ensuring efficient and effective learning from chemical graphs.
\item[-] We incorporate the \GNN\ model into the \molinfer\ framework,
and manage to formulate the inverse problem using MILP,
thereby guaranteeing both the optimality and exactness
of the inferred chemical structures.
While previous studies (e.g., ~\cite{McDonald:2024aa, Zhang:2024aa})
have used MILP formulations to simulate
the computation process of a given GNN,
these approaches have been restricted in the variety of chemical graphs they can generate.
In contrast, our method benefits from the flexibility of the two-layered model,
allowing for a significantly broader range of chemical structures to be inferred.
\item[-] We conduct numerical experiments 
on the QM9 dataset~\cite{Wu:2018aa, Ramakrishnan:2014aa}
to evaluate the predictive accuracy and inference efficiency of our proposed approach. 
Specifically, we demonstrate that the inverse problem can be solved efficiently enough to
infer a chemical graph comprising up to 20 non-hydrogen atoms.
Additionally, we use the open-source quantum chemistry software 
\psifour~\cite{Smith:2020aa} and \pyscf~\cite{Sun:2018aa, Sun:2020aa}
to compute the property values of the inferred molecules and compare them with the ones
obtained from MILP solutions,
and the experimental results demonstrate that the generated compounds are generally of good quality.
\end{itemize}


The paper is organized as follows.
We introduce some basic notations on graphs, the modeling of chemical compounds,
and the two-layered model in Section~\ref{sec:preliminary}.
Section~\ref{sec:GNN} introduces our newly-proposed GNN-based inverse QSAR/QSPR framework \molinfergnn.
We report some results on computational experiments conducted on the QM9 dataset in Section~\ref{sec:experiment}.
Section~\ref{sec:conclude} concludes the paper.
More details are available in the Appendix, including a full list of constraints in MILP formulations.
All program codes and experimental results are available at {\url{https://github.com/ku-dml/mol-infer/tree/master/2LGNN}}.


\section{Preliminary}\label{sec:preliminary}
In this section, we introduce some essential concepts and notations related to graph theory 
and the modeling of chemical compounds, which are the foundation for our proposed framework.
These definitions follow mainly the work of Zhu~et~al.~\cite{Zhu:2022ad},
with some necessary modifications tailored to our approach.

For two integers $a$ and $b$ such that $a \leq b$, let $[a,b]$ denote the set of 
integers $i$ with $a\leq i\leq b$.

\subsection{Graphs}

Throughout this study, we consider a \emph{graph} as
a simple connected undirected graph.
The sets of \emph{vertices} and \emph{edges} of $G$
are denoted by $V(G)$ and $E(G)$, respectively.
For any vertex $v\in V(G)$, the \emph{neighborhood} of $v$ is denoted by $N_G(v)$,
and the {\em degree} $\mathrm{deg}_G(v)$ of $v$ is defined to be
$\mathrm{deg}_G(v)=|N_G(v)|$.
 
We sometimes designate a vertex in a graph $G$ as a {\em root},
and call such a graph {\em rooted}. 
A {\em leaf-vertex} in a graph $G$ (possibly with a root) is defined to be 
be a non-root vertex  $v$ with degree 1.
An edge $uv$ incident to a leaf-vertex $v$ is called a {\em leaf-edge},
 and the sets of leaf-vertices and leaf-edges are denoted by $\Vleaf(G)$ and $\Eleaf(G)$, respectively.
 We define a sequence of graphs $G_i, i\in \mathbb{Z}_+$, for a graph $G$
 by removing the set of leaf-vertices iteratively as follows:
 \[ G_0:=G; ~~ G_{i+1}:=G_i - \Vleaf(G_i). \]
We call a vertex $v$ a {\em tree vertex} if $v\in \Vleaf(G_i)$ for some integer $i \geq 0$,
and define the  {\em height} $\h(v)$ of $v$ to be $i$.
For each non-tree vertex $v$ adjacent to a tree vertex,
we define $\h(v)$
to be $\h(u)+1$, where $u$ is the one with the maximum $\h(u)$ among the neighbors of $v$.
The height is left undefined for any non-tree vertex that is not adjacent to any tree vertex. 
The {\em height} $\h(T)$ of a rooted tree $T$ is defined
to be the maximum of $\h(v)$ of a vertex $v\in V(T)$.  

\subsection{Modeling of Chemical Compounds}\label{sec:chemical_model}

  
To represent a chemical compound, 
we employ the \emph{chemical graph} 
which abstracts a molecule as a graph where vertices represent atoms
and edges represent bonds. 
We refer~\cite{Zhu:2022ad} for a more detailed description of chemical graphs.
%
 
 A chemical compound $\bbC$ is represented by a {\em chemical graph} which is defined to be
a triplet $\bbC=(H,\alpha,\beta)$  of a graph $H$, $\alpha:V(H)\to \Lambda$  
assigns chemical elements to vertices,
and  $\beta: E(H)\to [1,3]$ assigns the bond-multiplicity to edges.
Here $\Lambda$ is a set of chemical elements,
and we denote a chemical element $\ta$ with a valence $i$ by $\ta_{(i)}$.
Such a suffix $(i)$ is omitted for a chemical element $\ta$ with a unique valence. 
The {\em hydrogen-suppressed chemical graph} $\anC$ of $\bbC$
is the graph obtained by removing all hydrogen atoms.

Two chemical graphs $(H_1,\alpha_1,\beta_1)$ and $(H_2, \alpha_2, \beta_2)$ are
 {\em isomorphic} if a bijection $\phi: V(H_1)\to V(H_2)$ exists
such that chemical and structural information are preserved under $\phi$;
i.e., 
 $uv\in E(H_1), \alpha_1(u)=\ta, \alpha_1(v)=\tb, \beta_1(uv)=m$
 if and only if
$\phi(u)\phi(v) \in E(H_2), \alpha_2(\phi(u))=\ta, 
\alpha_2(\phi(v))=\tb, \beta_2(\phi(u)\phi(v))=m$. 

%

\subsection{Two-layered Model}\label{sec:2LM}

This subsection reviews the {\em two-layered model} (2L-model) 
proposed  by Shi~et~al.\cite{Shi:2021aa} and further refined by Zhu~et~al.~\cite{Zhu:2022ad},
which divides the hydrogen-suppressed chemical graph $\anC$ into two parts: the {\em interior} and
the {\em exterior}.

\begin{figure}[t!]
\begin{center}
\includegraphics[width=.80\columnwidth]{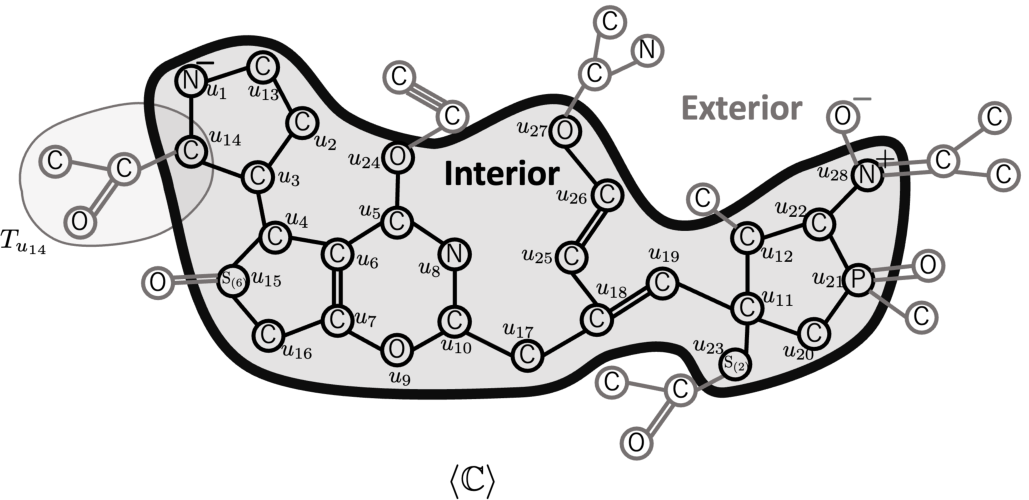}
\end{center}
\caption{An illustration of the 2L-model for a chemical graph $\bbC$. 
Here $\anC$ is the hydrogen-suppressed chemical graphs of $\bbC$.
The interior is represented by the shaded area enclosed by thick black lines, 
while the remaining parts form the exterior. 
Vertices $u_i, i\in[1,28]$ are the interior-vertices, 
and $T_{u_{14}}$ is the chemical tree rooted at vertex $u_{14}$, outlined by a thin gray line.}
\label{fig:example_chemical_graph}
\end{figure}

Let $\rho \geq 1$ be an integer, which we call a {\em branch-parameter} and
$\rho=2$ is the standard value.
For a chemical graph $\bbC=(H,\alpha,\beta)$, 
we categorize each vertex $v \in V(\anC)$ in the hydrogen-suppressed chemical graph 
as follows;   {\em exterior-vertex} if $\h(v) < \rho$ for the height defined on $\anC$, and {\em interior-vertex} otherwise.
An edge $e \in E(\anC)$ is called an {\em exterior-edge} if $e$ is incident to an exterior-vertex,
and {\em interior-edge} otherwise.
We denote the sets of exterior-vertices, exterior-edges, interior-vertices and interior-edges of $\bbC$
by $V^\ex(\bbC)$, $E^\ex(\bbC)$, $V^\inte(\bbC)$ and $E^\inte(\bbC)$, respectively.
The {\em interior} of $\bbC$ is defined to be the subgraph
 $\bbC^\inte := (V^\inte(\bbC),E^\inte(\bbC))$ of $\anC$. 
Notice that the set $E^\ex(\bbC)$ of exterior-edges consists of
a collection of connected graphs, each of which can be viewed 
as a rooted tree $T$ rooted at an interior-vertex $v\in V(T)$. 
We denote the set of these chemical rooted trees in $\anC$ as $\mathcal{T}^\ex(\anC)$.
 For each interior-vertex $u\in V^\inte(\bbC)$,
let $T_u\in \mathcal{T}^\ex(\anC)$ denote the chemical tree rooted at $u$
(where $T_u$ may consist of only one vertex $u$).
 See Figure~\ref{fig:example_chemical_graph} for an illustration of these concepts.
 
\begin{figure}[t!] 
\begin{center}
\includegraphics[width=.90\columnwidth]{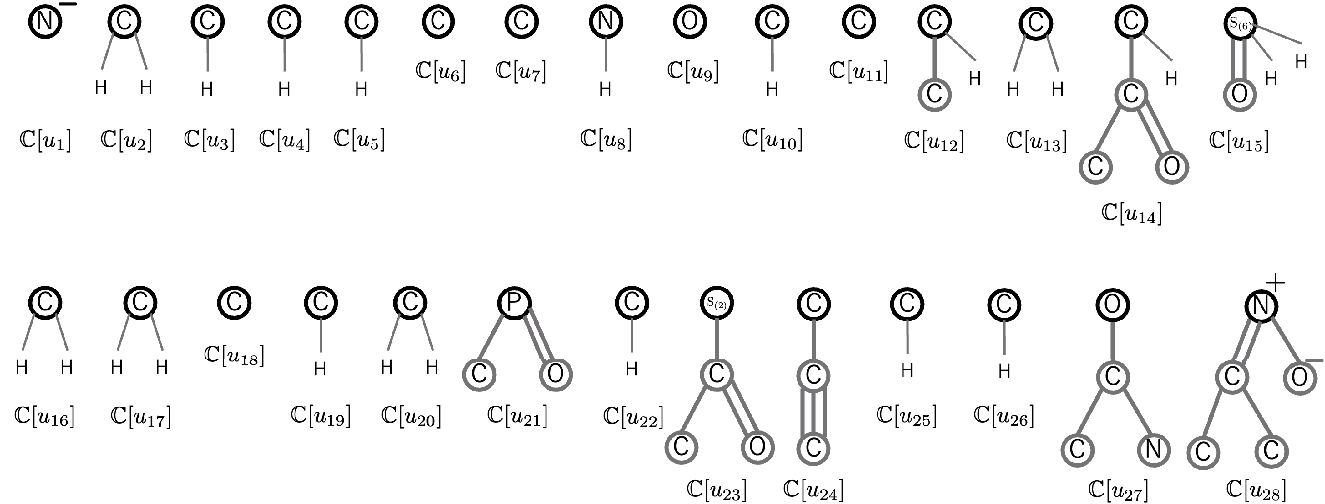}
\end{center} 
\caption{
An illustration of 2-fringe-trees $\bbC[u_i], u\in[1,28]$ of the example $\bbC$
depicted in Figure~\ref{fig:example_chemical_graph}.
We depict the root of each 2-fringe-tree with a black circle
and omit the hydrogens attached to non-root vertices.
 }
\label{fig:example_fringe-tree} 
\end{figure}

 The {\em $\rho$-fringe-tree} $\bbC[u]$ is defined to be
the chemical rooted tree obtained from $T_u$ by putting back
 the hydrogens originally attached with $T_u$ in $\bbC$. 
 We denote the set of $\rho$-fringe-trees in $\bbC$ as $\mathcal{T}(\bbC)$.
Figure~\ref{fig:example_fringe-tree}  illustrates
the set  $\mathcal{T}(\bbC)=\{\bbC[u_i]\mid i\in [1,28]\}$ of the 2-fringe-trees 
  of the example $\bbC$ with $\anC$ in Figure~\ref{fig:example_chemical_graph}. 
%


In order to describe the abstract structure of the chemical graph to be inferred,
a set of rules called  \emph{topological specification}  was introduced by
  Tanaka et~al.~\cite{Tanaka:2021aa}
  and later updated by Zhu~et~al.~\cite{Zhu:2022ad}.
It consists of the following three parts:
\begin{enumerate}[nosep]
\item[(i)]
A {\em seed graph} $\GC$ that provides an  abstract structure of  a target chemical graph $\bbC$;
\item[(ii)]
 A set $\mathcal{F}$ of chemical rooted trees  as candidates
 for the $\rho$-fringe-tree  $\bbC[u]$ rooted at each interior-vertex $u$ in $\bbC$; 
and 
\item[(iii)]
Lower and upper bounds on the number of components 
 in a target chemical graph such as  chemical elements, 
double/triple bonds and the interior-vertices in $\bbC$. 
\end{enumerate}

We refer Appendix~\ref{sec:specification} and~\cite{Zhu:2022ad}
for a detailed description of topological specification and how the MILP in Stage~4 expands the seed graph to get a complete chemical graph.

%
%


\section{\molinfergnn: A GNN-based Inverse QSAR/QSPR Framework}\label{sec:GNN}

In this section, we describe our proposed molecular inference framework, \molinfergnn, which uses GNN as the learning method so that
chemical graphs are directly used as the input. 
As a variant of \molinfer,
\molinfergnn\ inherits the most important feature 
that the optimality and exactness of the obtained solution are guaranteed by solving MILP formulations.

\begin{figure}[t!]
\begin{center} 
 \includegraphics[width=.95\columnwidth]{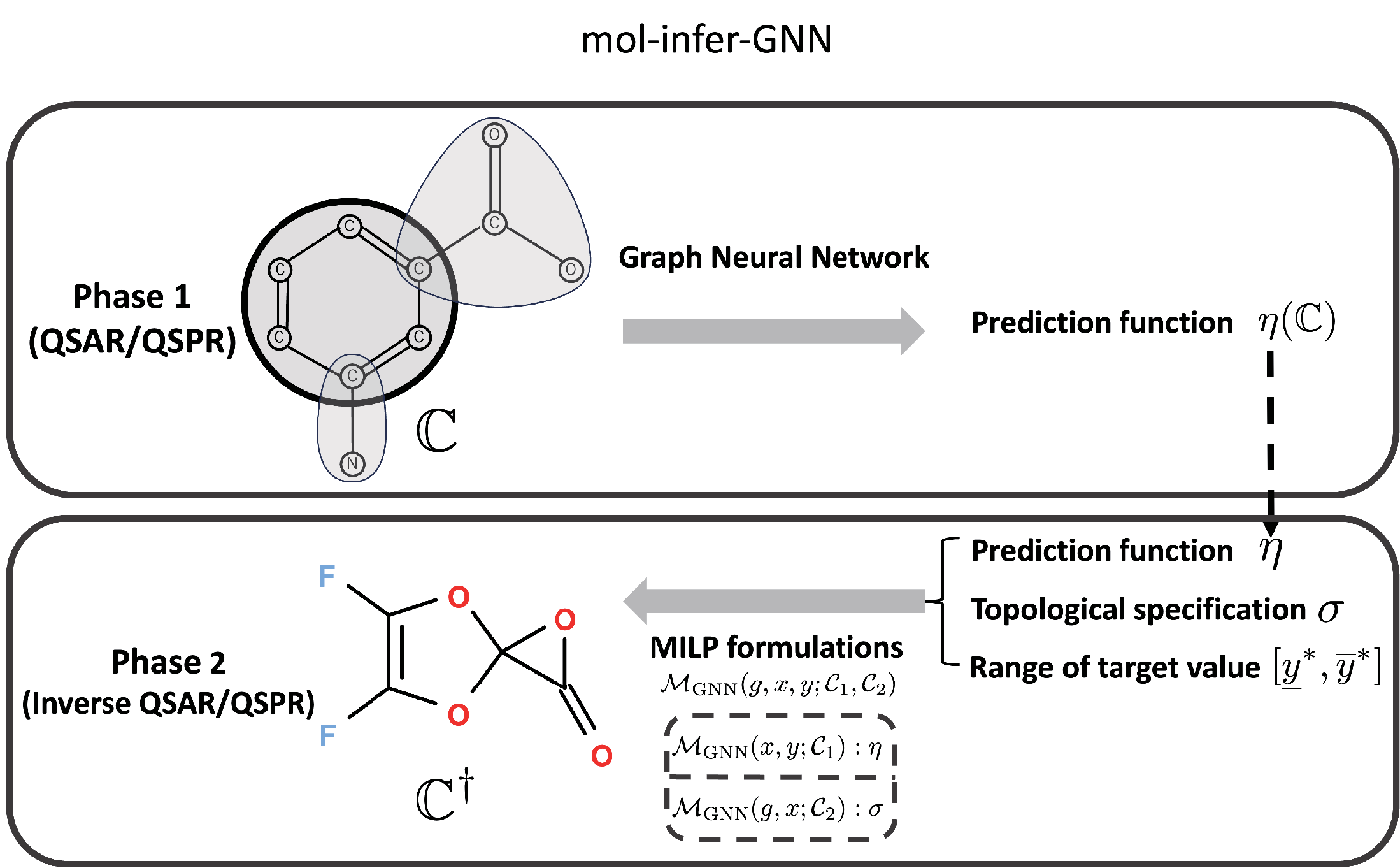}
\end{center}
\caption{An illustration of the two-phase GNN-based molecular inference framework \molinfergnn.
}
\label{fig:figure_GNN_framework}  
\end{figure}

The framework consists of two phases, and the basic idea of each phase remains the same as \molinfer.
We illustrate the framework in Figure~\ref{fig:figure_GNN_framework}.

\subsection{Phase~1: QSAR/QSPR Phase}\label{sec:GNN_phase1}
Given a property $\pi$ and a dataset consisting of chemical graphs $\bbC$ and their observed values $a(\bbC) \in \mathbb{R}$,
the target of Phase~1 is to construct a prediction function $\eta: \mathcal{G} \to \mathbb{R}$ between
a chemical graph $\bbC=(H,\alpha,\beta)$ and its observed property value $a(\bbC)$.

\paragraph{Graph Neural Networks}
%
Graph neural networks (GNNs)~\cite{Gilmer:2017aa, Kipf:2017aa, Micheli:2009aa} are a specialized class of neural network models
designed to process graph-structured data.
Unlike traditional neural networks, which require fixed-sized vectors as inputs,
GNNs can handle the irregular, non-Euclidean nature of graphs, making them
ideal for tasks involving relational data, such as molecular graphs, social networks, and so on~\cite{Wu:2021aa}.
We will use the term \emph{node} interchangeably with vertex in the context of graph neural networks.

A fundamental concept behind GNNs is {\em message passing},
a mechanism that enables each node in a graph to iteratively update its feature representation 
by aggregating information from its neighbors. 
This iterative exchange allows GNNs to capture both local structures 
(i.e., relationships between adjacent nodes) 
and global structures 
(i.e., long-range dependencies within the graph). 
This gives GNNs the capacity to learn highly expressive representations 
that encode both topological and attribute-based information, 
making them powerful tools for graph-based learning tasks.
One of the key advantages of GNNs over traditional machine learning methods is the ability to 
generalize across varying graph sizes and structures, eliminating the need for handcrafted feature engineering,
which greatly limits the learning performance on some datasets.

Formally, for a GNN with $L$ layers, given a graph $G=(V,E)$ 
with initial node feature vectors $\theta_v^{(0)}$ for each node $v\in V$,
the feature update process at the $\ell$-th ($\ell \in [1,L]$) layer can be described as:
\[
a^{(\ell)}_v \gets \textrm{AGGREGATE}^{(\ell)} ( \{ \theta^{(\ell-1)}_u \mid u \in N_G(v) \}   ),
\]
\[ 
\theta^{(\ell)}_v \gets \textrm{COMBINE}^{(\ell)}(\theta^{(\ell-1)}_v, a^{(\ell)}_v),
\]
where $\textrm{AGGREGATE}^{(\ell)}(\cdot)$ is a learnable function that collects features from neighboring nodes, and
$\textrm{COMBINE}^{(\ell)}(\cdot)$ is a learnable function that merges the aggregated features with the current node's features.

After $L$ layers of message passing, the node features $\theta_v^{(L)}$ will capture the structural information
from the $L$-hop neighborhood of each node~\cite{Xu:2019aa}.
A graph-level representation vector $\theta_G$ can then be derived by a readout function, typically
an aggregation like mean pooling, max pooling, or some more complex function 
based on attention mechanisms:
\[
\theta_G \gets \textrm{READOUT}(\{ \theta_v^{(L)} \mid v \in V \}).
\]

GNNs have shown remarkable success in various applications, especially in molecular property prediction,
where they outperform traditional feature-based models by directly leveraging the graph structure of molecules.

%
%

\paragraph{2L-GNN}
While GNNs have demonstrated their superior performance in QSAR/QSPR tasks,
integrating GNNs into inverse QSAR/QSPR frameworks faces significant challenges.
Typical GNN-based methods often focus on predicting properties from molecular graphs,
and they do not guarantee the optimality or exactness of the inferred structures 
when applied to the inverse problem.
Additionally, many models rely on 3D structural information (e.g., bond angles and interatomic distances),
which can be computationally expensive to obtain and process~\cite{Gilmer:2017aa}.

\begin{figure}[t!]
\begin{center} 
 \includegraphics[width=.98\columnwidth]{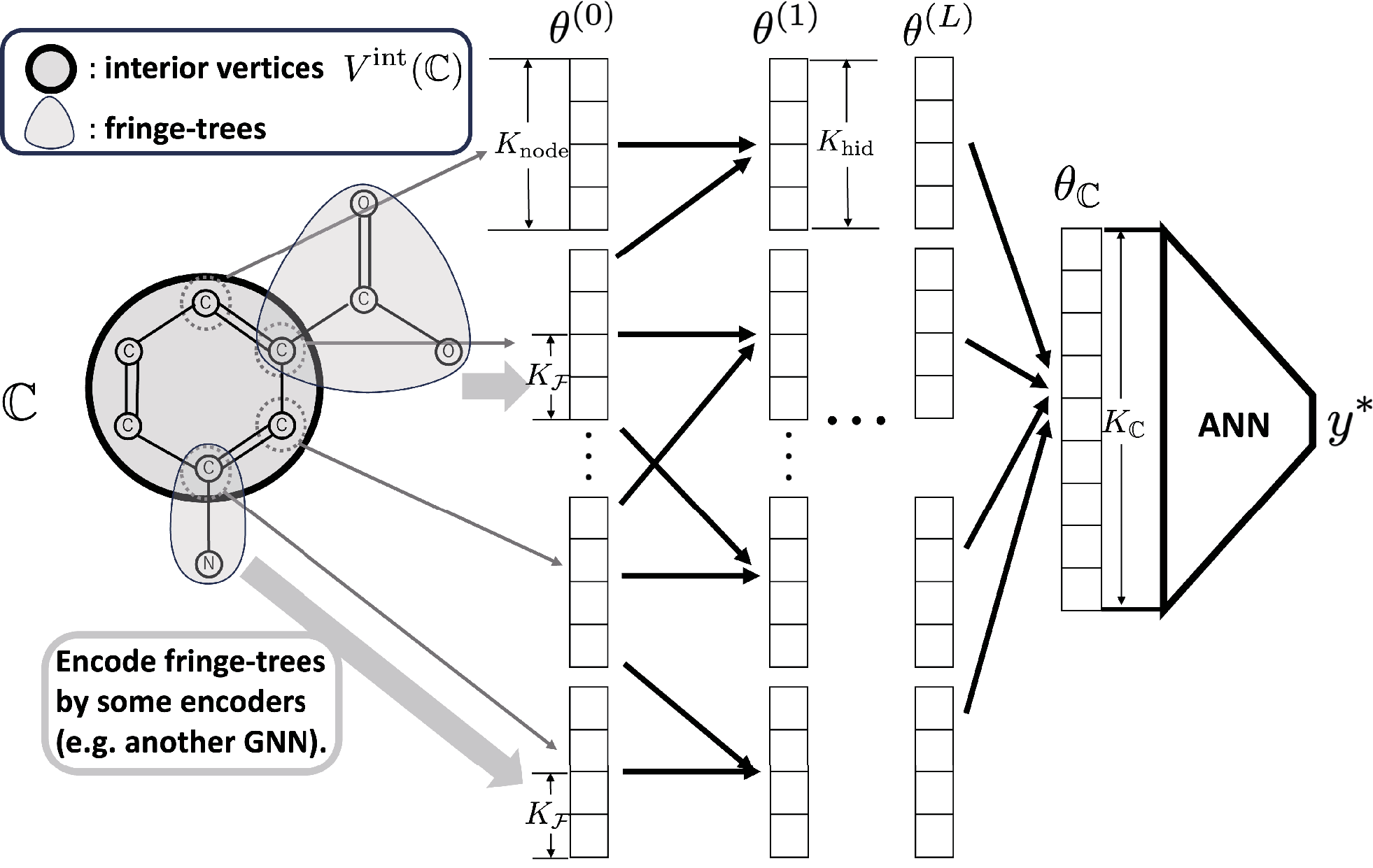}
\end{center}
\caption{An illustration of the graph neural network \GNN\ that is used in \molinfergnn\ to construct a prediction function $\eta$.
}
\label{fig:figure_GNN}  
\end{figure}

To overcome these limitations, here we propose a novel GNN architecture, \GNN,
which is designed specifically to enhance learning performance and maintain 
the flexibility of the 2L-model of chemical graphs introduced in
Section~\ref{sec:2LM}.
Unlike most of the existing approaches that process all nodes (including the hydrogen atoms),
\GNN\ uses only on the interior-vertices $V^\inte(\bbC)$ as the nodes in GNN architecture.
This selective processing reduces computational overhead while preserving essential information.
%
%
An illustration of \GNN\ is shown in Figure~\ref{fig:figure_GNN}.

Let $\llmax, \zmax, \zf, \phid,  \pcnv \in \Z_+$ be five positive integers,
where 
$\llmax$ denotes the number of layers in \GNN,
$\zmax$ denotes the length of an initial node feature vector,
$\zf$ denotes the length of the encoded feature vector that represents the local structure,
$\phid$ denotes the length of a node feature vector at the other layers (hidden layers),
and $\pcnv$ denotes the length of the representation vector of a chemical graph $\bbC$.
The initial feature vector $\theta_v^{(0)}\in \bbR^{\zmax}$ for each interior vertex $v \in V^\inte(\bbC)$
consists of the following components:
\begin{itemize}
\item[-] Atom type: One-hot encoding of the atom type of $\alpha(v)$ ($\ttC, \ttO, \ttN$ or not);
\item[-] Degree: The degree $\deg_\bbC(v)$ of the vertex in the chemical graph;
\item[-] Valence: The valence of the atom $\alpha(v)$;
\item[-] Hydrogen count: The number of hydrogens attached to the vertex $v$;
\item[-] Ion-valence: The ion-valence of the atom;
\item[-] Fringe-tree encoding: A feature vector $\theta_\psi \in \bbR^{\zf}$
representing the local structure ($\rho$-fringe-tree $\mathcal{T}_v \in \mathcal{T}(\bbC)$) rooted at $v$, obtained via a secondary GNN model.
\end{itemize}
Notice that we include only 2D information about a molecule
since the spatial information like bond angles or interatomic distances is hard to tract in our MILP formulations of Phase~2.


Because of the need to simulate the computational process by MILP formulations in Phase~2,
our choice of the functions $\textrm{AGGREGATE}^{(\ell)}(\cdot)$ and $\textrm{COMBINE}^{(\ell)}(\cdot)$
are rather simple.
The features of each interior vertex are updated through $\llmax$ layers using a straightforward yet
effective message-passing scheme.
At each layer $\ell, \ell \in [1, \llmax]$, the update rule 
for the node feature vector $\theta^{\ell}_v \in \bbR^{\phid}$ is given by:
\[
\theta^{(\ell)}_v \gets \textrm{LReLU}(\sum_{u \in N_{\bbC^\inte}(v) \cup \{ v \} } W^{{\ell}}  \theta_v^{(\ell-1)} + B^{(\ell)}), 
\]
where $W^{(\ell)} \in \bbR^{\phid \times \phid}$ ($W^{(1)} \in \bbR^{\zmax \times \phid}$) and $B^{(\ell)} \in \bbR^{\phid}$
are learnable weight matrices and biases and the LReLU (Leaky ReLU) activation $\textrm{LReLU}(x) := \textrm{max}(\alpha x, x)$ is used with a slope parameter of $\alpha=0.1$ to
avoid vanishing gradients.

The graph-level representation vector $\theta_\bbC \in \bbR^\pcnv$ is computed by aggregating the final 
layer features of all interior vertices:
\[
\theta_\bbC \gets \textrm{LReLU}(\sum_{v \in V^\inte(\bbC)} \theta_v^{(\llmax)}).
\]

The representation vector  $\theta_\bbC$ is then fed into a fully connected artificial neural network
with ReLU activation function $\textrm{ReLU}(x):=\textrm{max}(0,x)$ (except the final layer)
to output the predicted property value $y^* := \eta(\bbC)$ of the chemical graph $\bbC$.

\subsection{Phase~2: Inverse QSAR/QSPR Phase}\label{sec:GNN_phase2}

Phase~2 is designed to address the core task of inferring a chemical graph
that satisfies property and structural requirements.
The challenge of ensuring both the optimality and exactness of the inferred chemical graph
is achieved by formulating the problem as an MILP problem,
which we denote as $\mathcal{M}_{\mathrm{GNN}}(g,x,y;\mathcal{C}_1, \mathcal{C}_2)$.

Formally, the primary goal of Phase~2 is to infer a chemical graph $\bbC^\dagger$ given:
\begin{itemize}
\item[-] the prediction function $\eta$ learned in Phase~1;
\item[-] a topological specification $\sigma$, which defines the desired abstract structure of the chemical graph
to be inferred; and
\item[-] a target property range $[\ylb, \yub]$, indicating the desired range of the predicted property value.
\end{itemize}
The objective is to find a chemical graph $\bbC^\dagger$ that belongs to the set of all chemical graphs
satisfying the specification $\sigma$ and whose predicted property value lies within the specified range, i.e.:
\[
\bbC^\dagger \in \mathcal{G}_\sigma, \ \textrm{and} \ \eta(\bbC^\dagger) \in [\ylb, \yub],
\]
 where $\calG_\sigma$ denotes the set of all chemical graphs satisfying $\sigma$.

For this, we design an MILP formulation 
$\mathcal{M}_{\mathrm{GNN}}(g,x,y;\mathcal{C}_1, \mathcal{C}_2)$ consisting of two parts:
\begin{itemize}
\item[(i)] $\mathcal{M}_{\mathrm{GNN}}(x,y;\mathcal{C}_1)$: 
the part that simulates the computation process of the prediction function $\eta$, 
i.e., the graph neural network \GNN\ from Phase~1; and
\item[(ii)] $\mathcal{M}_{\mathrm{GNN}}(g,x;\mathcal{C}_2)$: 
the part of encoding the structural and chemical constraints of the target graph to ensure it
adheres the topological specification $\sigma$,
and the constraints to compute the initial node feature vectors $\theta_v^{(0)}$.
\end{itemize}

The basic idea of this formulation
remains similar to \molinfer.
In particular, the part $\mathcal{M}_{\mathrm{GNN}}(g,x;\mathcal{C}_2)$ 
 will be somehow similar to the one used in \molinfer, 
but the other one $\mathcal{M}_{\mathrm{GNN}}(x,y;\mathcal{C}_1)$ 
that simulates the computation process of a GNN is a challenging task,
and the extremely flexible rules of the topological specification make it even harder.
Here, 
we manage to realize the computation process of \GNN\ that is proposed in Section~\ref{sec:GNN_phase1} 
under the context of the two-layered model 
by paying special attention to the situation that a vertex/edge may be not selected 
in the resultant chemical graph.
We list the complete set of constraints for $\mathcal{M}_{\mathrm{GNN}}(g,x,y;\mathcal{C}_1, \mathcal{C}_2)$ 
in Appendix~\ref{sec:full_milp}, especially the ones for $\mathcal{M}_{\mathrm{GNN}}(x,y;\mathcal{C}_1)$
 in Appendix~\ref{sec:GNN_constraint}, since it is quite lengthy.

We acknowledge that several studies have already been done on 
utilizing MILP formulations to represent the computation process of a GNN
under the context of molecular optimization, e.g., \cite{McDonald:2024aa, Zhang:2024aa}.
However, our MILP formulation for the inverse problem differs significantly from these previous works,
in several key aspects:
\begin{itemize}
\item[-] Unlike methods that focus on optimizing the property value,
our approach aims to solve a feasibility problem,
find any valid chemical graph that satisfies the specified property range.
\item[-] The use of the 2L-model and topological specification provides
greater flexibility in defining the abstract structure of the target chemical graph,
while previous approaches like~\cite{McDonald:2024aa, Zhang:2024aa} often rely on 
simple or predetermined graph structures 
(e.g., fixed presence of one benzene ring, at least one sulfur atom, etc.),
greatly limiting the class of chemical graphs that can be inferred.
\end{itemize}

\section{Experimental Results}\label{sec:experiment}

We implemented our proposed GNN-based molecular inference framework \molinfergnn\
 and
conducted numerical experiments to evaluate its computational effectiveness. 
The model \GNN\ was implemented using the library PyTorch Geometric 2.7~\cite{Fey:2019aa}.
All the experiments were executed on a workstation with 
 a Core i9-9900K processor (3.6 GHz; 5.0 GHz at the maximum),
128 GB DDR4 RAM memory, and
an NVIDIA Quadro RTX 5000 GPU.

\subsection{Results of QSAR/QSPR Phase}

To assess the learning capabilities of our proposed \GNN\ model for predicting chemical properties, 
we used the publicly available benchmark QM9 dataset~\cite{Wu:2018aa, Ramakrishnan:2014aa}.
QM9 dataset comprises molecules composed of
Hydrogen ($\ttH$), Carbon ($\ttC$), Oxygen ($\ttO$), Nitrogen
($\ttN$), and Fluorine ($\ttF$) atoms, with each molecule containing up to 9 heavy (non-hydrogen) atoms.
In total, 
this dataset encompasses 133,885 drug-like organic molecules,
representing a diverse array of chemistry.
For each molecule in the QM9 dataset, {\em density functional theory} (B3LYP/6-31G(2df,p) based DFT)  is 
employed to determine a reasonable low-energy structure, 
and thereby providing access to atom positions, 
enable a set of intriguing and fundamental chemical properties to be computed~\cite{Wu:2018aa}.


Although a larger size of network will obtain a better learning performance in general,
the network size, and the time needed to solve the MILP formulation for the inverse problem in Phase~2 will
be increased drastically as well.
After some trials of preliminary experiments, 
we selected the following two variants of our proposed GNN model
for the sake of balancing the learning performance and the time effectiveness
of solving MILP formulations:
\begin{itemize}
\item[-] \GNN$_{16}$: a compact architecture with $\llmax=3, \phid = 16, \pcnv=32$; and
\item[-] \GNN$_{32}$: a larger architecture with $\llmax=3, \phid = 32,  \pcnv=32$.
\end{itemize}

Both models were trained and tested on the following chemical properties from the QM9 dataset:
\begin{itemize}
\setlength{\itemsep}{0cm}
\item[-] $\smu$: electric dipole moment (mD);
\item[-] $\sAlpha$: isotropic polarizability ($a_0^3$);
\item[-] $\sHomo$: energy of highest occupied molecular orbital (meV);
\item[-] $\sLumo$ energy of lowest occupied molecular orbital (meV);
\item[-] $\sGap$: the energy difference between $\sHomo$ and $\sLumo$ (meV);
\item[-] $\sRR$: electronic spatial extent ($a_0^2$);
\item[-] $\sZPVE$: zero point vibrational energy  (meV);
\item[-] $\sCv$: heat capacity at 298.15K (cal/molK).
\end{itemize}
To be consistent with previous works, 
we first remove 3,054 molecules that fail a geometric consistency check or are difficult to converge,
We use 110,000 molecules for training, 
10,000 for validation,
and the remaining 20,831 molecules are used for testing. 
We evaluate  the \emph{mean absolute error} (MAE) and the \emph{coefficient of determination} ($R^2$)
of each property.


\begin{table}[t!]
\caption{Results in Phase 1 on QM9 dataset with performance comparison to LLR in \molinfer~\cite{Zhu:2022ad}, 
SchNet~\cite{Schutt:2017aa}, MGCN~\cite{Lu:2019aa}, DimeNet++~\cite{Gasteiger:2022aa} and PAMNet~\cite{Zhang:2023ab}. The results of SchNet, MGCN, DimeNet++ and PAMNet are adapted from~\cite{Zhang:2023ab}.
}
\begin{center}
\scalebox{0.74}{
\begin{tabular}{c | c c | c | c  c c c c} \toprule
$\pi$ &  \GNN$_{16}$ & \GNN$_{32}$ & \molinfer\ (LLR)~\cite{Zhu:2022ad} &  SchNet~\cite{Schutt:2017aa} & MGCN~\cite{Lu:2019aa} & DimeNet++~\cite{Gasteiger:2022aa} & PAMNet~\cite{Zhang:2023ab} \\ 
Metric & MAE; $R^2$ &  MAE; $R^2$ &  MAE; $R^2$ & MAE & MAE & MAE & MAE \\ \midrule
$\smu$ &  588.930; 0.694 & 550.239; 0.733  & 759.067; 0.538 & 21 & 56 & 29.7 & 10.8 \\
$\sAlpha$ & 2.156; 0.850 & 1.799; 0.911 & 0.898; 0.975 &  0.124 & 0.030 & 0.0435 & 0.0447 \\
$\sHomo$ &  185.550; 0.828 & 158.878; 0.870 & 222.917; 0.774 & 47 & 42.1 & 24.6 & 22.8 \\
$\sLumo$ &  205.654; 0.952 & 171.943; 0.967 & 364.419; 0.873 & 39 & 57.4 & 19.5 & 19.2 \\
$\sGap$ & 277.472; 0.916  & 232.903; 0.940 & 432.729; 0.820 &  74 & 64.2 & 32.6 & 31.0 \\
$\sRR$ &  90.020; 0.796  & 79.073; 0.847  & 75.571; 0.849 & 0.158 & 0.11 & 0.331 & 0.093 \\
$\sZPVE$ &  110.482; 0.970 & 79.067; 0.984 & 16.743; 0.999 & 1.616 & 1.12  & 1.21 & 1.17\\
$\sCv$ &  0.9188; 0.896 & 0.7947; 0.934 & 0.4610; 0.977 & 0.034 & 0.038  & 0.0230 & 0.0231 \\
 \bottomrule
\end{tabular}
}
\end{center}\label{table:phase1a}
\end{table}

\begin{table}[t!]
\caption{Comparison of $R^2$ scores on the test set for \GNN$_{16}$ and \GNN$_{32}$ with other approaches in the \molinfer\ framework. The results for LLR, ANN, ALR, and R-MLR are adapted from~\cite{Zhu:2025aa}, and for HPS are adapted from~\cite{Zhu:2024aa}. ``-" means the corresponding result is not available in the
literature.
}
\begin{center}
\scalebox{0.95}{
\begin{tabular}{c | c c | c  c  c c c c} \toprule
$\pi$ &  \GNN$_{16}$ & \GNN$_{32}$ & LLR & ANN & ALR & R-MLR & HPS \\ \midrule 
$\smu$ &  0.694 &  0.733  & 0.367 & 0.409 & 0.403 & 0.645 & 0.708 \\
$\sAlpha$ &  0.850 & 0.911 & 0.961 &  0.888 & 0.953 & 0.980 & - \\
$\sHomo$ &  0.828 & 0.870 & 0.841 & 0.689 & 0.689 & 0.804 & 0.847 \\
$\sLumo$ &  0.952 & 0.967 & 0.841 & 0.860 & 0.833 & 0.920 & 0.948 \\
$\sGap$ &  0.916  &  0.940 & 0.784 &  0.795 & 0.755 & 0.876 & 0.907 \\
$\sCv$ &  0.896 & 0.934 & 0.970 & 0.911 & 0.966  & 0.978 & - \\
 \bottomrule
\end{tabular}
}
\end{center}\label{table:phase1a_molinfer}
\end{table}

Table~\ref{table:phase1a} presents a comparative analysis of \GNN$_{16}$ and \GNN$_{32}$
against the traditional LLR (Lasso linear regression)-based \molinfer\ approach and several existing state-of-the-art GNN-based models,
including SchNet~\cite{Schutt:2017aa}, MGCN~\cite{Lu:2019aa}, DimeNet++~\cite{Gasteiger:2022aa}, and PAMNet~\cite{Zhang:2023ab}.
Unlike previous \molinfer\ approaches,
the LLR experiments in Table~\ref{table:phase1a} 
were conducted on the whole QM9 dataset for a better comparison with \GNN.
A key distinction of our approach to other GNN-based models
 is that \GNN\ does not utilize 3D coordinate information
such as bond angles and interatomic distances, 
which are commonly leveraged by other models, including
the four mentioned models.
In addition, these models typically employ deeper architectures with more than 10 hidden layers.
Our model makes use of only 2D graphical information and 3 hidden layers,
and thus generally underperforms compared to them with no surprise. 
Nevertheless, our model can still obtain a commendable
result with $R^2$ exceeding 0.9 for the properties such as $\sAlpha, \sLumo, \sGap, \sZPVE$, and $\sCv$.

We also compared our new GNN-based approaches to other machine learning methods 
(LLR~\cite{Zhu:2022ad}, ANN~\cite{Azam:2021ab}, ALR~\cite{Zhu:2022ab}, R-MLR~\cite{Zhu:2025aa}, and HPS~\cite{Zhu:2024aa}) within the \molinfer\ framework, as shown 
in Table~\ref{table:phase1a_molinfer}.
Although a direct comparison is not entirely fair due to the differences in the dataset (the baseline models
were trained on a randomly selected subset of 1000 molecules from the whole QM9 dataset),
our results demonstrate that \GNN\ significantly improves the learning performance
for the properties such as $\smu, \sHomo, \sLumo$ and $\sGap$--propertues for which \molinfer\ approaches struggled to achieve
good learning performance.
These improvements highlight the limitations of handcrafted features used in previous models and showcase the advantage of GNNs in learning directly from molecular graph structures.
While the performance of \GNN\ on certain properties like $\sAlpha$ and $\sCv$
was slightly worse than the traditional LLR-based \molinfer\ approach,
it still achieved strong predictive performance--exceeding an $R^2$ score of 0.9 
with the \GNN$_{32}$ model.
This suggests that for some properties, carefully crafted features may still be critical
and capture key physicochemical characteristics more effectively than GNNs trained 
solely on 2D structures.
A hybrid approach like~\cite{Cai:2022aa}, 
combining GNN-based learning with selected handcrafted descriptors, 
may improve predictive accuracy.
This is left as our future work.

\subsection{Results of Inverse QSAR/QSPR Phase}


\begin{figure}[t!]
\begin{center}
\begin{tabular}{ccccc}
\includegraphics[width=.11\columnwidth]{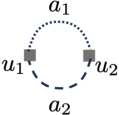} &
\includegraphics[width=.21\columnwidth]{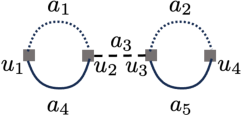} &
\includegraphics[width=.21\columnwidth]{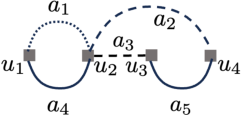} &
\includegraphics[width=.21\columnwidth]{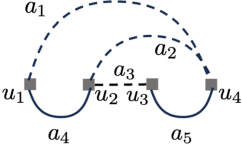} &
\includegraphics[width=.11\columnwidth]{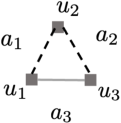} \\
(i) $\GC^1$ &
(ii) $\GC^2$ &
(iii) $\GC^3$ &
(iv) $\GC^4$ &
(v) $\GC^5$ \\
\end{tabular}
\end{center}
\caption{Illustrations of the seed graph $\GC^i$ of instance $I_i$, $i\in[1,5]$, respectively.
The edges in
$\Et$ are depicted with dotted lines,
the edges in $\Ew$ are depicted with dashed lines, the edges in $\Ez$ are depicted in gray bold lines 
and the edges in $\Eew$ are depicted with black solid lines. See Appendix~\ref{sec:specification}
for the description of the seed graphs and the sets $\Et$, $\Ew$, $\Ez$ and $\Eew$.
}
\label{fig:specification_example_b}
\end{figure}

In Phase~2, we formulated and solved MILP formulations to infer chemical graphs with desired property values
and pre-described abstract structures.
For the experiments, we use a set of four instances originally prepared by Zhu~et~al.~\cite{Zhu:2022ad},
namely $I_i, i\in[1,4]$, with slight modifications for this research,
and one new instance $I_5$. 
Instance $I_1$ is designed to infer chemical graphs with cycle index one (i.e., chemical graphs with one cycle), 
and $I_i, i\in[2,4]$ are designed to infer chemical graphs with cycle index two (i.e., chemical graphs with two cycles).
The basic settings for $I_i, i\in[1,4]$ are the same as in~\cite{Zhu:2022ad}, except for the set $\mathcal{F}$ of chemical rooted trees, and the lower and upper bounds on the number
of components in the target chemical graph. 
We also prepared an instance $I_5$ to imitate the chemical compounds appearing in the QM9 dataset. 
A 0/1 edge (depicted in gray, see Appendix~\ref{sec:specification} for the details) is introduced in $I_5$ to enable the generation of both 
a tree-like chemical graph and a chemical graph with cycle index one. 
The seed graphs $\GC^i$ of the five instances $I_i, i\in[1,5]$ are illustrated in 
Figure~\ref{fig:specification_example_b}.
See Appendix~\ref{sec:test_instances} for a detailed description of these instances.

We selected the properties $\sHomo$, $\sLumo$ and $\sGap$, and utilized the prediction functions 
constructed by \GNN$_{16}$ and \GNN$_{32}$ for the numerical experiments.
The three properties were selected because of the good learning performance obtained by GNN,
and the fact that it is easy to compute them by free and open-source softwares.
For each property and the corresponding prediction function, 
we formulated an MILP for each instance and
ten target value ranges which were selected based on the distribution of property values in the QM9 dataset.
The MILP formulations were solved using \textsc{CPLEX} 12.10, 
with a time limit of 1 hour for each formulation.

\begin{table}[t!]
\caption{
Number of inverse problems that are determined as feasible/infeasible/timeout among those having 10 different target ranges (\GNN$_{16}$): The time limit is set to 1 hour.
}
\begin{center}
\scalebox{1.00}{
\begin{tabular}{c | l l l l l  } \toprule
$\pi$ & 
\multicolumn{1}{c}{$I_1$} & \multicolumn{1}{c}{$I_2$} & \multicolumn{1}{c}{$I_3$} & 
\multicolumn{1}{c}{$I_4$} & \multicolumn{1}{c}{$I_5$} \\ \midrule
$\sHomo$ & 3/0/7 & 1/0/9 & 2/0/8 & 3/0/7 & 4/0/6 \\
$\sLumo$ & 7/0/3 & 0/0/10 & 1/0/9 & 4/0/6 & 5/0/5 \\
$\sGap$ & 4/0/6 & 1/0/9 & 0/0/10 & 1/0/9 & 7/0/3 \\
 \bottomrule
\end{tabular}
}
\end{center}\label{table:phase2_sol_16}
\end{table}

\begin{table}[t!]
\caption{
Number of inverse problems that are determined as feasible/infeasible/timeout among those having 10 different target ranges (\GNN$_{32}$): The time limit is set to 1 hour.
}
\begin{center}
\scalebox{1.00}{
\begin{tabular}{c | l l l l l  } \toprule
$\pi$ &  
\multicolumn{1}{c}{$I_1$} & \multicolumn{1}{c}{$I_2$} & \multicolumn{1}{c}{$I_3$} & 
\multicolumn{1}{c}{$I_4$} & \multicolumn{1}{c}{$I_5$} \\ \midrule
$\sHomo$ & 0/0/10 & 0/0/10 & 0/0/10 & 0/0/10 & 4/0/6 \\
$\sLumo$ & 0/0/10 & 0/0/10 & 0/0/10 & 0/0/10 & 6/0/4 \\
$\sGap$ & 1/0/9 & 0/0/10 & 1/0/9 & 2/0/8 & 6/0/4 \\
 \bottomrule
\end{tabular}
}
\end{center}\label{table:phase2_sol_32}
\end{table}

Tables~\ref{table:phase2_sol_16} and~\ref{table:phase2_sol_32} summarize the number of formulations
that could be determined within 1 hour for each property and instance with the prediction functions
constructed by \GNN$_{16}$ and \GNN$_{32}$, respectively.
In total, 43 out of 150 (28.7\%) formulations for \GNN$_{16}$ 
and 20 out of 150 (13.3\%) formulations for \GNN$_{32}$ were successful in obtaining a feasible solution within the time limit,
showing 
the hardness of the inverse problem.
Basically, the more complex model \GNN$_{32}$ needs more time to be solved,
matching the intuition that while a larger model improves learning performance in Phase~1,
the computational time for solving the corresponding MILP formulations
significantly increases in Phase~2.
Instance $I_5$ had the highest number of solved formulations, likely because it was designed
to resemble molecules in the QM9 dataset, while instances with more complex structures (e.g., $I_2$ and $I_3$)
showed lower success rates.

\begin{table}[t!]
\caption{Selected results of Phase 2 on properties $\sHomo$, $\sLumo$, and $\sGap$ 
with the model \GNN$_{16}$.
}
\begin{center}
\scalebox{0.95}{
\begin{tabular}{c | c c  c | c c c c c | c c  } \toprule
No. & $\pi$ & inst. & $\underline{y^*}, \overline{y^*}$ (eV) & \#v & \#c & I-time & $n$ & $\eta$ 
& $\eta_{\pyscf}$ & $\eta_{\psifour}$ \\ \midrule
(a) & $\sHomo$ & $I_1$ & -8.00, -7.50 & 10517 & 55112 & 40.644 & 19 & -7.63504 & -7.39256 & -7.71839 \\
(b) &                  & $I_2$ & -7.00, -6.50 & 10163 & 57570 & 4.134 & 10 & -6.51518 & -6.25516 & -7.31964 \\
(c) &                  & $I_4$ & -6.50, -6.00 & 10151 & 58342 & 4.123 & 10 & -6.27818 & -5.87852 & -6.38071 \\
(d) &                  & $I_5$ & -7.50, -7.00 & 7377 & 34508 & 19.008 & 9 & -7.01391 & -5.84124 & -6.36345 \\ \midrule
(e) & $\sLumo$ & $I_1$ & -2.00, -1.50 & 10517 & 55112 & 60.889 & 19 & -1.76264 & -2.01727 & -2.15115 \\
(f) &                   & $I_3$ & -3.50, -3.00 & 10159 & 57958 & 4.013 & 11 & -3.35504 & -4.74164 & Err.  \\
(g) &                  & $I_5$ & 1.00, 1.50 & 7377 & 34508 & 881.593 & 9 & 1.03216 & 1.14088 & 0.73261 \\ \midrule
(h) & $\sGap$ & $I_1$ & 6.00, 6.50 & 10517 & 55112 & 62.145 & 20 & 6.27711 & 6.07699 & 6.28504 \\
(i) &                 & $I_5$ & 7.00, 7.50 & 7377 & 34508 & 753.747 & 9 & 7.13251 & 6.45730 & 6.48175 \\
 \bottomrule
\end{tabular}
}
\end{center}\label{table:phase2_result_16}
\end{table}

\begin{table}[t!]
\caption{Selected results of Phase 2 on properties $\sHomo$, $\sLumo$, and $\sGap$ 
with the model \GNN$_{32}$.
}
\begin{center}
\scalebox{0.95}{
\begin{tabular}{c | c c  c | c c c c c | c c  } \toprule
No. & $\pi$ & inst. & $\underline{y^*}, \overline{y^*}$ (eV) & \#v & \#c & I-time & $n$ & $\eta$ 
& $\eta_{\pyscf}$ & $\eta_{\psifour}$ \\ \midrule
(j) & $\sHomo$ &  $I_5$ & -6.00, -5.50 & 12785 & 67516 & 3262.404 & 9 & -5.51096 & -6.01995 & Err. \\ \midrule
(k) & $\sLumo$ & $I_5$ & 1.00, 1.50 & 12785 & 67516 & 6.355 & 9 & 1.49046 & 1.53046 & Err. \\ \midrule
(l) & $\sGap$ &  $I_1$ & 4.50, 5.00 & 18325 & 108328 & 618.418 & 17 & 4.98681 & 4.86001 & 5.13423 \\ 
 \bottomrule
\end{tabular}
}
\end{center}\label{table:phase2_result_32}
\end{table}

To further assess the quality of our inferred chemical graphs,
we used two free and open-source quantum chemistry software \pyscf~\cite{Sun:2018aa, Sun:2020aa}
and \psifour~\cite{Smith:2020aa} to compute the property values
using DFT at the B3LYP/6-31G(2df,p) level
and compare them against the predicted values obtained from our constructed prediction functions.

We summarize some selected results about MILP formulations in Tables~\ref{table:phase2_result_16} and~\ref{table:phase2_result_32} for the model \GNN$_{16}$ and \GNN$_{32}$, respectively. 
The following notations are used:
\begin{itemize}
\item[-] No.: numbering of the MILP;
\item[-] $\pi$: the property;
\item[-] inst.: instance;
\item[-] $\underline{y^*}, \overline{y^*}$: range $[\underline{y^*}, \overline{y^*}]$ (in eV) of target value
of the chemical graph to be inferred;
\item[-] \#v: the number of variables in the MILP formulation;
\item[-] \#c: the number of constraints in the MILP formulation;
\item[-] I-time: the time (in seconds) to solve the MILP by CPLEX 12.10;
\item[-] $n$: the number of non-hydrogen atoms in the inferred chemical graph;
\item[-] $\eta$: the predicted property value by our constructed prediction function;
\item[-] $\eta_{\pyscf}$: the predicted property value by the software \pyscf~\cite{Sun:2018aa, Sun:2020aa}; and
\item[-] $\eta_{\psifour}$: the predicted property value by the software \psifour~\cite{Smith:2020aa} (Err. indicates that an error occurred during the computation).
\end{itemize}

\begin{figure}[t!]
\begin{center}
\begin{tabular}{cccc}
\includegraphics[width=.20\columnwidth]{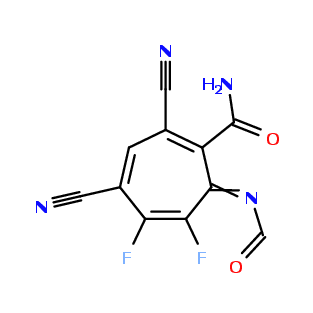} &
\includegraphics[width=.20\columnwidth]{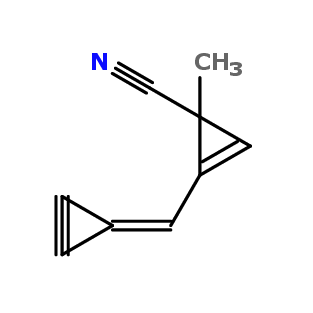} &
\includegraphics[width=.20\columnwidth]{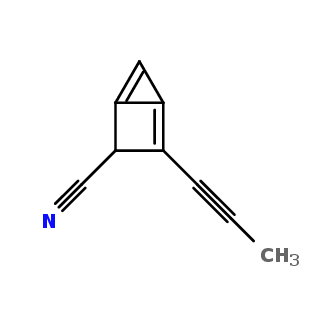} &
\includegraphics[width=.20\columnwidth]{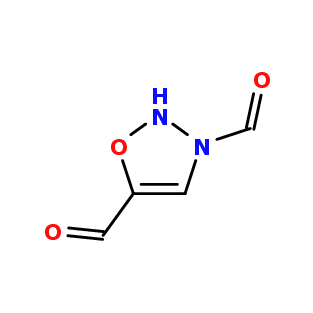} \\
(a) & (b) & (c) & (d) \\
\includegraphics[width=.20\columnwidth]{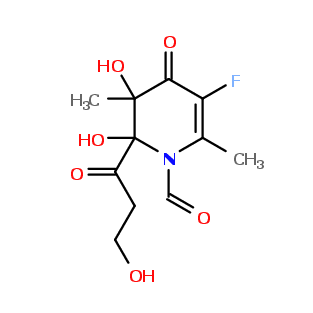} &
\includegraphics[width=.20\columnwidth]{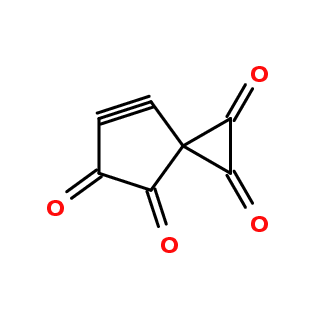} &
\includegraphics[width=.20\columnwidth]{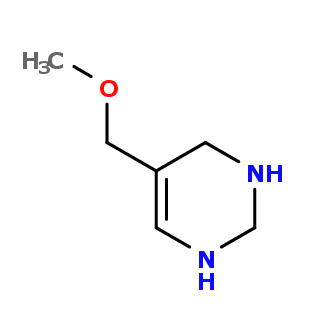} &
\includegraphics[width=.20\columnwidth]{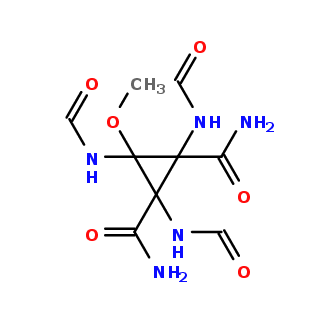} \\
(e) & (f) & (g) & (h) \\
\includegraphics[width=.20\columnwidth]{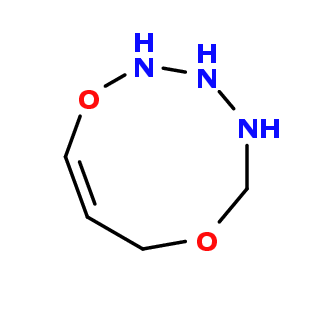} &
\includegraphics[width=.20\columnwidth]{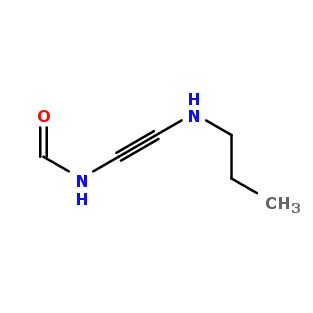} &
\includegraphics[width=.20\columnwidth]{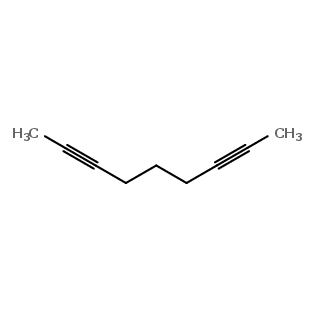} &
\includegraphics[width=.20\columnwidth]{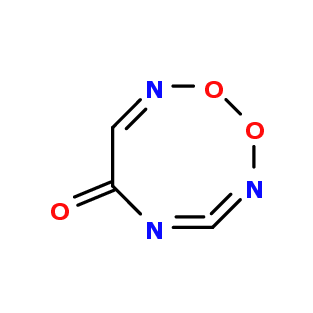} \\
(i) & (j) & (k) & (l) \\
\end{tabular}
\end{center}
\caption{Illustrations of the inferred chemical graphs. (a)-(l) correspond to the first column No. in Tables~\ref{table:phase2_result_16} and~\ref{table:phase2_result_32}, respectively.}
\label{fig:phase2_result_svg}
\end{figure}

The inferred molecules contained up to 20 non-hydrogen atoms, demonstrating
the practical applicability of our approach.
We notice that the predicted values obtained from our prediction function sometimes differ
significantly from those computed by \pyscf\ and \psifour\ (e.g., the chemical graphs (d), (e) and (j)).
We can also see that
the differences between the values computed by \pyscf\ and \psifour\ are big,
pointing out the instability of DFT methods.
In certain cases, \psifour\ calculations failed 
due to chemically unstable structures, 
indicating potential limitations in the MILP-generated molecular graphs.
The issue of unstable generated structures is common in molecular inference studies and 
highlights the need to incorporate chemical knowledge~\cite{Cheng:2021aa}.
Nevertheless, the inferred chemical graphs were of rather good quality,
demonstrating the potential of our proposed framework.
Figure~\ref{fig:phase2_result_svg} provides graphical representations of the inferred chemical graphs.
The results of our experiments indicate that \molinfergnn\ successfully 
integrates GNN-based learning with MILP-driven molecular inference. 

\section{Concluding Remarks}\label{sec:conclude}

In this study, we proposed \molinfergnn, a novel inverse QSAR/QSPR framework that integrates
GNNs into the existing \molinfer\ approach.
A key contribution of our work is the successful formulation
of the inverse problem of GNN within the 2L-model of \molinfer\
as an MILP formulation.
This enables the combination of
the deep learning-based molecular property prediction with the
rigorous inference capabilities of MILP.
The use of the 2L-model plays a crucial role in our framework,
as it significantly broadens the class of chemical graphs that can be effectively inferred
compared with previous studies.
By leveraging GNNs, our framework allows to use the chemical graphs directly,
eliminating the need for handcrafted feature vectors while enhancing prediction accuracy.
We validated our approach through computational experiments on the QM9 dataset.
The results demonstrated that our proposed simple GNN model, \GNN, can effectively
predict molecular properties using only 2D graphical information, which
is necessary for the inverse problem to be formulated as MILP formulations at present.
Our experiments also indicated that the framework can infer chemical graphs with up to
20 non-hydrogen atoms, 
highlighting its potential applications in cheminformatics and molecular design.

However, several challenges remain.
The computational cost of solving MILP formulations remains high,
particularly as the complexity of the target molecular graphs and GNN architectures increases.
Future research should focus more on optimizing the inverse problem formulations, 
using heuristics to solve MILP formulations,
and exploring alternative graph learning architectures.
Discrepancies between predicted and software-computed property values 
suggest that further improvements are needed in both GNN training and MILP constraint formulations. 
Enhancing the graph representation learning and incorporating additional chemical constraints could mitigate these issues.
Additionally, trying to incorporate 3D information about molecules is also desired since they can
significantly improve prediction accuracy and enhance the overall effectiveness of the framework.
These are left as future works.
We believe with further refinements and a more integrated hybrid approach, 
our framework has the potential to significantly advance computational molecular design and inverse QSAR/QSPR research.

\section*{Acknowledgement}\label{sec:acknowledgement}
This work is partially supported by JSPS KAKENHI Grant Numbers JP22H00532 and
JP22KJ1979.

\bibliographystyle{abbrv}
\bibliography{./chemgraph.bib}
 
\appendix
 \newpage
 \section*{Appendix}

\section{Specifying Target Chemical Graphs}\label{sec:specification} 

In this section, we review the way of specifying target chemical graphs
in the two-layered model introduced by Zhu~et~al.~\cite{Zhu:2022ad},
with some modifications for this work.

\begin{figure}[h!] \begin{center}
\includegraphics[width=.98\columnwidth]{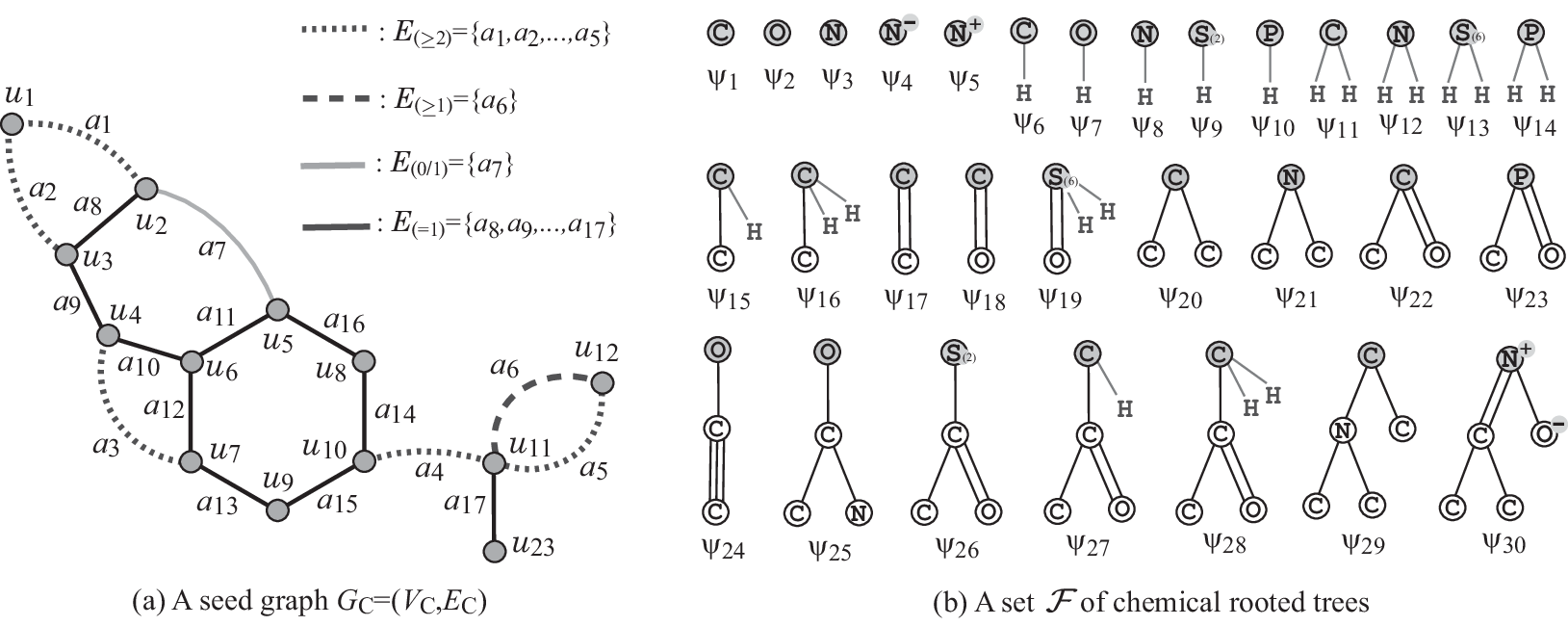}
\end{center} \caption{
(a) An illustration of a seed graph $\GC$ with $\mathrm{r}(\GC)=5$
where the vertices in $\VC$ are depicted with gray circles,
the edges in $\Et$ are depicted with dotted lines,
the edges in $\Ew$ are depicted with dashed lines,
the edges in $\Ez$ are depicted with gray bold lines and  
the edges in $\Eew$ are depicted with black solid lines;
(b) A set $\mathcal{F}=\{\psi_1,\psi_2,\ldots,\psi_{30}\}\subseteq
\mathcal{F}(D_\pi)$ of 30 chemical rooted trees
$\psi_i, i\in [1,30]$, where the root of each tree is depicted with a gray circle, 
where  the hydrogens attached to non-root vertices are omitted in the figure.    }
\label{fig:specification_example_1} \end{figure}  

Given a prediction function $\eta$ and 
a target value $y^*\in \mathbb{R}$, 
we call a chemical graph $\C^*$ such that $\eta(\C^*)=y^*$
 a {\em target chemical graph}.
This section  presents a set of rules for 
 specifying  topological substructure
  of a target chemical graph in a flexible way in Phase~2.

We first describe how to reduce a chemical graph $\C=(H,\alpha,\beta)$ into
an abstract form based on which our specification rules will be defined.
To illustrate the reduction process,
we use the chemical graph $\C=(H,\alpha,\beta)$
such that $\anC$ is given in Figure~\ref{fig:example_chemical_graph}.
 
\begin{figure}[t!] \begin{center}
\includegraphics[width=.65\columnwidth]{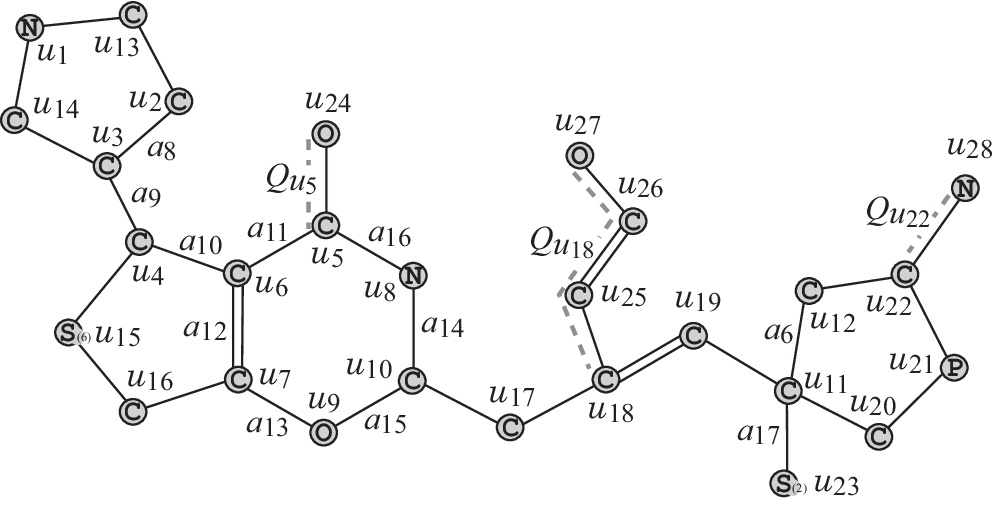}
\end{center} \caption{The interior $H^\inte$ of
chemical graph $\C$ with $\anC$ 
  in Figure~\ref{fig:example_chemical_graph} for ${\rho}=2$.
}
\label{fig:specification_example_interior} \end{figure}

\begin{figure}[t!] \begin{center}
\includegraphics[width=.98\columnwidth]{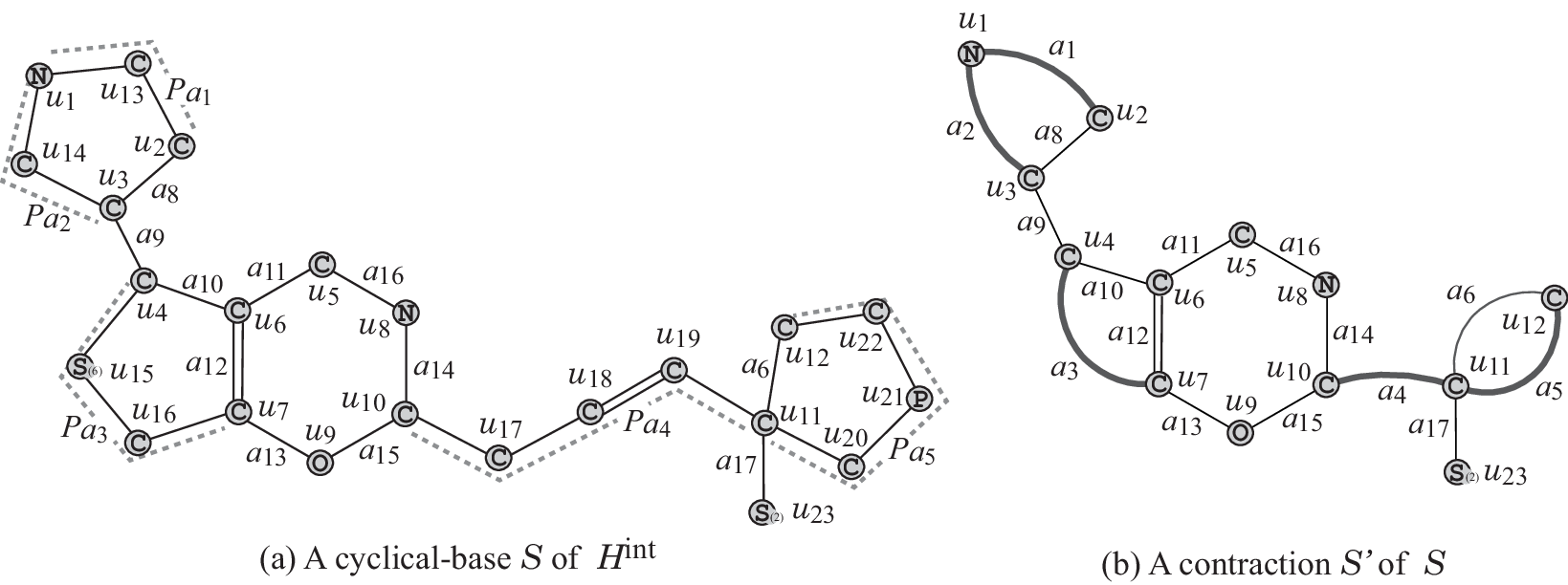}
\end{center} \caption{
(a) A cyclical-base  
$S=H^\inte- \bigcup_{u\in \{u_5,u_{18},u_{22}\}}(V(Q_u)\setminus \{u\})$
of the interior  $H^\inte$ in Figure~\ref{fig:specification_example_interior};
(b) A contraction $S'$ of  $S$ for a pure path set 
 $\mathcal{P}=\{P_{a_1},P_{a_2},\ldots,P_{a_5}\}$ 
in (a),
where a new edge obtained by contracting a pure path is depicted
with a thick line.   
}
\label{fig:specification_example_R2_3} \end{figure}

 \begin{enumerate}
 \item[R1] {\bf Removal of all ${\rho}$-fringe-trees: } 
The interior $H^\inte=(V^\inte(\C),E^\inte(\C))$ of $\C$ 
is obtained by removing the non-root vertices of 
each ${\rho}$-fringe-trees $\C[u]\in\mathcal{T}(\C), u\in V^\inte(\C)$. 
Figure~\ref{fig:specification_example_interior} illustrates
the interior $H^\inte$ of 
chemical graph $\C$ with ${\rho}=2$
  in Figure~\ref{fig:example_chemical_graph}. 
  
 \item[R2] {\bf Removal of some leaf paths: } 
 We call a $u,v$-path $Q$ in $H^\inte$  a {\em leaf path} if 
  vertex $v$ is a leaf-vertex of $H^\inte$
  and the degree of each internal vertex of $Q$  in $H^\inte$  is 2,
  where we regard that $Q$ is rooted at vertex $u$. 
A connected subgraph $S$ of the interior $H^\inte$ of $\C$  
is called a {\em cyclical-base}
if $S$ is obtained from $H$
by removing the vertices in $V(Q_u)\setminus \{u \}, u\in X$ 
for a subset $X$ of interior-vertices  and a set  $\{Q_u \mid u\in X\}$ of leaf 
 $u,v$-paths $Q_u$  such that    
 no two paths $Q_u$ and $Q_{u'}$ share a vertex.
Figure~\ref{fig:specification_example_R2_3}(a) illustrates
a cyclical-base  $S=H^\inte- \bigcup_{u\in X}(V(Q_u)\setminus \{u\})$
of the interior  $H^\inte$  
for a set 
$\{Q_{u_5}=(u_5,u_{24}), 
     Q_{u_{18}}=(u_{18},u_{25},u_{26},u_{27}),
     Q_{u_{22}}=(u_{22},u_{28})\}$ of leaf  paths 
in Figure~\ref{fig:specification_example_interior}.  

 \item[R3] {\bf Contraction of some pure paths: } 
 A path in $S$ is called {\em pure} 
 if  each internal vertex of the path  is of degree 2. 
 Choose a set $\mathcal{P}$ of several pure paths in $S$ 
 so that no two paths share  vertices except for their end-vertices. 
 A graph $S'$ is called a {\em contraction} of a graph $S$
  (with respect to $\mathcal{P}$) 
 if $S'$ is obtained from $S$ by replacing 
 each pure $u,v$-path  with a single edge $a=uv$,
 where $S'$ may contain multiple edges between the same pair of adjacent vertices.
Figure~\ref{fig:specification_example_R2_3}(b) illustrates
a contraction $S'$ obtained from 
the chemical graph  $S$
by contracting each $uv$-path $P_a\in  \mathcal{P}$ into a new edge $a=uv$,
where $a_1=u_1 u_{2},  a_2=u_1 u_{3},  a_3=u_4 u_{7}, a_4=u_{10}u_{11}$
and $a_5=u_{11}u_{12}$ and 
 $\mathcal{P}=\{
 P_{a_1}=(u_1,u_{13},u_{2}), 
 P_{a_2}=(u_{1},u_{14},u_{3}),
 P_{a_3}=(u_{4},u_{15},u_{16},u_{7}), 
 P_{a_4}=(u_{10},u_{17},u_{18},u_{19},u_{11}),
 P_{a_5}=(u_{11},u_{20},u_{21},u_{22},u_{12})\}$ of pure paths 
in Figure~\ref{fig:specification_example_R2_3}(a). 
\end{enumerate}

We will define a set of rules so that 
a chemical graph can be obtained 
from a graph (called a seed graph in the next section)
by applying processes R3 to R1 in a reverse way. 
We specify topological substructures of a target chemical graph
with a tuple  $(\GC,\sint,\sce)$  called  a {\em target specification}
defined under the set of the following rules. 

\subsection*{Seed Graph}

A  {\em seed graph} $\GC=(\VC,\EC)$ is defined
to be a graph (possibly with multiple edges) such that 
the edge set $\EC$ consists of four sets 
$\Et$, $\Ew$, $\Ez$ and $\Eew$, 
where each of them can be empty.
A seed graph plays a role of the most abstract form $S'$ in R3.  
Figure~\ref{fig:specification_example_1}(a) illustrates an example of a seed graph
$\GC$ with $\mathrm{r}(\GC)=5$,   
where $\VC=\{u_1,u_2,\ldots,u_{12},u_{23}\}$, 
$\Et=\{a_1,a_2,\ldots,a_5\}$, 
$\Ew=\{a_6\}$,
$\Ez=\{a_7\}$ and 
$\Eew=\{a_8,a_9,\ldots,a_{16}\}$.

 A {\em subdivision} $S$ of $\GC$  
is a graph constructed from a seed graph $\GC$ 
according to the following rules:
\begin{enumerate}[leftmargin=*]
\item[-]
Each edge $e=uv\in \Et$ is replaced
with a $u,v$-path $P_e$ of length at least 2;

\item[-] 
Each edge $e=uv\in \Ew$ is replaced
with a $u,v$-path $P_e$ of length at least 1
(equivalently $e$ is directly used or replaced with
a $u,v$-path $P_e$ of length at least 2);

\item[-] 
Each edge $e\in \Ez$ is either used or discarded, where 
 $\Ez$ is required to be chosen as a non-separating edge subset of
 $E(\GC)$ since otherwise the connectivity of a final chemical graph $\Co$
 is not guaranteed; 
$\mathrm{r}(\Co)= \mathrm{r}(\GC)-|E'|$ holds
for a subset $E'\subseteq \Ez$ of edges discarded 
in a  final chemical graph $\Co$; 
and 

\item[-]
Each edge $e\in \Eew$ is always used directly. 
\end{enumerate}

We allow a possible elimination of edges in $\Ez$ as an optional rule
in constructing a target chemical graph from a seed graph, 
even though such an operation has 
not been included in the process R3. 
A subdivision  $S$ plays a role of a cyclical-base   in R2. 
A target chemical graph $\C=(H,\alpha,\beta)$ will contain  $S$  as a subgraph
of the interior $H^\inte$ of $\C$.


\subsection*{Interior-specification}

A graph $H^*$ that serves as the interior $H^\inte$ of
a target chemical graph $\C$ will be constructed as follows.
First construct a subdivision  $S$ of a seed graph $\GC$ 
by replacing each edge $e=u u'\in \Et\cup\Ew$
with a pure $u,u'$-path $P_e$.
Next construct a supergraph $H^*$ of $S$ by 
attaching a leaf path $Q_v$ at each vertex $v\in \VC$ or
at an internal vertex $v\in V(P_e)\setminus\{u,u'\}$ 
of each pure $u,u'$-path $P_e$ for some edge $e=uu'\in \Et\cup\Ew$,
where possibly $Q_v=(v), E(Q_v)=\emptyset$ 
(i.e., we do not attach any new edges to $v$).
We introduce the following rules for specifying
 the size of $H^*$, the length $|E(P_e)|$  of
a pure path  $P_e$,  the length $|E(Q_v)|$ of
a   leaf path $Q_v$, the number of  leaf paths $Q_v$
and a bond-multiplicity of each interior-edge,
where we call the set of prescribed constants  
 an  {\em interior-specification}   $\sint$: 
\begin{enumerate}[leftmargin=*]
 \item[-]
  Lower and upper bounds $\nint_\LB, \nint_\UB\in \mathbb{Z}_+$ 
  on   the number of interior-vertices 
of a target chemical graph~$\C$. 
  
\item[-] 
For each edge $e=u u'\in \Et\cup\Ew$, 
\begin{description} 
\item[]
 a lower bound $\ell_{\LB}(e)$ and 
 an upper bound $\ell_{\UB}(e)$  on the length $|E(P_e)|$ of
 a pure $u,u'$-path $P_e$. 
(For a notational convenience, set 
$\ell_\LB(e):=0$, $\ell_\UB(e):=1$, $e\in \Ez$ and
$\ell_\LB(e):=1$, $\ell_\UB(e):=1$, $e\in \Eew$.)
   
\item[]  
 a lower bound $\bl_{\LB}(e)$ and 
 an upper bound $\bl_{\UB}(e)$ on the number of leaf paths $Q_v$ attached 
 at  internal vertices $v$ of a pure $u,u'$-path $P_e$.   

\item[] 
 a lower bound $\ch_{\LB}(e)$ and 
 an upper bound $\ch_{\UB}(e)$  on the maximum 
 length  $|E(Q_v)|$ of a leaf path $Q_v$ attached  
 at an internal vertex $v\in V(P_e)\setminus\{u,u'\}$ 
 of a pure $u,u'$-path $P_e$.   
\end{description} 

\item[-]
For each vertex $v\in \VC$, 
\begin{description} 
\item[]
 a lower bound $\ch_{\LB}(v)$ and 
 an upper bound $\ch_{\UB}(v)$  on  
 the number of leaf paths $Q_v$ attached to $v$,
 where $0\leq \ch_{\LB}(v)\leq \ch_{\UB}(v)\leq 1$.
 
\item[]
 a lower bound $\ch_{\LB}(v)$ and 
 an upper bound $\ch_{\UB}(v)$  on the
 length $|E(Q_v)|$ of a leaf path $Q_v$ attached to $v$. 
\end{description}  

\item[-] 
For each edge $e=u u'\in \EC$, 
a lower bound $\bd_{m, \LB}(e)$ 
and an  upper bound $\bd_{m, \UB}(e)$  on
the number of edges with bond-multiplicity $m\in [2,3]$ in
$u,u'$-path $P_e$, where we regard $P_e$, $e  \in \Ez\cup \Eew$ 
as single edge $e$.
\end{enumerate}

We call a graph $H^*$ that satisfies an interior-specification $\sint$
a {\em $\sint$-extension of $\GC$}, 
where the bond-multiplicity of each edge has been determined.

Table~\ref{table:interior-spec}  shows an example of 
an interior-specification  $\sint$ to the seed graph  $\GC$ in 
Figure~\ref{fig:specification_example_1}. 

\begin{table}[t!]\caption{Example~1 of an interior-specification  $\sint$. }
\begin{tabular}{ |  c | c |  } \hline 
$\nint_\LB=20$ & $\nint_\UB = 28$ \\\hline 
\end{tabular}

 \begin{tabular}{ |  c | c c c c c c |  } \hline
                        & $a_1$ &  $a_2$ &   $a_3$ &   $a_4$ &   $a_5$ &   $a_6$   \\\hline
 $\ell_\LB(a_i)$&  2 &  2 &  2 & 3 &  2 &  1 \\ \hline
 $\ell_\UB(a_i)$&  3 & 4 &  3 & 5 & 4 &  4 \\\hline
 $\bl_\LB(a_i)$&  0 &  0 &   0 & 1 &  1 &   0 \\ \hline
 $\bl_\UB(a_i)$&  1 & 1 &   0 & 2 & 1 &   0 \\\hline
 $\ch_\LB(a_i)$&  0 &  1 & 0 & 4 &  3 &  0 \\ \hline
 $\ch_\UB(a_i)$&  3 & 3 &  1 & 6 & 5 &  2 \\\hline
\end{tabular} 

\begin{tabular}{ |  c | c c c c c c   c c c c  c c c |  } \hline
                        & $u_1$ &  $u_2$ &   $u_3$ &   $u_4$ &   $u_5$ &   $u_6$ 
                       & $u_7$ &   $u_8$ &   $u_9$ &   $u_{10}$ &   $u_{11}$ 
                       &   $u_{12}$ &   $u_{23}$ \\\hline 
 $\bl_\LB(u_i)$&  0 &  0 &   0 & 0 &  0 &   0
                       & 0 &   0 &  0 &   0 &  0 &  0 &  0 \\ \hline
 $\bl_\UB(u_i)$&  1 & 1 &   1 & 1 & 1 &   0
                       & 0 &   0 &  0 &   0 &  0 &  0 &  0\\\hline
 $\ch_\LB(u_i)$&  0 &  0 &   0 & 0 &  1 &   0
                       & 0 &   0 &  0 &   0 &  0 &  0 &  0 \\ \hline
 $\ch_\UB(u_i)$&  1 & 0 &   0 & 0 & 3 &   0
                       & 1 &   1 &  0 &   1 &  2 & 4 &  1 \\\hline
\end{tabular} 

\begin{tabular}{ |  c | c c c c c c   c c c c c c  c c c c c |  } \hline
                               & $a_1$ &  $a_2$ &   $a_3$ &   $a_4$ &   $a_5$ &   $a_6$ 
                               & $a_7$ &  $a_8$ &   $a_9$ &   $a_{10}$ &   $a_{11}$ &   $a_{12}$ 
                               & $a_{13}$ &   $a_{14}$ &   $a_{15}$ &   $a_{16}$ &   $a_{17}$  \\\hline
 $\bd_{2, \LB}(a_i)$ &  0    &  0 &   0 & 1 &  0 &   0
                                &  0   &  0 &  0 & 0 &  0 &   1
                                &  0    &  0 &   0 & 0     & 0  \\ \hline
 $ \bd_{2, \UB}(a_i)$&  1    & 1 &   0 & 2  & 2 &   0  
                                &  0    & 0&   0 & 0 &  0 &   1
                                &  0    &  0 &   0 & 0   & 0   \\ \hline
 $\bd_{3, \LB}(a_i)$ &  0    &  0 &   0 & 0 &  0 &   0
                                &  0   &  0 &  0 & 0 &  0 &   0
                                &  0    &  0 &   0 & 0   & 0   \\ \hline
 $ \bd_{3, \UB}(a_i)$&  0    & 0 &   0 & 0  & 1 &   0 
                                &  0    &  0 &   0 & 0 &  0 &   0
                                &  0    &  0 &   0 &  0    & 0   \\ \hline
\end{tabular} 
\label{table:interior-spec}  
\end{table}

\begin{figure}[t!] \begin{center}
\includegraphics[width=.58\columnwidth]{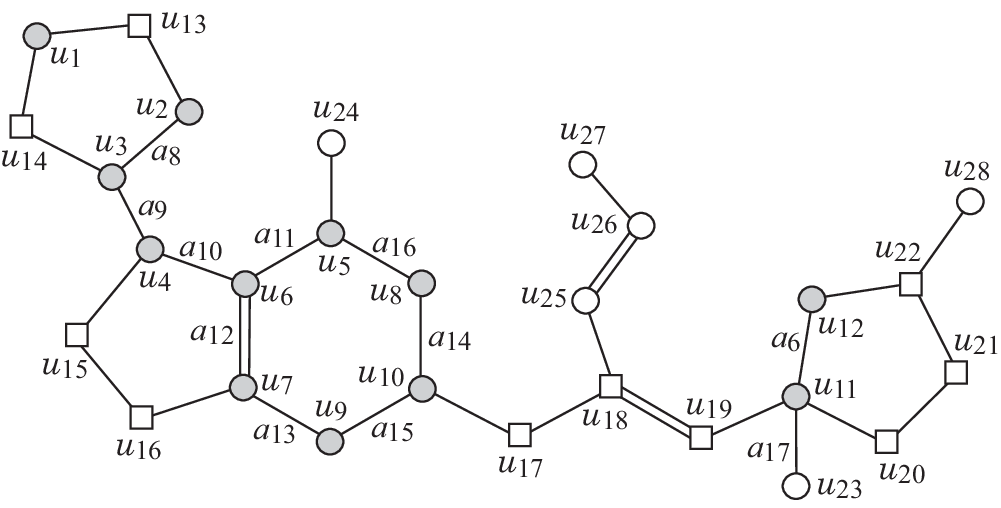}
\end{center} \caption{
An illustration of a graph 
$H^*$ that is obtained from  the seed graph  $\GC$ in 
Figure~\ref{fig:specification_example_1}
under the interior-specification  $\sint$ in 
Table~\ref{table:interior-spec},
where the vertices newly introduced by pure paths $P_{a_i}$
and leaf paths $Q_{v_i}$ are depicted with white squares and circles,
respectively.    }
\label{fig:specification_example_3} \end{figure}

Figure~\ref{fig:specification_example_3} illustrates an example of 
an $\sint$-extension $H^*$ of seed graph  $\GC$ in 
Figure~\ref{fig:specification_example_1}
under the interior-specification  $\sint$ in 
Table~\ref{table:interior-spec}.  

\subsection*{Chemical-specification}
 
 Let $H^*$ be a graph that serves as 
 the interior $H^\inte$ of a target chemical graph $\C$,
 where the bond-multiplicity of each edge in $H^*$ has be determined.
 Finally we introduce a set of rules for constructing 
   a target chemical graph $\C$ from $H^*$ 
   by choosing  a chemical element $\ta\in \Lambda$ 
and assigning a ${\rho}$-fringe-tree $\psi$
 to each interior-vertex $v\in V^\inte$. 
We introduce the following rules for specifying
the size of $\C$, a set of chemical rooted trees  
that are allowed to use as  ${\rho}$-fringe-trees 
and lower and upper bounds on the frequency of
a chemical element, a chemical symbol, 
and an edge-configuration,
where we call the set of prescribed constants   
 a  {\em chemical specification} $\sce$:   
 
\begin{enumerate}[leftmargin=*]
\item[-] 
Lower and upper bounds $n_\LB,  n^*\in \mathbb{Z}_+$
on the number of vertices, where $\nint_\LB \leq n_\LB\leq n^*$.
 
\item[-] 
Subsets  $\mathcal{F}(v) \subseteq \mathcal{F}(D_\pi), v\in \VC$ 
and $\mathcal{F}_E \subseteq \mathcal{F}(D_\pi)$ 
 of chemical rooted trees $\psi$ with $\h(\anpsi)\leq {\rho}$, where 
 we require that 
 every ${\rho}$-fringe-tree $\C[v]$ rooted at a vertex $v\in \VC$
 (resp., at an internal vertex $v$ not in $\VC$)   in  $\C$ 
 belongs to $\mathcal{F}(v)$ (resp.,   $\mathcal{F}_E$).  
Let  $\mathcal{F}^*:=\mathcal{F}_E\cup \bigcup_{v\in \VC}\mathcal{F}(v)$
and 
$\Lambda^\ex$ denote the set of  chemical elements assigned to non-root
vertices over all chemical rooted trees in $\mathcal{F}^*$.  
 
\item[-] 
A subset  $\Lambda^\inte\subseteq \Lambda^\inte(D_\pi)$, where 
 we require that every chemical element $\alpha(v)$ 
 assigned to an interior-vertex  $v$ in $\C$ belongs to $\Lambda^\inte$.
Let $\Lambda:= \Lambda^\inte\cup \Lambda^\ex$ and
 $\na_\ta(\C)$ (resp., $\na_\ta^\inte(\C)$ and $\na_\ta^\ex(\C)$) 
 denote the number of vertices   (resp.,   interior-vertices and  exterior-vertices)
  $v$ such that $\alpha(v)=\ta$   in  $\C$.
 
\item[-] 
A set $\Ldg^\inte\subseteq \Lambda\times [1,4]$  of chemical  symbols
and  a set $\Gamma^\inte \subseteq \Gamma^\inte(D_\pi)$  
of  edge-configurations  $(\mu,\mu' ,m)$ with $\mu \leq \mu'$, where 
 we require that the edge-configuration $\ec(e)$ of an interior-edge $e$ in $\C$ 
 belongs to $\Gamma^\inte$.
We do not distinguish  $(\mu,\mu' ,m)$ and $(\mu' , \mu,m)$.

\item[-] 
Define  $\Gac^\inte$ to be the set of   adjacency-configurations such that  
$\Gac^\inte:=\{(\ta, \tb, m) \mid (\ta d, \tb d',m)\in \Gamma^\inte\}$.   
Let  $\ac_\nu^\inte(\C), \nu\in \Gac^\inte$   
denote  the number of  interior-edges $e$ such that $\ac(e)=\nu$  in $\C$.
  
\item[-] 
 Subsets $\Lambda^*(v)\subseteq \{\ta\in \Lambda^\inte\mid \val(\ta)\geq 2\}$, 
 $v\in \VC$,  
 we require that every chemical element $\alpha(v)$ 
 assigned to   a vertex $v\in  \VC$
 in the seed graph  belongs to $\Lambda^*(v)$.  

\item[-] Lower and upper bound functions 
$\na_\LB,\na_\UB: \Lambda\to  [1,n^*]$  and 
$\na_\LB^\inte,\na_\UB^\inte: \Lambda^\inte\to  [1,n^*]$ 
on the number of   interior-vertices  $v$ such that  $\alpha(v)=\ta$  in $\C$. 

%
%
 
 \item[-] Lower and upper bound functions  
$\fc_\LB,\fc_\UB: \mathcal{F}^*\to  [0,n^*]$ 
  on the number of   interior-vertices $v$ 
  such that $\C[v]$ is r-isomorphic to $\psi\in \mathcal{F}^*$  in $\C$.   
  
 \item[-] Lower and upper bound functions  
$\ac^\lf_\LB,\ac^\lf_\UB: \Gac^\lf \to  [0,n^*]$ 
  on the number of  leaf-edges $uv$ in $\acC$
  with adjacency-configuration $\nu$.  
\end{enumerate}
 
We call a chemical graph $\C$ that satisfies a chemical specification $\sce$
a {\em $(\sint,\sce)$-extension of $\GC$},
and denote by $\mathcal{G}(\GC, \sint,\sce)$ the set of
all $(\sint,\sce)$-extensions of $\GC$. 

Table~\ref{table:chemical_spec}  shows an example of 
a chemical-specification  $\sce$ to the seed graph  $\GC$
 in Figure~\ref{fig:specification_example_1}. 
 

\begin{table}[t!]\caption{Example~2 of a chemical-specification  $\sce$.  
}
\begin{tabular}{ |  l |  } \hline
 $n_\LB=30$,  $n^* =50$. \\\hline
  branch-parameter:   ${\rho}=2$  \\\hline
\end{tabular}

\begin{tabular}{ |  l |  } \hline
 Each of sets $\mathcal{F}(v), v\in \VC$ and
 $\mathcal{F}_E$ is set to be \\
 the set $\mathcal{F}$  of chemical rooted trees $\psi$ with $\h(\anpsi)\leq {\rho}=2$
in Figure~\ref{fig:specification_example_1}(b). \\\hline
\end{tabular}

\begin{tabular}{ |  c | c |   } \hline
  $\Lambda=\{ \ttH,\ttC,\ttN,\ttO, \ttS_{(2)},\ttS_{(6)}, \ttP=\ttP_{(5)}\}$ & 
  $\Ldg^{\inte} =\{ \ttC2 , \ttC3,  \ttC4, \ttN2, \ttN3, \ttO2,
    \ttS_{(2)}2,  \ttS_{(6)}3, \ttP4   \}$  
\\\hline
\end{tabular}

\begin{tabular}{ |  c | l |  } \hline
  $\Gac^{\inte}$ &
  $ \nu_1 \!=\!(\ttC   , \ttC  , 1) ,   \nu_2 \!=\!(\ttC   , \ttC  , 2) ,   
   \nu_3 \!=\!(\ttC   , \ttN  , 1) ,  \nu_4 \!=\!(\ttC  , \ttO  , 1), 
    \nu_5 \!=\! (\ttC, \ttS_{(2)}, 1),\nu_6 \!=\!(\ttC  , \ttS_{(6)}, 1), 
    \nu_7 \!=\! (\ttC  , \ttP  , 1) $  \\ \hline
\end{tabular}

\begin{tabular}{ |  c | l |  } \hline
  $\Gamma^{\inte}$ &
  $ \gamma_1 \!=\! (\ttC 2 , \ttC 2, 1) ,
   \gamma_2 \!=\!(\ttC 2 , \ttC 3, 1) ,  
   \gamma_3 \!=\!(\ttC 2 , \ttC 3, 2) ,  
   \gamma_4 \!=\!(\ttC 2 , \ttC 4, 1) , 
   \gamma_5 \!=\!(\ttC 3 , \ttC 3, 1) , 
   \gamma_6 \!=\!(\ttC 3 , \ttC 3, 2) , $ \\
   &
  $   
    \gamma_7 \!=\!(\ttC 3 , \ttC 4, 1), 
   \gamma_8 \!=\!(\ttC 2 , \ttN 2, 1) ,  
   \gamma_9 \!=\!(\ttC 3 , \ttN 2, 1) ,  
   \gamma_{10} \!=\!(\ttC 3 , \ttO 2, 1), 
    \gamma_{11} \!=\!(\ttC 2 , \ttC 2, 2),  
    \gamma_{12} \!=\!(\ttC 2 , \ttO 2, 1) ,$ \\
   &
  $  
    \gamma_{13} \!=\!(\ttC 3 , \ttN3, 1), 
    \gamma_{14} \!=\!(\ttC 4, \ttS_{(2)} 2, 2),  
    \gamma_{15} \!=\!(\ttC 2 , \ttS_{(6)}3, 1), 
   \gamma_{16} \!=\!(\ttC 3 , \ttS_{\tiny (6)}3, 1), 
    \gamma_{17} \!=\!(\ttC 2, \ttP4, 2), $ \\
   &
  $  
    \gamma_{18} \!=\!(\ttC 3, \ttP4, 1)  
     $ \\ \hline
\end{tabular}
    
\begin{tabular}{ |  l|  } \hline
$\Lambda^*(u_1)=\Lambda^*(u_8)=\{{\tt C,  N}\}$, 
$\Lambda^*(u_9)=\{{\tt C, O}\}$, 
   $\Lambda^*(u)=\{\ttC\}$, $u\in \VC\setminus\{u_1,u_8,u_9\}$
   \\\hline
\end{tabular}

\begin{tabular}{ |  c | c c c c  c c c |  } \hline
                         & ${\tt H}$  & ${\tt C}$ &   ${\tt N}$ &     ${\tt O}$ 
                         & $\ttS_{(2)}$ & $\ttS_{(6)}$ & $\ttP$  \\\hline
 $\na_\LB(\ta)$ & 40 &  27 &  1 &   1 & 0 & 0 & 0   \\ \hline 
 $\na_\UB(\ta)$ & 65 & 37 & 4 &  8  &   1 &   1 &   1 \\\hline
\end{tabular} 
\begin{tabular}{ |  c | c c c  c c c   |  } \hline
   & $\ttC$ &   $\ttN$ &     $\ttO$  & $\ttS_{(2)}$ & $\ttS_{(6)}$ & $\ttP$  \\\hline
 $\na_\LB^{\inte}(\ta)$ &   9 &  1 &   0  & 0 & 0 & 0      \\ \hline
 $\na_\UB^{\inte}(\ta) $&  23 & 4 & 5 &   1 &   1 &   1  \\\hline
\end{tabular} 

%
%

\begin{tabular}{ |  c | c   c   |  } \hline 
& $\psi\in\{\psi_i\mid i=1,6,11\}$ 
& $\psi\in \mathcal{F}^*\setminus \{\psi_i\mid i=1,6,11\}$ \\\hline
 $\fc_\LB(\psi)$  &  1 &    0   \\ \hline 
 $\fc_\UB(\psi)$ &  10 &  3\\\hline
\end{tabular}

\begin{tabular}{ |  c | c   c   |  } \hline 
& $\nu\in\{(\ttC,\ttC,1),(\ttC,\ttC,2)  \}$ 
& $\nu\in \Gac^\lf \setminus \{(\ttC,\ttC,1),(\ttC,\ttC,2)  \}$   \\\hline
 $\ac^\lf_\LB(\nu)$  &  0 &    0   \\ \hline 
 $\ac^\lf_\UB(\nu)$ &  10 &  8 \\\hline
\end{tabular} 

\label{table:chemical_spec}
\end{table}

Figure~\ref{fig:example_chemical_graph} 
 illustrates an example $\Co$ of 
a   $(\sint,\sce)$-extension of $\GC$   obtained 
from the  $\sint$-extension $H^*$  
 in Figure~\ref{fig:specification_example_3} 
under the chemical-specification $\sce$ in Table~\ref{table:chemical_spec}.  
Note that $\mathrm{r}(\Co)= \mathrm{r}(H^*)= \mathrm{r}(\GC)-1=4$
 holds since the edge in $\Ez$ is discarded in $H^*$.

\section{Test Instances for Phase~2}\label{sec:test_instances} 
We prepared the following instances $I^i, i\in[1,5]$ for conducting experiments
 in Phase~2.

\begin{itemize} 
  \item[(1)] $I_i=(\GC^i,\sint^i, \sce^i)$, $i=1,2,3,4$:
 An instance for inferring chemical graphs with rank at most 2.  
In the four instances $I_i$, $i=1,2,3,4$, 
the following specifications in $(\sint,\sce)$ are common. 
\begin{enumerate}
\item[] 
Set  $\Lambda:=\Lambda(\pi)$
 for a given property $\pi\in \{${\sc Homo, Lumo, Gap}$\}$, 
 set $\Ldg^\inte$ to be
the set of all possible symbols in $\Lambda\times[1,4]$  
that appear in the data set $D_\pi$  
and set $\Gamma^\inte$
to be the set  of  all  edge-configurations that appear in the data set $D_\pi$. 
Set  $\Lambda^*(v):= \Lambda$,  $v\in \VC$. 
 
\item[] 
The lower bounds  
 $\ell_\LB $, $\bl_\LB $, $\ch_\LB $,  
 $\bd_{2,\LB}$,   $\bd_{3,\LB}$,  
 $\na_\LB$,  $\na^\inte_\LB$,  $\ns^\inte_\LB$,  
$\ac^\inte_\LB$, $\ec^\inte_\LB$ and $\ac^\lf_\LB$  are all set to be 0.

\item[] 
The upper bounds  
 $\ell_\UB $, $\bl_\UB $, $\ch_\UB $,  
 $\bd_{2,\UB}$,   $\bd_{3,\UB}$,  
 $\na_\UB$,  $\na^\inte_\UB$,  $\ns^\inte_\UB$,  
$\ac^\inte_\UB$, $\ec^\inte_\UB$ and $\ac^\lf_\UB$ 
are all set to be an upper bound $n^*$  on $n(G^*)$.
 
\item[] 
   For each property $\pi$, let $\mathcal{F}(D_\pi)$ denote
    the set of 2-fringe-trees in the compounds in $D_\pi$,
   and select a subset $\mathcal{F}_\pi^i\subseteq  \mathcal{F}(D_\pi)$ with
   $|\mathcal{F}_\pi^i|=45-5i$, $i\in [1,5]$.
   For each instance $I_i$, 
   set $\mathcal{F}_E :=\mathcal{F}(v):=  \mathcal{F}_\pi^i$, $v\in \VC$ and 
$\fc_\LB(\psi):=0, \fc_\UB(\psi):=10, \psi\in  \mathcal{F}_\pi^i$. 
\end{enumerate}
 
  Instance $I_1$ is given   by the rank-1 seed graph $\GC^1$ 
  in Figure~\ref{fig:specification_example_b}(i)
  and   Instances $I_i$, $i=2,3,4$ are
   given by  the rank-2 seed graph $\GC^i$, $i=2,3,4$ in 
   Figure~\ref{fig:specification_example_b}(ii)-(iv).

\begin{itemize} 
 \item[(i)]  For the instance $I_1$, select as a seed graph 
  the monocyclic graph   $\GC^1=(\VC,\EC=\Et\cup \Ew)$
  in Figure~\ref{fig:specification_example_b}(i),
  where $\VC=\{u_1,u_2\}$, $\Et=\{a_1\}$ and  $ \Ew=\{a_2\}$. 
Set $\nint_\LB:=6, \nint_\UB:=8, n_\LB:=15$ and $n^*:=20$.

 \item[(ii)]
 For the instance $I_2$, select as a seed graph 
  the  graph   $\GC^2=(\VC,\EC=\Et\cup \Ew\cup \Eew)$ 
  in Figure~\ref{fig:specification_example_b}(ii),
  where
$\VC=\{u_1,u_2,u_3,u_4\}$, 
$\Et=\{a_1,a_2\}$, 
$\Ew=\{a_3\}$  and 
$\Eew=\{a_4,a_5\}$. 
Set $\nint_\LB:=6, \nint_\UB:=12, n_\LB:=10$ and $n^*:=15$. 
%
    
 \item[(iii)]
 For the instance $I_3$, select as a seed graph 
  the  graph   $\GC^3=(\VC,\EC=\Et\cup \Ew\cup \Eew)$ 
  in Figure~\ref{fig:specification_example_b}(iii),   where
$\VC=\{u_1,u_2,u_3,u_4\}$, 
$\Et=\{a_1\}$, 
$\Ew=\{a_2, a_3\}$  and 
$\Eew=\{a_4,a_5\}$. 
Set $\nint_\LB:=6, \nint_\UB:=12, n_\LB:=10$ and $n^*:=15$. 

 \item[(iv)] 
 For the instance $I_4$, select as a seed graph 
  the  graph   $\GC^4=(\VC,\EC=\Et\cup \Ew\cup \Eew)$ 
  in Figure~\ref{fig:specification_example_b}(iv),   where
$\VC=\{u_1,u_2,u_3,u_4\}$, 
$\Ew=\{a_1, a_2, a_3\}$  and 
$\Eew=\{a_4,a_5\}$. 
Set $\nint_\LB:=6, \nint_\UB:=12, n_\LB:=10$ and $n^*:=15$. 
 \end{itemize}
 
  \item[(2)] $I_5=(\GC^5,\sint^5, \sce^5)$:
 An instance for inferring chemical graphs 
 resembling the molecules in the QM9 dataset.
 Below is the specification $(\sint^5, \sce^5)$ for $I_5$.
 
 \begin{enumerate}
\item[] 
Set  $\Lambda:=\Lambda(\pi)$
 for a given property $\pi\in \{${\sc Homo, Lumo, Gap}$\}$, 
 set $\Ldg^\inte$ to be
the set of all possible symbols in $\Lambda\times[1,4]$  
that appear in the data set $D_\pi$  
and set $\Gamma^\inte$
to be the set  of  all  edge-configurations that appear in the data set $D_\pi$. 
Set  $\Lambda^*(v):= \Lambda$,  $v\in \VC$. 
 
\item[] 
The lower bounds  
 $\ell_\LB $, $\bl_\LB $, $\ch_\LB $,  
 $\bd_{2,\LB}$,   $\bd_{3,\LB}$,  
 $\na_\LB$,  $\na^\inte_\LB$,  $\ns^\inte_\LB$,  
$\ac^\inte_\LB$, $\ec^\inte_\LB$ and $\ac^\lf_\LB$  are all set to be 0.

\item[] 
The upper bounds  
 $\ell_\UB $, $\bl_\UB $, $\ch_\UB $,  
 $\bd_{2,\UB}$,   $\bd_{3,\UB}$,  
 $\na_\UB$,  $\na^\inte_\UB$,  $\ns^\inte_\UB$,  
$\ac^\inte_\UB$, $\ec^\inte_\UB$ and $\ac^\lf_\UB$ 
are all set to be an upper bound $n^*$  on $n(G^*)$.
 
\item[] 
   For each property $\pi$, let $\mathcal{F}(D_\pi)$ denote
    the set of 2-fringe-trees in the compounds in $D_\pi$,
   and select a subset $\mathcal{F}_\pi^5\subseteq  \mathcal{F}(D_\pi)$ with
   $|\mathcal{F}_\pi^5|=50$.
   For each instance $I_i$, 
   set $\mathcal{F}_E :=\mathcal{F}(v):=  \mathcal{F}_\pi^i$, $v\in \VC$ and 
$\fc_\LB(\psi):=0, \fc_\UB(\psi):=10, \psi\in  \mathcal{F}_\pi^i$. 
\end{enumerate}
 
 Instance $I_5$ is given by the seed graph $\GC^5$ in Figure~\ref{fig:specification_example_b}(v). 
 
 \begin{itemize}
 \item[(v)]
 For the instance $I_5$, select as a seed graph 
  the  graph   $\GC^5=(\VC,\EC=\Et\cup \Ez)$ 
  in Figure~\ref{fig:specification_example_b}(v),
  where
$\VC=\{u_1,u_2,u_3\}$, 
$\Et=\{a_1,a_2\}$ and
$\Ez=\{a_3\}$. 
Set $\nint_\LB:=3, \nint_\UB:=9, n_\LB:=3$ and $n^*:=9$. 
\end{itemize}
 
 \end{itemize}

\section{All Constraints in an MILP Formulation for Chemical Graphs}\label{sec:full_milp}


We define a standard encoding of a finite set $A$ of elements
to be a bijection $\sigma: A \to [1, |A|]$, 
where we denote by $[A]$   the set $[1, |A|]$ of integers
and by $[{\tt e}]$ the encoded element $\sigma({\tt e})$.
Let $\epsilon$ denote {\em null}, a fictitious chemical element 
that does not belong to any set of chemical elements,
chemical symbols, adjacency-configurations and
edge-configurations in the following formulation.
Given a finite set $A$, let $A_\epsilon$ denote the set $A\cup\{\epsilon\}$
and define a standard encoding of $A_\epsilon$
  to be a bijection $\sigma: A \to [0, |A|]$ such that
$\sigma(\epsilon)=0$, 
where we denote by $[A_\epsilon]$   the set $[0, |A|]$ of integers
and by $[{\tt e}]$ the encoded element $\sigma({\tt e})$,
where $[\epsilon]=0$.

 \bigskip 
 Let $\sigma=(\GC,\sint,\sce)$ be a target specification,
 ${\rho}$ denote  the branch-parameter in the specification $\sigma$
 and  $\C$ denote a chemical   graph in $\mathcal{G}(\GC, \sint,\sce)$. 

 \subsection{Selecting  a Cyclical-base} 
\label{sec:co}
 
Recall that  
\[ \begin{array}{ll}
   \Eew = \{e\in \EC\mid \ell_\LB(e)=\ell_\UB(e)=1 \}; &
   \Ez =\{e\in \EC\mid \ell_\LB(e)=0, \ell_\UB(e)=1 \}; \\
  \Ew=\{e\in \EC\mid \ell_\LB(e)=1,  \ell_\UB(e)\geq 2 \}; &
  \Et= \{e\in \EC\mid \ell_\LB(e)\geq 2 \}; \end{array} \]
\begin{enumerate} [leftmargin=*]
\item[-]
Every edge $a_i\in \Eew$ is  included in  $\anC$;

\item[-]
Each edge $a_i\in \Ez$ is   included in $\anC$ if necessary;
 
\item[-]
For each edge  $a_i  \in \Et$, edge $a_i$ is not included in $\anC$
and instead a path 
\[P_i=(\vC_{\tail(i)}, \vT_{j-1},\vT_{j},\ldots,
    \vT_{j+t}, \vC_{\hd(i)})\]
     of length at least 2
  from vertex $\vC_{\tail(i)}$ to vertex $\vC_{\hd(i)}$ 
  visiting some  vertices in $\VT$ is constructed in $\anC$; and  
 
\item[-]
For each edge $a_i  \in \Ew$, either  edge $a_i$   is directly used in $\anC$ or
the above path $P_i$ of length at least 2   is constructed in $\anC$.  
 \end{enumerate}
 
Let  $\tC\triangleq |\VC|$ and denote $\VC$ by 
$\{\vC_{i}\mid i\in [1,\tC]\}$.
Regard the seed graph $\GC$ as a digraph such that
each edge $a_i$ with end-vertices $\vC_{j}$ and $\vC_{j'}$
is directed from  $\vC_{j}$ to $\vC_{j'}$ when $j<j'$.
 For each directed edge $a_i  \in \EC $,
 let $\hd(i)$ and $\tail(i)$ denote the head and tail of $\eC(i)$;
 i.e., $a_i=(\vC_{\tail(i)}, \vC_{\hd(i)})$. 
  
Define 
 \[ \kC \triangleq  |\Et\cup \Ew| , ~~ \widetilde{\kC} \triangleq  |\Et| ,\]
 and denote   $\EC=\{a_i\mid i\in[1,\mC]\}$,
$\Et=\{a_k\mid k\in[1,\widetilde{\kC}]\}$,
$\Ew=\{a_k\mid k\in[\widetilde{\kC}+1,\kC]\}$,
$\Ez=\{a_i\mid i\in[\kC+1,\kC+|\Ez|]\}$ and 
$\Eew=\{a_i\mid i\in[\kC+|\Ez|+1,\mC]\}$.
Let $\Iew$ denote the set of indices $i$ of edges $a_i\in \Eew$.
Similarly for $\Iz$, $\Iw$  and $\It$.

To control the construction of such a path $P_i$
 for each edge  $a_k\in  \Et\cup \Ew $,
we regard the index $k\in [1,\kC]$ of each edge $a_k\in  \Et\cup \Ew$
as the ``color'' of the edge.
To introduce necessary linear constraints 
that can construct such a path $P_k$ properly   in our MILP,
we assign the color $k$ to the vertices $\vT_{j-1},\vT_{j},\ldots,$ 
$\vT_{j+t}$ in $\VT$
when the above path  $P_k$ is used in $\anC$.
 
For each index $s\in [1,\tC]$, let  
$\IC(s)$ denote the set of edges $e\in \EC$ incident to vertex $\vC_{s}$,
and 
 $\Eew^+(s)$ (resp., $\Eew^-(s)$) denote the set of 
 edges $a_i\in \Eew$ such that 
the tail (resp., head) of $a_i$ is vertex $\vC_{s}$.
Similarly for 
$\Ez^+(s)$,  $\Ez^-(s)$, $\Ew^+(s)$,  $\Ew^-(s)$,
$\Et^+(s)$ and $\Et^-(s)$.
Let $\IC(s)$ denote the set of indices $i$ of edges $a_i\in \IC(s)$.
Similarly for   
$\Iew^+(s)$,  $\Iew^-(s)$,
$\Iz^+(s)$,  $\Iz^-(s)$, 
$\Iw^+(s)$,  $\Iw^-(s)$,
$\It^+(s)$ and $\It^-(s)$.
Note that $[1, \kC]=\It\cup \Iw$ and 
$[\widetilde{\kC}+1,\mC]=\Iw\cup \Iz\cup\Iew$.

\smallskip\noindent
{\bf constants: } 
\begin{enumerate} [leftmargin=*]
\item[-] $\tC=|\VC|$, $\widetilde{\kC}=  |\Et|$, $\kC= |\Et\cup \Ew|$,
      $\tT=\nint_\UB-|\VC|$, $\mC=|\EC|$.
      Note that 
      $a_i\in \EC\setminus (\Et\cup \Ew)$ holds $i\in [\kC+1,\mC]$;   

\item[-] $\ell_\LB(k), \ell_\UB(k)\in [1, \tT]$, $k\in [1,\kC]$: 
lower and upper bounds on the length of path $P_k$;  
       
\item[-] $r_{\GC}\in[1,\mC]$: the rank $\mathrm{r}(\GC)$ of
seed graph $\GC$;   
\end{enumerate}

\smallskip\noindent
{\bf variables: } 
\begin{enumerate}[leftmargin=*]
\item[-] $\eC(i)\in[0,1]$,  $i\in [1, \mC]$: 
$\eC(i)$ represents edge $a_i\in \EC$, $i\in [1,\mC]$  
 ($\eC(i)=1$, $i\in \Iew$;  $\eC(i)=0$, $i\in \It$)     
  ($\eC(i)=1$ $\Leftrightarrow $   edge $a_i$ is  used in  $\anC$);    
\item[-]  $\vT(i)\in[0,1]$,   $i\in [1,\tT]$:  
  $\vT(i)=1$ $\Leftrightarrow $ vertex $\vT_{i}$ is used in  $\anC$;   
\item[-]  $\eT(i)\in[0,1]$, $i\in [1,\tT+1]$:  $\eT(i)$ represents edge 
$\eT_{i}=(\vT_{i-1}, \vT_{i})\in \ET$,  
where $\eT_{1}$ and $\eT_{\tT+1}$ are fictitious edges
  ($\eT(i)=1$ $\Leftrightarrow $   edge $\eT_{i}$ is  used in  $\anC$);    
\item[-]  $\chiT(i)\in [0,\kC]$, $i\in [1,\tT]$: $\chiT(i)$ represents
 the color assigned to vertex $\vT_{i}$ 
  ($\chiT(i)=k>0$
   $\Leftrightarrow $  vertex $\vT_{i}$ is  assigned color $k$;
   $\chiT(i)=0$ means that vertex $\vT_{i}$ is not used in $\anC$);    
   
\item[-]  $\clrT(k)\in [\ell_\LB(k)-1, \ell_\UB(k)-1]$, $k\in [1,\kC]$, 
$\clrT(0)\in [0, \tT]$: the number of vertices 
$\vT_{i}\in \VT$  with color $c$;
%
\item[-]  $\dclrT(k)\in [0,1]$,   $k\in [0,\kC]$:
      $\dclrT(k)=1$    $\Leftrightarrow $ $\chiT(i)=k$ 
      for some $i\in [1,\tT]$;
      
\item[-]    $\chiT(i,k)\in[0,1]$,  $i\in [1,\tT]$, $k\in [0,\kC]$  
  ($\chiT(i,k)=1$    $\Leftrightarrow $ $\chiT(i)=k$);  
\item[-]   $\tldgC^+(i)\in [0,4]$, $i\in [1,\tC]$: 
the out-degree of vertex $\vC_{i}$ with the used edges $\eC$ in $\EC$; 

\item[-]   $\tldgC^-(i)\in [0,4]$, $i\in [1,\tC]$: 
the in-degree of vertex $\vC_{i}$  with the used edges $\eC$ in $\EC$; 

\item[-] $\mathrm{rank}$:  the rank $\mathrm{r}(\C)$ of a target 
chemical graph $\C$;   
\end{enumerate}
  
\smallskip\noindent
{\bf constraints: }   
\begin{align} 
 \mathrm{rank} =  r_{\GC} -\sum_{i\in \Iz}(1-\eC(i)), && \label{eq:co_rank} \\
  \eC(i)=1,  ~~~  i\in \Iew,       &&  \label{eq:co_first}  \\
  \eC(i)=0,  ~~ \clrT(i)\geq 1,   ~~~  i\in \It,     &&   \label{eq:co_first} \\
  \eC(i)+ \clrT(i)\geq 1,  ~~~~~  \clrT(i)\leq \tT\cdot (1-\eC(i) ), 
~~~  i\in \Iw,    &&    \label{eq:co1c} 
\end{align}   
  
\begin{align}  
\sum_{ c\in \Iw^-(i)\cup \Iz^-(i)\cup \Iew^-(i) }\!\!\!\!\!\! \eC(c) 
 = \tldgC^-(i),  ~~ 
\sum_{ c\in \Iw^+(i)\cup \Iz^+(i)\cup \Iew^+(i) }\!\!\!\!\!\! \eC(c) 
 = \tldgC^+(i),  &&   i\in [1,\tC],   \label{eq:co_5}
\end{align}   
\begin{align} 
\chiT(i,0)=1 -\vT(i), ~~~
\sum_{k\in [0,\kC]} \chiT(i,k)=1,  ~~~ 
\sum_{k\in [0,\kC]}k\cdot \chiT(i,k)=\chiT(i),  && i\in[1,\tT],  \label{eq:co2} 
\end{align}   

\begin{align}  
\sum_{i\in[1,\tT]} \chiT(i,k)=\clrT(k), ~~
\tT\cdot \dclrT(k)\geq  \sum_{i\in [1,\tT]} \chiT(i,k)
\geq \dclrT(k), &&  k\in [0,\kC],    \label{eq:co3}   
\end{align}     
 
\begin{align}  
\vT(i-1)\geq \vT(i), && \notag \\
 \kC\cdot (\vT(i-1)-\eT(i )) \geq \chiT(i-1)-\chiT(i )
  \geq \vT(i-1) - \eT(i ), && i\in[2,\tT]. \label{eq:co_last} 
 \end{align}

\subsection{Constraints for Including Leaf Paths} 
\label{sec:int}

Let
$\widetilde{\tC}$  denote the number of vertices $u\in \VC$ such that 
$\bl_\UB(u)=1$ and assume that 
$\VC=\{u_1,u_2,\ldots, u_p\}$ so that 
\[ \mbox{ 
$\bl_\UB(u_i)=1$, $i\in [1,\widetilde{\tC}]$ and 
$\bl_\UB(u_i)=0$, $i\in[\widetilde{\tC}+1, \tC]$. }\]
Define the set of colors for the vertex set 
$\{u_i\mid i\in [1,\widetilde{\tC}] \}\cup \VT$
 to be $[1,\cF]$ with 
\[ \cF \triangleq \widetilde{\tC} + \tT 
=|\{u_i\mid i\in[1,\widetilde{\tC}]\}\cup \VT|. \]
Let each  vertex   $\vC_{i}$, $i\in[1,\widetilde{\tC}]$ 
(resp., $\vT_{i}\in \VT$)
  correspond to 
a color $i\in [1,\cF]$ (resp., $i+\widetilde{\tC} \in [1,\cF]$). 
When a path $P=(u, \vF_{j}, \vF_{j+1},\ldots, \vF_{j+t})$ 
from a vertex $u\in \VC\cup \VT$ 
  is used in $\anC$, we assign the color $i\in [1,\cF]$ of the vertex $u$
to the vertices $\vF_{j}, \vF_{j+1},\ldots, \vF_{j+t}\in \VF$.

\smallskip\noindent
{\bf constants: } 
\begin{enumerate}[leftmargin=*]
\item[-] $\cF$: the maximum number of different colors 
assigned to the vertices in $\VF$;  

\item[-]   $n^*$: an upper bound  
on the number $n(\C)$ of non-hydrogen atoms in $\C$;  

\item[-] $\nint_\LB, \nint_\UB \in [2,n^* ]$:
 lower and upper bounds on
the number of interior-vertices in $\C$; 

\item[-] $\bl_\LB(i) \in [0,1]$,  $i\in [1, \widetilde{\tC}]$: 
a lower   bound  on the number of leaf ${\rho}$-branches  in
the leaf path rooted  at a vertex $\vC_{i}$; 

\item[-]  $\bl_\LB(k),\bl_\UB(k)\in [0,\ell_\UB(k)-1]$, 
 $k\in[1,\kC]=\It\cup\Iw$: 
lower and upper bounds on the number of 
leaf ${\rho}$-branches in the trees rooted at internal vertices 
of a pure path $P_k$  for an edge $a_k\in \Ew\cup \Et$; 
\end{enumerate}

\smallskip\noindent
{\bf variables: } 
\begin{enumerate}[leftmargin=*]
  
\item[-]   $\nint_G\in [\nint_\LB, \nint_\UB]$: 
the number of interior-vertices in $\C$; 

\item[-]  $\vF(i)\in[0,1]$,   $i\in [1,\tF]$:
 $\vF(i)=1$ $\Leftrightarrow $ vertex $\vF_{i}$ is used in  $\C$;   
 
\item[-]  $\eF(i)\in[0,1]$, $i\in [1,\tF+1]$:  $\eF(i)$ represents edge 
$\eF_{i}=\vF_{i-1} \vF_{i}$,  
where $\eF_{1}$ and $\eF_{\tF+1}$ are fictitious edges
  ($\eF(i)=1$ $\Leftrightarrow $   edge $\eF_{i}$ is  used in  $\C$);    
\item[-]  $\chiF(i)\in [0,\cF]$, $i\in [1,\tF]$: $\chiF(i)$ represents
 the color assigned to  vertex $\vF_{i}$  
  ($\chiF(i)=c$ $\Leftrightarrow $  vertex $\vF_{i}$ is  assigned color $c$);   
  
\item[-]  $\clrF(c)\in [0, \tF]$, $c\in [0,\cF]$: the number of vertices $\vF_{i}$
 with color $c$;  
\item[-]  $\dclrF(c)\in [\bl_\LB(c), 1]$,  $c\in [1, \widetilde{\tC}]$:
      $\dclrF(c)=1$    $\Leftrightarrow $ $\chiF(i)=c$ for some $i\in [1,\tF]$;  
\item[-]  $\dclrF(c)\in[0,1]$,  $c\in [\widetilde{\tC}+1,\cF]$:
      $\dclrF(c)=1$    $\Leftrightarrow $ $\chiF(i)=c$ for some $i\in [1,\tF]$;  
\item[-]    $\chiF(i,c)\in[0,1]$,
 $i\in [1,\tF]$, $c\in [0,\cF]$:  
   $\chiF(i,c)=1$    $\Leftrightarrow $ $\chiF(i)=c$;     
\item[-]  $\bl(k,i)\in [0,1]$, $k\in[1,\kC]= \It\cup\Iw$,  $i\in[1,\tT]$: 
    $\bl(k,i)=1$ $\Leftrightarrow$ path $P_k$ contains vertex $\vT_{i}$ 
    as an internal vertex
    and the ${\rho}$-fringe-tree rooted at $\vT_{i}$ contains a leaf ${\rho}$-branch;
\end{enumerate}
  
\smallskip\noindent
{\bf constraints: }   
\begin{align} 
\chiF(i,0)=1 -\vF(i), ~~~
\sum_{c\in [0,\cF]} \chiF(i,c)=1,  ~~~ 
\sum_{c\in [0,\cF]}c\cdot \chiF(i,c)=\chiF(i),  &&  i\in[1,\tF],  \label{eq:int_first} 
\end{align}   

\begin{align}  
\sum_{i\in[1,\tF]} \chiF(i,c)=\clrF(c), ~~~ \tF\cdot \dclrF(c)\geq
\sum_{i\in [1,\tF]} \chiF(i,c)\geq \dclrF(c), &&  c\in [0,\cF],    \label{eq:int3}   
\end{align}   
 
\begin{align}  
 \eF(1)=\eF(\tF+1)=0,  && \label{eq:int4} 
 \end{align}   
 
\begin{align}  
\vF(i-1)\geq \vF(i), && \notag \\
 \cF\cdot (\vF(i-1)-\eF(i)) \geq \chiF(i-1)-\chiF(i) 
 \geq \vF(i-1)- \eF(i), && i\in[2,\tF], \label{eq:int6} 
 \end{align}



\begin{align}  
 \bl(k,i)\geq  \dclrF(\widetilde{\tC} + i)+\chiT(i,k)-1 , ~~~
 ~~~~~   k \in[1,\kC],   i\in[1,\tT], &&    
 \end{align}   
 
\begin{align}  
 \sum_{k \in[1,\kC],  i\in[1,\tT]} \bl(k,i)
 \leq \sum_{i\in[1,\tT]}\dclrF( \widetilde{\tC} +i),   &&    
  \label{eq:int12} 
 \end{align}   
  
  \begin{align}  
 \bl_\LB(k)\leq  \sum_{ i\in[1,\tT]} \bl(k,i) \leq  \bl_\UB(k) , ~~~~~~
     k \in[1,\kC], &&      
       \label{eq:int_last} 
 \end{align}

\begin{align}  
 \tC +\sum_{i\in [1,\tT]} \vT(i) + \sum_{i\in [1,\tF]} \vF(i) =\nint_G.  &&  
  \label{eq:int_last} 
 \end{align}   
 

\subsection{Constraints for Including Fringe-trees} \label{sec:ex}
 
 Recall that   $\mathcal{F}(D_\pi)$ denotes the set of 
chemical rooted trees $\psi$  
r-isomorphic to a chemical rooted tree in $\mathcal{T}(\C)$
  over all chemical graphs $\C\in D_\pi$,
  where possibly a chemical rooted tree $\psi\in \mathcal{F}(D_\pi)$
  consists of a single chemical element $\ta\in \Lambda\setminus \{{\tt H}\}$.

To express the condition that
the ${\rho}$-fringe-tree is chosen from a rooted tree $C_i$, $T_i$  or  $F_i$, 
we introduce the following set of variables and constraints.  
  
\smallskip\noindent
{\bf constants: } 
\begin{enumerate}[leftmargin=*]
\item[-]   $n_\LB$: a lower bound  
on the number $n(\C)$ of non-hydrogen atoms in $\C$,
where $n_\LB, n^*\geq \nint_\LB$;  

\item[-]   $\ch_{\LB}(i),\ch_{\UB}(i)\in [0,n^* ]$, $i\in [1,\tT]$: 
lower and upper bounds on $\h(\langle T_i\rangle)$ of the tree $T_i$ rooted 
at a vertex $\vC_{i}$; 

\item[-]   $\ch_{\LB}(k),\ch_{\UB}(k)\in [0,n^* ]$, $k \in[1,\kC]= \It\cup\Iw$: 
lower and upper bounds on the maximum  
 height $\h(\langle T \rangle)$ of the tree $T\in \F(P_k)$ rooted at 
an internal vertex of a path $P_k$   for an edge $a_k\in \Ew\cup \Et$;  


\item[-]  Prepare a coding of the set  $\mathcal{F}(D_\pi)$ and let 
    $[\psi]$ denote  the coded integer of  
     an element $\psi$ in $\mathcal{F}(D_\pi)$;  

\item[-]   Sets  $\mathcal{F}(v) \subseteq \mathcal{F}(D_\pi), v\in \VC$
and $\mathcal{F}_E \subseteq \mathcal{F}(D_\pi)$ 
 of chemical rooted trees $T$ with $\h(T)\in [1,{\rho}]$;  
 
\item[-]  Define
$\mathcal{F}^*:=\bigcup_{v\in \VC}\mathcal{F}(v)\cup \mathcal{F}_E$, 
 $\FrC_i:= \mathcal{F}(\vC_i)$, $i\in[1,\tC]$,
$\FrT_i:= \mathcal{F}_E$, $i\in[1,\tT]$  and 
$\FrF_i:= \mathcal{F}_E$, $i\in[1,\tF]$;

\item[-]    
 $\fc_\LB(\psi),\fc_\UB(\psi)\in[0,n^*], \psi\in \mathcal{F}^*$:
lower and upper bound functions  
  on the number of   interior-vertices $v$ 
  such that $\C[v]$ is r-isomorphic to $\psi $  in $\C$;
  
\item[-]  
$\FrX_i[p], p\in [1,{\rho}], \mathrm{X}\in\{\mathrm{C,T,F}\}$:
the set of  chemical rooted trees  $T\in  \FrX_i$
with   $\h(\langle T\rangle)= p$;  

\item[-]  
$n_{\oH}([\psi])\in [0, 3^{\rho}], \psi\in \mathcal{F}^*$: 
the number $n(\langle \psi\rangle)$ 
of non-root hydrogen vertices in a chemical rooted tree  $\psi$; 
 
\item[-]  
$\h_{\oH}([\psi])\in [0,{\rho}], \psi\in \mathcal{F}^*$: 
 the height $\h(\langle \psi\rangle)$  of the
 hydrogen-suppressed chemical rooted tree  $\langle \psi\rangle$; 

\item[-]  
$\deg_\mathrm{r}^{\oH}([\psi])\in [0,3], \psi\in \mathcal{F}^*$: 
the number $\deg_\mathrm{r}(\anpsi)$ of non-hydrogen children of the root $r$
 of a chemical rooted tree  $\psi$; 
 
\item[-]  
$\deghyd_\mathrm{r}([\psi])\in [0,3], \psi\in \mathcal{F}^*$: 
the number $\deg_\mathrm{r}(\psi)-\deg_\mathrm{r}(\anpsi)$ 
of hydrogen children of the root $r$ of a chemical rooted tree  $\psi$; 
 
\item[-] 
$\vion(\psi)\in [-3,+3], \psi\in \mathcal{F}^*$: 
  the ion-valence of the root in  $\psi$; 

\item[-] 
  $\ac^\lf_\nu(\psi), \nu\in \Gac^\lf$:
the frequency of leaf-edges with adjacency-configuration $\nu$ in $\psi$;

 \item[-] 
$\ac^\lf_\LB,\ac^\lf_\UB: \Gac^\lf \to  [0,n^*]$:
lower and upper bound functions    on the number of  leaf-edges $uv$ in $\acC$
  with adjacency-configuration $\nu$; 
\end{enumerate}

\smallskip\noindent
{\bf variables: }   
\begin{enumerate}[leftmargin=*]
\item[-]
  $n_G\in [n_\LB, n^*]$: the number $n(\C)$ of non-hydrogen atoms in $\C$;  
\item[-] $\vX(i)\in[0,1], i\in [1,\tX]$,   $\mathrm{X}\in\{\mathrm{T,F}\}$: 
 $\vX(i)=1$ $\Leftrightarrow $ vertex $\vX_{i}$ is used in $\C$; 
       
\item[-] 
$\dlfrX(i,[\psi])\in [0,1], 
    i\in[1,\tX], \psi\in \FrX_i, \mathrm{X}\in \{\mathrm{C,T,F}\}$:  
$\dlfrX(i,[\psi])=1$  $\Leftrightarrow $
 $\psi$ is  the ${\rho}$-fringe-tree rooted at vertex $\vX_i$ in $\C$;  

\item[-]    
$\fc([\psi])\in [\fc_\LB(\psi),\fc_\UB(\psi)], \psi\in \mathcal{F}^*$:
  the number of   interior-vertices $v$ 
  such that $\C[v]$ is r-isomorphic to $\psi$  in $\C$;  
  
\item[-]    
$\ac^\lf([\nu])\in [\ac^\lf_\LB(\nu),\ac^\lf_\UB(\nu)], \nu\in \Gac^\lf$: 
  the number of leaf-edge with adjacency-configuration $\nu$  in $\C$;  
  
\item[-]
 $\degXex(i)\in [0,3],  i\in [1,\tX],     \mathrm{X}\in\{\mathrm{C,T,F}\}$:
the number of non-hydrogen children of the root
 of  the ${\rho}$-fringe-tree rooted at vertex $\vX_i$ in $\C$;  
  
\item[-]  $\hyddegX(i)\in [0,4]$,  $i\in [1,\tX]$, 
 $\mathrm{X}\in \{\mathrm{C,T,F}\}$: 
 the number of  hydrogen atoms adjacent to  vertex $\vX_{i}$
 (i.e.,  $\hyddeg(\vX_{i})$) in $\C=(H,\alpha,\beta)$; 
 
\item[-] 
 $\eledegX(i)\in [-3,+3]$,  $i\in [1,\tX]$, 
 $\mathrm{X}\in \{\mathrm{C,T,F}\}$: 
 the  ion-valence $\vion(\psi)$ of vertex $\vX_{i}$
 (i.e.,  $\eledegX(i)=\vion(\psi)$ 
 for the ${\rho}$-fringe-tree $\psi$ rooted at $\vX_{i}$) in $\C=(H,\alpha,\beta)$; 

\item[-] $\hX(i)\in [0,{\rho}]$, $i\in [1,\tX]$,
$\mathrm{X}\in \{\mathrm{C,T,F}\}$: the height $\h(\langle T\rangle)$ of
the hydrogen-suppressed chemical rooted tree $\langle T\rangle$ of  
 the ${\rho}$-fringe-tree $T$ rooted at vertex $\vX_i$ in $\C$;  
\item[-] $\sigma(k,i)\in[0,1]$, $k \in[1,\kC]=\It\cup\Iw,  i\in [1,\tT]$: 
    $\sigma(k,i)=1$ $\Leftrightarrow$ 
    the ${\rho}$-fringe-tree $T_v$ rooted at  vertex $v=\vT_{i}$ 
      with color $k$  has the largest height $\h(\langle \T_v \rangle)$ among such trees
      $T_v, v\in \VT$;
\end{enumerate}

\smallskip\noindent
{\bf constraints: }    
\begin{align}    
\sum_{\psi\in \FrC_i}\!\!\dlfrC(i,[\psi]) =1, &&  i\in [1,\tC], \notag \\
\sum_{\psi\in \FrX_i }\!\!\dlfrX(i,[\psi]) =\vX(i),  
&&   i\in [1,\tX],  \mathrm{X}\in\{\mathrm{T,F}\},  \label{eq:ex_first}   
\end{align}     
 
\begin{align}    
\sum_{\psi\in \FrX_i }\!\! \deg_\mathrm{r}^{\oH}([\psi]) \cdot \dlfrX(i,[\psi]) 
 = \degXex(i),  &&  \notag \\
\sum_{\psi\in \FrX_i }\!\!   \deghyd_\mathrm{r}([\psi]) \cdot \dlfrX(i,[\psi])
 = \hyddegX(i),   &&  \notag \\
\sum_{\psi\in \FrX_i }\!\! \vion([\psi]) \cdot \dlfrX(i,[\psi]) 
 = \eledegX(i),
&&   i\in [1,\tX],  \mathrm{X}\in\{\mathrm{C,T,F}\},  \label{eq:ex_1}  
\end{align}    


\begin{align}    
\sum_{\psi\in \FrF_i[{\rho}] }\dlfrF(i,[\psi]) 
\geq \vF(i) - \eF(i+1),
      && i\in [1,\tF]~(\eF(\tF+1)=0), \label{eq:ex3}  
\end{align}   
 
\begin{align}   
\sum_{\psi\in \FrX_i } \h_{\oH}([\psi]) \cdot \dlfrX(i,[\psi]) =  \hX(i), && 
   i\in[1,\tX],  \mathrm{X}\in \{\mathrm{C,T,F}\}, \label{eq:ex5}  
\end{align}   

\begin{align}     
\sum\limits_{\substack{  \psi\in \FrX_i \\
              i\in [1,\tX],    \mathrm{X}\in \{\mathrm{C,T,F}\}  }}\!\!
                 n_{\oH}([\psi])  \cdot \dlfrX(i,[\psi])    
 +    \sum_{  i\in [1,\tX], \mathrm{X}\in \{\mathrm{T,F}\} } \vX(i)
   +\tC
  = n_G, ~~
 && 
  \label{eq:ex2} 
\end{align}   

\begin{align} 
\sum_{ i\in [1,\tX], \mathrm{X}\in\{\mathrm{C,T,F}\}} 
\dlfrX(i,[\psi]) =  \fc([\psi]), &&  \psi\in \mathcal{F}^*, 
 \label{eq:ex3} 
\end{align}    

\begin{align}    
\sum_{\psi\in \FrX_i, i\in[1,\tX], \mathrm{X}\in \{\mathrm{C,T,F}\}}
\ac^\lf_\nu(\psi)\cdot \dlfrX(i,[\psi]) = \ac^\lf([\nu]), &&
\nu\in \Gac^\lf, \label{eq:ex4}  
\end{align}

\begin{align}  
\hC(i)    \geq \ch_\LB(i)- n^* \cdot \dclrF(i),  ~~
\clrF(i)+{\rho} \geq \ch_\LB(i) , ~~~~~~~~~~~~~~   &&\notag \\
\hC(i)          \leq \ch_\UB(i) ,  ~~
\clrF(i)+{\rho} \leq \ch_\UB(i)+ n^* \cdot (1-\dclrF(i)),  ~ 
        &&   i\in [1,\widetilde{\tC}],         
            \label{eq:int14} 
 \end{align}   
 
\begin{align}  
 \ch_\LB(i) \leq  \hC(i)   \leq  \ch_\UB(i) ,  ~~ 
        &&    i\in [\widetilde{\tC}+1,\tC],       
             \label{eq:int14} 
 \end{align}   
 
\begin{align}   
 \hT(i)    \leq \ch_\UB(k) 
  + n^*\cdot (\dclrF( \widetilde{\tC}+ i)+1-\chiT(i,k)),  &&\notag \\
\clrF(\widetilde{\tC}+i)+{\rho}
 \leq \ch_\UB(k)+ n^*\cdot (2-\dclrF( \widetilde{\tC}+ i)-\chiT(i,k)),    
&& k \in[1,\kC],  i\in [1,\tT],      
    \label{eq:int15} 
 \end{align}   
 
\begin{align}   
 \sum_{i\in[1,\tT]}\sigma(k,i) =\dclrT(k),   &&   k \in[1,\kC],      
 \label{eq:int16} 
 \end{align}   
 
\begin{align}  
 \chiT(i,k)\geq \sigma(k,i), && \notag\\
 \hT(i)    \geq \ch_\LB(k) - n^*\cdot (\dclrF( \widetilde{\tC}+ i)+1-\sigma(k,i) ),
  && \notag\\ 
\clrF(\widetilde{\tC}+i)+{\rho}
 \geq \ch_\LB(k) - n^* \cdot (2-\dclrF( \widetilde{\tC}+ i)-\sigma(k,i)),   
&&  k \in[1,\kC],  i\in [1,\tT]. 
    \label{eq:ex_last} 
 \end{align}   

\subsection{Descriptor for the  Number of Specified Degree} 
\label{sec:Deg}

We include constraints to compute descriptors for degrees in $\C$. \\

\smallskip\noindent
{\bf variables: } 
\begin{enumerate}[leftmargin=*]
\item[-]  $\degX(i)\in [0,4]$,  $i\in [1,\tX]$, 
 $\mathrm{X}\in \{\mathrm{C,T,F}\}$: 
 the number of non-hydrogen atoms adjacent to  vertex $v=\vX_{i}$
 (i.e.,  $\deg_{\anC}(v)=\deg_H(v)-\hyddeg_{\C}(v)$) in $\C=(H,\alpha,\beta)$; 

\item[-] $\degCT(i)\in [0,4]$,  $i\in [1, \tC]$: the number of edges
from vertex $\vC_{i}$ to vertices $\vT_{j}$, $j\in [1,\tT]$;  
\item[-]  $\degTC(i)\in [0,4]$,  $i\in [1, \tC]$: the number of edges
from  vertices $\vT_{j}$, $j\in [1,\tT]$ to vertex $\vC_{i}$;    
\item[-]    $\ddgC(i,d)\in[0,1]$,  $i\in [1,\tC]$, $d\in [1,4]$, 
  $\ddgX(i,d)\in[0,1]$,  $i\in [1,\tX]$,
 $d\in [0,4]$,  $\mathrm{X}\in \{\mathrm{T,F}\}$: 
        $\ddgX(i,d)=1$ $\Leftrightarrow$   $\degX(i)+\hyddegX(i)=d$;  
       
\item[-]   $\dg(d)\in[\dg_\LB(d),\dg_\UB(d)]$,  $d \in[1,4]$:
    the number  of interior-vertices $v$ with 
       $\mathrm{deg}_H(\vX_{i})=d$   in $\C=(H,\alpha,\beta)$;

\item[-] $\degCint(i)\in [1,4]$,  $i\in [1, \tC]$, 
 $\degXint(i)\in [0,4]$,  $i\in [1, \tX], \mathrm{X}\in \{\mathrm{T,F}\}$: 
the interior-degree $\deg_{H^\inte}(\vX_i)$ 
  in the interior $H^\inte=(V^\inte(\C),E^\inte(\C))$ of  $\C$; i.e., 
the number of interior-edges incident to vertex $\vX_{i}$;

\item[-]    $\ddgCint(i,d)\in[0,1]$,  $i\in [1,\tC]$,  $d\in [1,4]$,  
  $\ddgXint(i,d)\in[0,1]$,  $i\in [1,\tX]$,
 $d\in [0,4]$,  $\mathrm{X}\in \{\mathrm{T,F}\}$: 
       $\ddgXint(i,d)=1$ $\Leftrightarrow$   $\degXint(i)=d$;  
       
\item[-]   $\dg^\inte(d)\in[\dg_\LB(d),\dg_\UB(d)]$,  $d \in[1,4]$:
    the number  of interior-vertices $v$ with
    the interior-degree  $\deg_{H^\inte}(v)=d$
  in the interior $H^\inte=(V^\inte(\C),E^\inte(\C))$ of  $\C=(H,\alpha,\beta)$.
  
\end{enumerate}
   
\smallskip\noindent
{\bf constraints: }   
\begin{align}   
\sum_{   k\in \It^+(i)\cup \Iw^+(i)} \dclrT(k) = \degCT(i), ~~
 \sum_{   k\in \It^-(i)\cup \Iw^-(i)} \dclrT(k) = \degTC(i), 
    &&    i\in [1, \tC],     \label{eq:Deg_first}  
\end{align}

\begin{align}   
\tldgC^-(i)+\tldgC^+(i)   + \degCT(i)  + \degTC(i) + \dclrF(i) = \degCint(i),  
    &&    i\in [1, \widetilde{\tC}],     \label{eq:Deg2}  
\end{align}   

\begin{align}      
\tldgC^-(i)+\tldgC^+(i)  + \degCT(i)  + \degTC(i)   = \degCint(i),  
     &&   i\in [\widetilde{\tC}+1,\tC],     \label{eq:Deg2b}  
\end{align}   

\begin{align}      
  \degCint(i)+ \degCex(i) = \degC(i),  
    &&    i\in [1, \tC],     \label{eq:Deg2c}  
\end{align}   

\begin{align}    
\sum_{\psi\in \FrC_i[{\rho}] }\dlfrC(i,[\psi]) \geq 2-\degCint(i)
      &&  i\in [1, \tC],    \label{eq:Deg2d} 
\end{align}   

\begin{align}   
  2\vT(i)   + \dclrF(\widetilde{\tC}+i)    =\degTint(i),   && \notag \\
 \degTint(i)+ \degTex(i)  =\degT(i),   && 
  i\in [1,\tT]~(\eT(1)=\eT(\tT+1)=0), \label{eq:Deg3}  
\end{align}   

\begin{align}
   \vF(i) +\eF(i+1)  =\degFint(i),  && \notag \\
   \degFint(i)  +\degFex(i)   = \degF(i),   && 
  i\in [1,\tF] ~(\eF(1)=\eF(\tF+1)=0),   \label{eq:Deg4}  
\end{align} 

\begin{align}   
\sum_{d\in [0,4]}\ddgX(i,d)=1, ~
\sum_{d\in [1,4]}d\cdot\ddgX(i,d)=\degX(i)+\hyddegX(i), && \notag \\
\sum_{d\in [0,4]}\ddgXint(i,d)=1, ~
\sum_{d\in [1,4]}d\cdot\ddgXint(i,d)=\degXint(i), && 
 i\in [1,\tX],  \mathrm{X}\in \{\mathrm{T, C, F}\}, \label{eq:Deg5}  
\end{align}   
 
\begin{align}   
\sum_{ i\in [1,\tC]} \ddgC(i,d) + \sum_{ i\in [1,\tT]} \ddgT(i,d) 
 + \sum_{ i\in [1,\tF] }  \ddgF(i,d) = \dg(d),   &&   \notag \\
\sum_{ i\in [1,\tC]} \ddgCint(i,d) + \sum_{ i\in [1,\tT]} \ddgTint(i,d) 
 + \sum_{ i\in [1,\tF] }  \ddgFint(i,d) = \dg^\inte(d),     
    && d\in [1,4].  
  \label{eq:Deg_last}  
\end{align}   

\subsection{Assigning Multiplicity} 
\label{sec:beta}

 We prepare an integer variable $\beta(e)$  
 for each edge $e$ in the scheme graph $\mathrm{SG}$ 
 to denote the bond-multiplicity of $e$ in a selected graph $H$ and
 include necessary constraints for the variables to satisfy in $H$. 
 
\smallskip\noindent
{\bf constants: }
\begin{enumerate}[leftmargin=*]
\item[-]
$\betar([\psi])$: the sum $\beta_\psi(r)$ of bond-multiplicities of edges
incident to  the root $r$ of a chemical rooted tree $\psi\in \mathcal{F}^*$; 
\end{enumerate}

\smallskip\noindent
{\bf variables: } 
\begin{enumerate}[leftmargin=*]
\item[-] $\bX(i)\in [0,3]$,   $i\in [2,\tX]$, $\mathrm{X}\in \{\mathrm{T,F}\}$:   
 the bond-multiplicity of edge  $\eX_{i}$ in $\C$;  
 
\item[-] $\bC(i)\in [0,3]$,     $i\in [\widetilde{\kC}+1,\mC]= \Iw\cup \Iz\cup\Iew$:    
     the bond-multiplicity of 
     edge  $a_{i}\in \Ew\cup \Ez\cup\Eew$ in $\C$;        
\item[-]
   $\bCT(k), \bTC(k)\in [0,3]$, $k\in [1, \kC]=\It\cup \Iw$: 
   the bond-multiplicity of the first (resp., last) edge of the pure path $P_k$ in $\C$;    
   
\item[-]
   $\bsF(c)\in [0,3], c\in [1,\cF=\widetilde{\tC} + \tT ]$:  
   the bond-multiplicity of the first edge of the leaf path $Q_c$
   rooted at vertex $\vC_{c}, c\leq\widetilde{\tC} $
    or $\vT_{c-\widetilde{\tC}}, c>\widetilde{\tC} $  in $\C$;   
    
\item[-] $\bXex(i)\in [0,4],  i\in [1,\tX],   \mathrm{X}\in\{\mathrm{C,T,F}\}$:
the sum $\beta_{\C[v]}(v)$ of bond-multiplicities of edges in the ${\rho}$-fringe-tree
$\C[v]$ rooted at  interior-vertex $v=\vX_{i}$;  

\item[-] $\delbX(i,m)\in [0,1]$, $i\in [2,\tX]$,   $m\in[0,3]$, 
       $\mathrm{X}\in \{\mathrm{T,F}\}$:  
  $\delbX(i,m)=1$  $\Leftrightarrow$  $\bX(i)=m$; 
\item[-] $\delbC(i,m)\in [0,1]$,  
   $i\in [\widetilde{\kC},\mC]=\Iw\cup \Iz\cup\Iew$,  $m\in[0,3]$:  
   $\delbC(i,m)=1$  $\Leftrightarrow$  $\bC(i)=m$; 
\item[-]
   $\delbCT(k,m), \delbTC(k,m)\in [0,1]$, $k\in [1, \kC]=\It\cup \Iw$,  $m\in[0,3]$:
     $\delbCT(k,m)=1$   (resp., $\delbTC(k,m)=1$)    $\Leftrightarrow$  
           $\bCT(k)=m$ (resp., $\bTC(k)=m$); 
           
\item[-]
   $\delbsF(c,m)\in [0,1]$, $c\in [1,\cF]$,  
    $m\in[0,3], \mathrm{X}\in \{\mathrm{C,T}\}$: 
     $\delbsF(c,m)=1$ $\Leftrightarrow$  $\bsF(c)=m$;  
\item[-] $\bd^\inte(m)\in[0, 2\nint_\UB]$, $m\in[1,3]$:
      the number of interior-edges with bond-multiplicity  $m$ in $\C$;  
      
\item[-] $\bdX(m)\in [0,2\nint_\UB],  \mathrm{X}\in \{\mathrm{C,T,CT,TC}\}$,
      $\bdX(m)\in [0,2\nint_\UB], \mathrm{X}\in \{\mathrm{F,CF,TF}\}$, $m\in[1,3]$:  
 the number of interior-edges $e\in \EX$ with bond-multiplicity  $m$ in  $\C$; 
\end{enumerate}
 
\smallskip\noindent
{\bf constraints: } 
\begin{align}    
\eC(i)\leq \bC(i)\leq 3\eC(i), 
  i\in [\widetilde{\kC}+1,\mC]=\Iw\cup \Iz\cup\Iew, \label{eq:beta_first} 
\end{align}   

\begin{align}   
  \eX(i)\leq \bX(i)\leq 3 \eX(i), 
  &&    i\in [2,\tX],   \mathrm{X}\in \{\mathrm{T, F}\},    \label{eq:beta1}  
\end{align}   

\begin{align}   
\dclrT(k)\leq \bCT(k)\leq 3 \dclrT(k), ~~~ 
\dclrT(k)\leq \bTC(k)\leq 3 \dclrT(k), &&  k\in [1, \kC], \label{eq:beta8} \\ 
\dclrF(c)\leq \bXF(c)\leq 3 \dclrF(c), &&   c\in [1,\cF], \label{eq:beta8}  
\end{align}

\begin{align} 
\sum_{m\in[0,3]} \delbX(i,m)=1,  ~~
\sum_{m\in[0,3]}m\cdot \delbX(i,m)=\bX(i), &&   i\in [2,\tX],  
  \mathrm{X}\in \{\mathrm{T,F}\},  \label{eq:beta10}    
\end{align}   

\begin{align} 
\sum_{m\in[0,3]} \delbC(i,m)=1,  ~~
\sum_{m\in[0,3]}m\cdot \delbC(i,m)=\bC(i), &&   i\in [\widetilde{\kC}+1,\mC],   \label{eq:beta11}    
\end{align}   
 
\begin{align}   
\sum_{m\in[0,3]} \delbCT(k,m)=1,    ~~ 
\sum_{m\in[0,3]}m\cdot\delbCT(k,m)=\bCT(k),
&&     k\in [1, \kC],  \notag \\      
\sum_{m\in[0,3]} \delbTC(k,m)=1,  ~~ 
\sum_{m\in[0,3]} m\cdot\delbTC(k,m)=\bTC(k),  &&
      k\in [1, \kC], \notag \\
\sum_{m\in[0,3]} \delbsF(c,m)=1,  ~~ 
\sum_{m\in[0,3]} m\cdot\delbsF(c,m)=\bsF(c),  &&    c\in [1,\cF],
   \label{eq:beta15}    
\end{align}    

 \begin{align}           
\sum_{\psi\in \FrX_i } \betar([\psi]) \cdot  \dlfrX(i,[\psi]) = \bXex(i),     
 &&  i\in [1,\tX],     \mathrm{X}\in\{\mathrm{C,T,F}\}, 
 \label{eq:beta16a}   
\end{align}

\begin{align} 
 \sum_{i\in [\widetilde{\kC}+1,\mC]} \delbC(i,m) =\bdC(m), ~~
  \sum_{i\in [2,\tT]} \delbT(i,m)    =\bdT(m),   \notag \\ 
   \sum_{k\in [1, \kC]}\delbCT(k,m)=\bdCT(m), ~~
   \sum_{k\in [1, \kC]}\delbTC(k,m)=\bdTC(m),    \notag \\ 
\sum_{i\in [2,\tF]}\!\!\! \delbF(i,m) =\bdF(m), ~~
 \sum_{c\in [1,\widetilde{\tC}]} \delbsF(c,m)  =\bdCF(m),   \notag \\ 
  \sum_{c\in [\widetilde{\tC}+1,\cF]}  \delbsF(c,m) =\bdTF(m),  
   \notag \\ 
 \bdC(m)+\bdT(m) + \bdF(m)
 +\bdCT(m)+\bdTC(m) +\bdTF(m)+\bdCF(m) = \bd^\inte(m),   \notag \\ 
  m\in [1,3].       \label{eq:beta_last} 
\end{align}

\subsection{Assigning Chemical Elements and  Valence Condition}
\label{sec:alpha} 

We include constraints so that each vertex $v$ in a selected graph $H$
satisfies the valence condition; i.e., 
$\beta_\C(v)=  \val(\alpha(v)) +\eledeg_\C(v)$, 
where $\eledeg_\C(v)=\vion(\psi)$ for the ${\rho}$-fringe-tree $\C[v]$
r-isomorphic to $\psi$. 
With these constraints, a chemical graph
   $\C=(H,\alpha,\beta)$ on a selected subgraph $H$
will be constructed. 
 
\smallskip\noindent
{\bf constants: }
 \begin{enumerate}[leftmargin=*]
\item[-] Subsets
 $\Lambda^\inte \subseteq \Lambda\setminus\{{\tt H}\}, 
 \Lambda^\ex \subseteq \Lambda$ of chemical elements,
 where we denote by $[{\tt e}]$ (resp., $[{\tt e}]^\inte$ and $[{\tt e}]^\ex$)  
 of a standard encoding of an element ${\tt e}$ in the set $\Lambda$ 
 (resp.,    $\Lambda^\inte_\epsilon$ and  $\Lambda^\ex_\epsilon$);  
\item[-]  A valence function: $\val: \Lambda \to [1,6]$;  

\item[-]  A function $\mathrm{mass}^*:\Lambda\to \mathbb{Z}$ 
(we let $\mathrm{mass}(\ta)$ denote  the observed mass of a chemical element  
$\ta\in \Lambda$, and define 
   $\mathrm{mass}^*(\ta)\triangleq
    \lfloor 10\cdot \mathrm{mass}(\ta)\rfloor$);  
        
\item[-]  
 Subsets $\Lambda^*(i)\subseteq \Lambda^\inte$, $i\in[1,\tC]$; 
 
\item[-] 
 $\na_\LB(\ta),\na_\UB(\ta)\in [0,n^* ]$,  $\ta\in  \Lambda$:
lower and upper bounds on the number of vertices  $v$    with $\alpha(v)=\ta$;  
\item[-] 
  $\na_\LB^\inte(\ta),\na_\UB^\inte(\ta)\in [0,n^* ]$,
 $\ta\in  \Lambda^\inte$:
lower and upper bounds on the number  of interior-vertices  
 $v$ with $\alpha(v)=\ta$; 

\item[-] 
$\alpha_\mathrm{r}([\psi])\in [\Lambda^\ex], \in \mathcal{F}^*$:
 the chemical element $\alpha(r)$  of the root $r$ of  $\psi$;

\item[-]   $\na_\ta^\ex([\psi])\in [0,n^*]$,  
$\ta\in \Lambda^\ex, \psi\in \mathcal{F}^*$: 
the frequency of chemical element $\ta$ in the set of  
non-rooted vertices   in   $\psi$, where possibly $\ta={\tt H}$;


\item[-]   $\mathrm{M}$: 
an upper bound for the average $\overline{\mathrm{ms}}(\C)$ of mass$^*$ 
over all atoms in $\C$;

\end{enumerate}
      
\smallskip\noindent
{\bf variables: } 
\begin{enumerate}[leftmargin=*]
\item[-]
   $\bCT(i),\bTC(i)\in [0,3], i\in [1,\tT]$:
the bond-multiplicity of edge $\eCT_{j,i}$ (resp., $\eTC_{j,i}$)
if one exists;  

\item[-] 
 $\bCF(i), \bTF(i)\in [0,3], i\in [1,\tF]$:
the bond-multiplicity of $\eCF_{j,i}$ (resp., $\eTF_{j,i}$)
if one exists;  

\item[-]  $\aX(i)\in [\Lambda^\inte_\epsilon ],
       \delaX(i,[\ta]^\inte)\in [0,1],  \ta\in \Lambda^\inte_\epsilon, i\in [1,\tX],
         \mathrm{X}\in \{\mathrm{C,T,F}\}$:  
$\aX(i)= [\ta]^\inte\geq 1$  (resp., $\aX(i)=0$)
  $\Leftrightarrow$ $\delaX(i,[\ta]^\inte)=1$ (resp., $\delaX(i,0)=0$)  
  $\Leftrightarrow$ $\alpha(\vX_{i})= \ta\in \Lambda$ 
(resp., vertex $\vX_{i}$ is not used in $\C$); 

\item[-] 
$\delaX(i,[\ta]^\inte)\in [0,1], i\in [1,\tX],  
\ta  \in \Lambda^\inte,   \mathrm{X}\in\{\mathrm{C,T,F}\}$:   
   $\delaX(i,[\ta]^\typ)=1$   $\Leftrightarrow$   $\alpha(\vX_{i})=\ta$;  
\item[-]  $\mathrm{Mass}\in \mathbb{Z}_+$: 
 $\sum_{v\in V(H)} \mathrm{mass}^*(\alpha(v))$;  
 
\item[-]  $\overline{\mathrm{ms}}\in \mathbb{R}_+$: 
 $\sum_{v\in V(H)} \mathrm{mass}^*(\alpha(v)) / |V(H)|$;  
 \item[-] 
$\delta_{\mathrm{atm}}(i)\in [0,1], i\in [n_\LB + \na_\LB({\tt H}), 
n^* + \na_\UB({\tt H})]$:   
   $\delta_{\mathrm{atm}}(i)=1$   $\Leftrightarrow$   $|V(H)| = i$;  

\item[-]   $\na([\ta])\in[\na_\LB(\ta),\na_\UB(\ta)]$,
 $\ta \in \Lambda$:
    the number  of vertices $v\in V(H)$
     with $\alpha(v)=\ta$, where possibly $\ta={\tt H}$; 
    
\item[-]   $\na^{\inte}([\ta]^\inte) \in[\na_\LB^\inte(\ta),\na_\UB^\inte(\ta)]$,
 $\ta \in \Lambda, \mathrm{X}\in \{\mathrm{C,T,F}\}$:
    the number  of interior-vertices  $v\in V(\C)$ 
    with $\alpha(v)=\ta$; 
    
\item[-]   $\naX^\ex([\ta]^\ex) , \na ^\ex([\ta]^\ex) \in [0,\na_\UB(\ta)]$,
 $\ta \in \Lambda$,   $\mathrm{X}\in \{\mathrm{C,T,F}\}$: 
    the number    of   exterior-vertices rooted at vertices $v\in\VX$ 
  and  the number    of   exterior-vertices $v$
     such that  $\alpha(v)=\ta$;
     

\end{enumerate} 
    
\smallskip\noindent
{\bf constraints: } 
\begin{align}    
 \bCT(k)-3(\eT(i)-\chiT(i,k)+1) \leq 
\bCT(i)\leq \bCT(k)+3(\eT(i)-\chiT(i,k)+1),  i\in [1,\tT], && \notag\\   
  \bTC(k)-3(\eT(i+1)-\chiT(i,k)+1) \leq 
\bTC(i)\leq \bTC(k)+3(\eT(i+1)-\chiT(i,k)+1),  i\in [1,\tT], && \notag\\ 
   k\in [1, \kC],   &&  \label{eq:alpha_first}  
\end{align}

\begin{align}   
 \bsF(c)-3(\eF(i)-\chiF(i,c)+1) \leq 
\bCF(i)\leq \bsF(c)+3(\eF(i)-\chiF(i,c)+1),   i\in [1,\tF], 
&&  c\in[1,\widetilde{\tC}] ,    \notag\\  
  \bsF(c)-3(\eF(i)-\chiF(i,c)+1) \leq 
\bTF(i)\leq \bsF(c)+3(\eF(i)-\chiF(i,c)+1),    i\in [1,\tF], 
&& c\in[\widetilde{\tC}+1,\cF] ,    \notag\\  
  \label{eq:alpha2}  
\end{align}

\begin{align}  
   \sum_{\ta\in \Lambda^\inte} \delaC(i,[\ta]^\inte)=1, ~~ 
   \sum_{\ta\in \Lambda^\inte} [\ta]^\inte\cdot\delaX(i,[\ta]^\inte)=\aC(i),   
     &&  i\in [1,\tC], \notag  \\
  \sum_{\ta\in \Lambda^\inte } \delaX(i,[\ta]^\inte)=\vX(i), ~~ 
   \sum_{\ta\in \Lambda^\inte} [\ta]^\inte\cdot\delaX(i,[\ta]^\inte)=\aX(i), 
    &&  i\in [1,\tX],  \mathrm{X}\in \{\mathrm{T,F}\},  
  \label{eq:alpha_first} 
\end{align}

\begin{align}  
\sum_{\psi\in \FrX_i } 
 \alpha_\mathrm{r}([\psi])\cdot \dlfrX(i,[\psi]) = \aX(i),   
 &&   i\in [1,\tX],   \mathrm{X}\in \{\mathrm{C,T,F}\},  
  \label{eq:alpha_1} 
\end{align}    
  

\begin{align}  
\sum_{j\in \IC(i)}\bC(j)  
+ \sum_{  k\in \It^+(i)\cup \Iw^+(i)} \bCT(k)
+ \sum_{  k\in \It^-(i)\cup \Iw^-(i)} \bTC(k)   &&  \notag\\
     + \bsF(i)  +\bCex(i) -\eledegC(i) 
     =
     \sum_{\ta\in \Lambda^\inte}\val(\ta)\delaC(i,[\ta]^\inte),  
 &&  i\in [1,\widetilde{\tC}],  \label{eq:alpha3} 
\end{align}

\begin{align}   
\sum_{j\in \IC(i)}\bC(j)  
+ \sum_{  k\in \It^+(i)\cup \Iw^+(i)} \bCT(k)
+ \sum_{  k\in \It^-(i)\cup \Iw^-(i)} \bTC(k)   &&  \notag\\
+\bCex(i) -\eledegC(i) 
   =
      \sum_{\ta\in \Lambda^\inte}\val(\ta)\delaC(i,[\ta]^\inte),  
  &&  i\in [\widetilde{\tC}+1,\tC],   \label{eq:alpha3b} 
\end{align} 

\begin{align}  
 \bT(i)+\bT(i\!+\!1)   +   \bTex(i)  
  + \bCT(i) + \bTC(i) \hspace{1cm} \notag\\
  + \bsF(\widetilde{\tC}+i) -\eledegT(i) 
   =
      \sum_{\ta\in \Lambda^\inte}\val(\ta)\delaT(i,[\ta]^\inte), 
   \notag \\
  i\in [1,\tT]~  (\bT(1)=\bT(\tT+1)=0),   \label{eq:alpha4} 
\end{align}
 
\begin{align} 
 \bF(i)+\bF(i\!+\!1) +\bCF(i) +\bTF(i)  \hspace{1cm}    \notag\\
  +\bFex(i)  -\eledegF(i) 
   =
      \sum_{\ta\in \Lambda^\inte}\val(\ta)\delaF(i,[\ta]^\inte),  
   \notag \\
  i\in [1,\tF] ~  (\bF(1)=\bF(\tF+1)=0),   \label{eq:alpha5} 
\end{align}

\begin{align}  
 \sum_{i\in [1,\tX] } \delaX(i,[\ta]^\inte) = \naX([\ta]^\inte) ,  
 &&  \ta\in \Lambda^\inte, \mathrm{X}\in \{\mathrm{C,T,F}\},  
     \label{eq:alpha6} 
\end{align}    

\begin{align}   
\sum_{\psi\in \FrX_i , i\in [1,\tX] } \na_\ta^\ex([\psi])\cdot  \dlfrX(i,[\psi])  
   = \naX^\ex([\ta]^\ex),     && 
 \ta\in \Lambda^\ex,  \mathrm{X}\in \{\mathrm{C,T,F}\},  \label{eq:alpha6} 
\end{align}    
       
\begin{align}  
 \naC([\ta]^\inte)+ \naT([\ta]^\inte)+ \naF([\ta]^\inte) =  \na^\inte([\ta]^\inte),  
    &&   \ta\in \Lambda^\inte,       \notag \\
  \sum_{  \mathrm{X}\in \{\mathrm{C,T,F}\} }
   \naX^\ex([\ta]^\ex)       =\na^\ex([\ta]^\ex),  
  &&  \ta\in \Lambda^\ex,    \notag \\
  \na^\inte([\ta]^\inte) + \na^\ex([\ta]^\ex)=\na([\ta]),  
 &&   \ta\in \Lambda^\inte\cap \Lambda^\ex,     \notag \\
  \na^\inte([\ta]^\inte)  =\na([\ta]),  
 &&      \ta\in \Lambda^\inte \setminus \Lambda^\ex,     \notag \\ 
   \na^\ex([\ta]^\ex)  =\na([\ta]),  
 &&      \ta\in \Lambda^\ex \setminus \Lambda^\inte,        
     \label{eq:alpha6} 
\end{align}

 \begin{align}    
 \sum_{\ta\in \Lambda^*(i)} \delaC(i,[\ta]^\inte) = 1,  
  &&  i\in [1,\tC],   \label{eq:alpha8} 
\end{align}

\begin{align}   
\sum_{ \ta\in\Lambda }\mathrm{mass}^*(\ta )\cdot \na([\ta])
 =\mathrm{Mass}, &&    \label{eq:alpha7} 
\end{align}  

 \begin{align}
 \sum_{i \in [n_\LB + \na_\LB({\tt H}), n^* + \na_\UB({\tt H})]} \delta_{\mathrm{atm}}(i) = 1, && \\
 \sum_{i \in [n_\LB + \na_\LB({\tt H}), n^* + \na_\UB({\tt H})]} i \cdot \delta_{\mathrm{atm}}(i) 
 = n_G + \na^\ex([{\tt H}]^\ex), &&  \\
 \mathrm{Mass} / i - \mathrm{M}\cdot (1 - \delta_{\mathrm{atm}}(i)) 
 \le
 \overline{\mathrm{ms}} 
 \le \mathrm{Mass} / i+ \mathrm{M}\cdot (1 - \delta_{\mathrm{atm}}(i)), 
   && i \in [n_\LB + \na_\LB({\tt H}), n^* + \na_\UB({\tt H})]. 
  \label{eq:alpha_last} 
 \end{align}

\subsection{Constraints for Bounds on the Number of Bonds}  
\label{sec:BDbond}

We include constraints for specification of lower and upper bounds
$\bd_\LB$ and $\bd_\UB$. 

\smallskip\noindent
{\bf constants: } 
\begin{enumerate}[leftmargin=*]
\item[-]
$\bd_{m, \LB}(i), \bd_{m, \UB}(i)\in [0,\nint_\UB]$, 
$i\in [1,\mC]$,  $m\in [2,3]$:  lower and upper bounds 
 on the number  of edges $e\in E(P_i)$ with bond-multiplicity $\beta(e)=m$
 in the pure path $P_i$ for edge $e_i\in \EC$; 
\end{enumerate}

\smallskip\noindent
{\bf variables : } 
\begin{enumerate}[leftmargin=*]
\item[-]
  $\bdT(k,i,m)\in [0,1]$, $k\in [1, \kC]$, $i\in [2,\tT]$, $m\in [2,3]$:  
  $\bdT(k,i,m)=1$  $\Leftrightarrow$ the pure path $P_k$ for edge $e_k\in \EC$ 
  contains edge $\eT_i$ with $\beta(\eT_i)=m$; 
\end{enumerate}
  
\smallskip\noindent
{\bf constraints: } 
\begin{align}    
\bd_{m,\LB}(i)\leq \delbC(i,m)\leq \bd_{m,\UB}(i), 
  i\in \Iew\cup \Iz, m\in [2,3], && 
  \label{eq:BDbond_first}  
\end{align}

\begin{align}   
\bdT(k,i,m)\geq \delbT(i,m)+\chiT(i,k)-1, 
~~~ k \in [1, \kC], i\in [2,\tT],  m\in [2,3], && 
 \label{eq:BDbond2}  
\end{align}   
 
\begin{align}    
\sum_{j\in[2,\tT]}\delbT(j,m) \geq 
\sum_{k\in[1, \kC], i\in [2,\tT]}\!\!\!\! \bdT(k,i,m) , 
~~ m\in [2,3],  \label{eq:BDbond3}  
\end{align}    

\begin{align}    
 \bd_{m, \LB}(k) \leq 
   \sum_{i\in [2,\tT]}\bdT(k,i,m) +\delbCT(k,m)+\delbTC(k,m)    
   \leq   \bd_{m, \UB}(k), ~~~~  \notag \\
    k\in [1, \kC],   m\in [2,3]. ~~ 
     \label{eq:BDbond_last}  
\end{align}

\subsection{Constraints for \GNN}  
\label{sec:GNN_constraint}
 
Recall that the node feature vector $\theta_v^{(0)}$ for each interior-vertex $v=\vX_i, i\in[1,\tX], \textrm{X} \in \{ \textrm{C,T,F} \}$ consists of the following $\zmax=15$ entries,
where a $\rho$-fringe-tree $\psi$ is encoded into a $\zf(=8)$-dimension real vector.

\begin{itemize}
\item[-] $\delaX(i, [\ttC]^\inte ) \in [0,1], i\in[1,\tX], \textrm{X} \in \{ \textrm{C,T,F} \}$; 
\item[-] $\delaX(i, [\ttO]^\inte ) \in [0,1], i\in[1,\tX], \textrm{X} \in \{ \textrm{C,T,F} \}$; 
\item[-] $\delaX(i, [\ttN]^\inte ) \in [0,1], i\in[1,\tX], \textrm{X} \in \{ \textrm{C,T,F} \}$; 
\item[-] $\dgX(i) + \hyddegX(i), i\in[1,\tX], \textrm{X} \in \{ \textrm{C,T,F} \}$; 
\item[-] $\sum_{\ta \in \Lambda^\inte} \val(\ta) \delaX(i, [\ta]^\inte) + \eledegX(i) , i\in[1,\tX], \textrm{X} \in \{ \textrm{C,T,F} \}$
\item[-] $\hyddegX(i), i\in[1,\tX], \textrm{X} \in \{ \textrm{C,T,F} \}$; 
\item[-] $\eledegX(i), i\in[1,\tX], \textrm{X} \in \{ \textrm{C,T,F} \}$; 
\item[-] $\theta_\psi(\cdot, [\psi]) \in \mathbb{R}^\zf$ : the encoded feature vector of the $\rho$-fringe-tree $\psi$.
\end{itemize}

\smallskip\noindent
{\bf constants: } 

$\kappa = 0.1$, the parameter used for the LeakyReLU activation function;

Let $I_a^+(i) \triangleq \Iw^+(i) \cup \Iz^+(i) \cup \Iew^+(i)$ and $I_a^-(i) \triangleq \Iw^-(i) \cup \Iz^-(i) \cup \Iew^-(i)$ ;

Let $I_b^+(i) \triangleq \It^+(i) \cup \Iw^+(i) $ and $I_b^-(i) \triangleq \It^-(i) \cup \Iw^-(i) $ ;

Let $N_a^+(i)$ denote the set of indices $i$ of the tails $\vC_j$ of edges in $(\vC_j, \vC_i) \in \Ew \cup \Ez \cup \Eew$ and $N_a^-(i)$ denote the set of indices $i$ of the heads $\vC_j$ of edges $(\vC_i, \vC_j) \in \Ew \cup \Ez \cup \Eew$;

$\zmax, \phid, \llmax, \pcnv \in \Z_+$;

$\w_0(i,j) \in \R, i \in[1,\zmax], j \in [1,\phid], \ell \in [1,\llmax -1 ]$;

$\wll(i,j) \in \R, i,j \in [1,\phid], \ell \in [1,\llmax -1 ]$;

$\wftr(i,j) \in \R, i\in[1,\phid], j\in[1,\pcnv]$;

$\bias(z;\ell) \in \R, z \in[1,\phid], \ell \in [0,\llmax - 1]$;

$\Mll \in \R_+, \ell \in [0,\llmax]$: An upper bound on the maximum value $\theta_{\max}(z,\ell)$ of entry $z$ in the $\ell$-th layer over all training data $D_\pi$.

\smallskip\noindent
{\bf variables: } 

(In order to improve the readability, we will sometimes use $\theta(v,z)$ instead of $\theta_v(z)$.)

$\thftr(p) \in \R, p\in[1,\pcnv]$: the $p$-th entry of a representation vector $\thftr \in \R^\pcnv$;
$\tau_{\bbC}(p) \in \R, \delta^\tau_{\bbC}(p) \in \{ 0, 1 \} , p\in[1,\pcnv]$;

$\thC(0;z;\ell) \in \R, i\in[1,\tC], z \in[1,\zmax]$: the $z$-th entry of vector $\theta^{(\ell)}_v$ of vertex $v=\vC_i$ in the 0-th layer;

$\thT(0;z;\ell) \in \R, i\in[0,\tT+1], z \in[1,\zmax]$: the $z$-th entry of vector $\theta^{(\ell)}_v$ of vertex $v=\vT_i$ in the 0-th layer, where $\vT_i$ may be not used in a target chemical graph;

$\thF(0;z;\ell) \in \R, i\in[0,\tF+1], z \in[1,\zmax]$: the $z$-th entry of vector $\theta^{(\ell)}_v$ of vertex $v=\vF_i$ in the 0-th layer, where $\vF_i$ may be not used in a target chemical graph;

$\thC(i;z;\ell) \in \R, i\in[1,\tC], z \in[1,\phid], \ell\in[1,\llmax]$: the $z$-th entry of vector $\theta^{(\ell)}_v$ of vertex $v=\vC_i$ in the $\ell$-th layer;
$\tauC(i;z;\ell) \in \R, \deltauC(i;z;\ell) \in \{0, 1 \}, i\in[1,\tC], z \in[1,\phid], \ell\in[1,\llmax]$;

$\thT(i;z;\ell) \in \R, i\in[0,\tT+1], z \in[1,\phid], \ell\in[1,\llmax]$: the $z$-th entry of vector $\theta^{(\ell)}_v$ of vertex $v=\vT_i$ in the $\ell$-th layer, where $\vT_i$ may be not used in a target chemical graph;
$\tauT(i;z;\ell) \in \R, \deltauT(i;z;\ell) \in \{0, 1 \}, i\in[0, \tT + 1], z \in[1,\phid], \ell\in[1,\llmax]$;

$\thF(i;z;\ell) \in \R, i\in[0,\tF+1], z \in[1,\phid], \ell\in[1,\llmax]$: the $z$-th entry of vector $\theta^{(\ell)}_v$ of vertex $v=\vF_i$ in the $\ell$-th layer, where $\vF_i$ may be not used in a target chemical graph;
$\tauF(i;z;\ell) \in \R, \deltauF(i;z;\ell) \in \{0, 1 \}, i\in[0, \tF + 1], z \in[1,\phid], \ell\in[1,\llmax]$;

$\thC_-(k;z;\ell), \thC_+(k;z;\ell)\in\R, k\in[\widetilde{\kC}+1,\mC], z\in[1,\phid],\ell\in[0,\llmax-1]$: the $z$-th entry of vector $\theta^{(\ell)}_v$ of the head and tail $v$ of edge $a_k \in \Ew \cup \Ez \cup \Eew$, where $a_k \in \Ew \cup \Ez $ may be not used in a target chemical graph;

$\thT_-(i;z;\ell), \thT_+(i;z;\ell)\in\R, i\in[1,\tT+1],z\in[1,\phid],\ell\in[0,\llmax-1]$: the value to \\
 $\sum_{z' \in[1,\zmax]}\wll(z',z)\theta^{(\ell)}(\vT_i,z'))$ for edge $e=\eT_i$ (resp., edge $e=\eT_{i+1}$), where such an edge $e\in \ET$ may be not used in a target chemical graph;

$\thF_-(i;z;\ell), \thF_+(i;z;\ell)\in\R, i\in[1,\tF+1],z\in[1,\phid],\ell\in[0,\llmax-1]$: the value to \\
 $\sum_{z' \in[1,\zmax]}\wll(z',z)\theta^{(\ell)}(\vF_i,z'))$ for edge $e=\eF_i$ (resp., edge $e=\eF_{i+1}$), where such an edge $e\in \EF$ may be not used in a target chemical graph;

$\thCTT(k;z;\ell), \thTCT(k;z;\ell)\in\R, k\in[1,\kC], z\in[1,\phid],\ell\in[0,\llmax-1]$: the value to \\
$\sum_{z'\in[1,\zmax]}\wll(z',z)\theta^{(\ell)}(v,z')$ for the edge $(u=\vC_j, v=\vT_i)\in\ECT$ (resp., $(v=\vT_i, u=\vC_j) \in\ETC$ and $(v=\vT_i, u=\vF_j) \in\ETF$), where such an edge $uv$ may be not used in a target chemical graph;

$\thCTC(i;z;\ell), \thTCC(i;z;\ell), \thTFF(i;z;\ell)\in\R, i\in[1,\tT], z\in[1,\phid],\ell\in[0,\llmax-1]$: the value to \\
$\sum_{z'\in[1,\zmax]}\wll(z',z)\theta^{(\ell)}(u,z')$ for the edge $(u=\vC_j, v=\vT_i)\in\ECT$ (resp., $(v=\vT_i, u=\vC_j) \in\ETC$ and $(v=\vT_i, u=\vF_j) \in\ETF$), where such an edge $uv$ may be not used in a target chemical graph;

$\thCFF(c;z;\ell)\in\R, c\in[1,\widetilde{\tC}], z\in[1,\phid],\ell\in[0,\llmax-1]$: the value to \\
$\sum_{z'\in[1,\zmax]}\wll(z',z)\theta^{(\ell)}(u,z')$ for the edge $(v=\vC_c, u=\vF_i)\in\ECF$, where such an edge $uv$ may be not used in a target chemical graph;

$\thCFC(i;z;\ell), \thTFT(i;z;\ell)\in\R, i\in[1,\tF], z\in[1,\phid],\ell\in[0,\llmax-1]$: the value to \\
$\sum_{z'\in[1,\zmax]}\wll(z',z)\theta^{(\ell)}(u,z')$ for the edge $(u=\vC_j, v=\vF_i)\in\ECF$ (resp., $(v=\vT_j, u=\vF_i) \in\ETF$), where such an edge $uv$ may be not used in a target chemical graph;

\smallskip\noindent
{\bf constraints: } 

Initializing vectors $\theta^{(0)}(v)$:
\begin{align}
& \thX(i;1;0) = \delaX(i;[\ttC]^\inte), \nonumber \\
& \thX(i;2;0) = \delaX(i;[\ttO]^\inte), \nonumber \\
& \thX(i;3;0) = \delaX(i;[\ttN]^\inte), \nonumber \\
& \thX(i;4;0) = \dgX(i) + \hyddegX(i), \nonumber \\
& \thX(i;5;0) = \sum_{\ta \in \Lambda^\inte} \val(\ta) \delaX(i, [\ta]^\inte) + \eledegX(i), \nonumber \\
& \thX(i;6;0) = \hyddegX(i), \nonumber \\
& \thX(i;7;0) = \eledegX(i), \nonumber  & i \in[1,\tX], \textrm{X} \in \{ \textrm{C,T,F} \} \\
& \thX(i;j;0) = \sum_{\psi\in \FrX_i } \dlfrX(i, [\psi]) \theta_\psi(j - 7; [\psi])  & i \in[1,\tX], j \in[8,\zmax], \textrm{X} \in \{ \textrm{C,T,F} \}
\end{align}

We denote the function $h$ such that $h(\ell) := \zmax$ when $\ell = 0$ and $h(\ell) := \phid$ otherwise in order to simplify the formulations.

Calculating the $(\ell+1)$-th vector from the $\ell$-th vecctor:

\begin{align}
-M_{\ell+1} \leq \sum_{z' \in [1,h(\ell)]} \wll(z',z)\thC(i;z';\ell) + \sum_{k \in I_a^-(i)}\thC_-(k;z;\ell) + \sum_{k\in I_a^+(i)}\thC_+(k;z;\ell)  & \nonumber \\
+ \sum_{k\in I_b^+(i)}\thCTT(k;z;\ell) + \sum_{k\in I_b^-(i)}\thTCT(k;z;\ell) + \thCFF(i;z;\ell) + \bias(z;\ell) = \tauC(i;z;\ell+1) \leq M_{\ell+1}, & \nonumber \\
-M_{\ell + 1} \deltauC(i;z;\ell+1) \le \thC(i;z;\ell+1) - \kappa \tauC(i;z;\ell+1) \le M_{\ell+1} \deltauC(i;z;\ell+1), & \nonumber \\
-M_{\ell + 1} (1 - \deltauC(i;z;\ell+1)) \le \thC(i;z;\ell+1) - \tauC(i;z;\ell+1) \le M_{\ell+1} (1 - \deltauC(i;z;\ell+1)), & \nonumber \\
-M_{\ell+1} \deltauC(i;z;\ell+1) \le \tauC(i;z;\ell+1) \le M_{\ell+1} (1 - \deltauC(i;z;\ell+1)), & \nonumber \\
z \in[1,\phid], \ell \in [0, \llmax-1], i\in[1,\widetilde{\tC}], 
\end{align}

\begin{align}
-M_{\ell+1} \leq \sum_{z' \in [1,h(\ell)]} \wll(z',z)\thC(i;z';\ell) + \sum_{k \in I_a^-(i)}\thC_-(k;z;\ell) + \sum_{k\in I_a^+(i)}\thC_+(k;z;\ell)  & \nonumber \\
+ \sum_{k\in I_b^+(i)}\thCTT(k;z;\ell) + \sum_{k\in I_b^-(i)}\thTCT(k;z;\ell) + \bias(z;\ell) = \tauC(i;z;\ell+1) \leq M_{\ell+1} & \nonumber \\
-M_{\ell + 1} \deltauC(i;z;\ell+1) \le \thC(i;z;\ell+1) - \kappa \tauC(i;z;\ell+1) \le M_{\ell+1} \deltauC(i;z;\ell+1), & \nonumber \\
-M_{\ell + 1} (1 - \deltauC(i;z;\ell+1)) \le \thC(i;z;\ell+1) - \tauC(i;z;\ell+1) \le M_{\ell+1} (1 - \deltauC(i;z;\ell+1)), & \nonumber \\
-M_{\ell+1} \deltauC(i;z;\ell+1) \le \tauC(i;z;\ell+1) \le M_{\ell+1} (1 - \deltauC(i;z;\ell+1)), & \nonumber \\
z \in[1,\phid], \ell \in [0, \llmax-1], i\in[\widetilde{\tC} + 1, \tC], 
\end{align}

\begin{align}
-M_{\ell + 1} \cdot (1 - \vT(i)) \leq 
\tauT(i;z;\ell +1 ) - (\sum_{z' \in [1,h(\ell)]} \wll(z',z)\thT(i;z';\ell) + \thT_-(i;z;\ell) + \thT_+(i;z;\ell)  & \nonumber \\
+\thCTC(i;z;\ell) + \thTCC(i;z;\ell) + \thTFF(i;z;\ell) + \bias(z;\ell))  \leq M_{\ell+1} \cdot (1-\vT(i)), & \nonumber \\
-M_{\ell + 1} \cdot \vT(i) \leq \thT(i;z;\ell + 1) \leq M_{\ell+1} \cdot \vT(i), & \nonumber \\
-M_{\ell + 1} \deltauT(i;z;\ell+1) \le \thT(i;z;\ell+1) - \kappa \tauT(i;z;\ell+1) \le M_{\ell+1} \deltauT(i;z;\ell+1), & \nonumber \\
-M_{\ell + 1} (1 - \deltauT(i;z;\ell+1)) \le \thT(i;z;\ell+1) - \tauT(i;z;\ell+1) \le M_{\ell+1} (1 - \deltauT(i;z;\ell+1)), & \nonumber \\
-M_{\ell+1} \deltauT(i;z;\ell+1) \le \tauT(i;z;\ell+1) \le M_{\ell+1} (1 - \deltauT(i;z;\ell+1)), & \nonumber \\
\thT(0;z;\ell) = \thT(\tT + 1;z;\ell) = 0, & \nonumber \\
z \in[1,\phid], \ell \in [0, \llmax-1], i\in[1, \tT], &
\end{align}

\begin{align}
-M_{\ell + 1} \cdot (1 - \vF(i)) \leq  
\tauF(i;z;\ell +1) - (\sum_{z' \in [1,h(\ell)]} \wll(z',z)\thF(i;z';\ell) + \thF_-(i;z;\ell) + \thF_+(i;z;\ell)  & \nonumber \\
+\thCFC(i;z;\ell)  + \thTFT(i;z;\ell) + \bias(z;\ell))
\leq  M_{\ell+1} \cdot (1-\vF(i)), & \nonumber \\
-M_{\ell + 1} \cdot \vF(i) \leq \thF(i;z;\ell + 1) \leq M_{\ell+1} \cdot \vF(i), & \nonumber \\
-M_{\ell + 1} \deltauF(i;z;\ell+1) \le \thF(i;z;\ell+1) - \kappa \tauF(i;z;\ell+1) \le M_{\ell+1} \deltauF(i;z;\ell+1), & \nonumber \\
-M_{\ell + 1} (1 - \deltauF(i;z;\ell+1)) \le \thF(i;z;\ell+1) - \tauF(i;z;\ell+1) \le M_{\ell+1} (1 - \deltauF(i;z;\ell+1)), & \nonumber \\
-M_{\ell+1} \deltauF(i;z;\ell+1) \le \tauF(i;z;\ell+1) \le M_{\ell+1} (1 - \deltauF(i;z;\ell+1)), & \nonumber \\
\thF(0;z;\ell) = \thF(\tF + 1;z;\ell) = 0, & \nonumber \\
z \in[1,\phid], \ell \in [0, \llmax-1], i\in[1, \tF], &
\end{align}

Preparing associated variables: /* The values for $\thC(i;z;\ell)$, $\thT(i;z;\ell)$ and $\thF(i;z;\ell)$ 
determine the values for $\thC_-(k;z;\ell)$, $\thC_+(k;z;\ell)$, $\thT_-(k;z;\ell)$, $\thT_+(k;z;\ell)$, 
$\thF_-(k;z;\ell)$, $\thF_+(k;z;\ell)$, $\thCTT(k;z;\ell)$,  $\thTCT(k;z;\ell)$, $\thCTC(k;z;\ell)$, 
$\thTCC(k;z;\ell)$, $\thTFF(k;z;\ell)$,$\thCFF(k;z;\ell)$, $\thCFC(k;z;\ell)$ and $\thTFT(k;z;\ell)$ 
by the following. */       

\begin{align}
\sum_{z' \in [1,h(\ell)]} \wll(z',z) \thC(\tail(k);z';\ell) = \thC_-(k;z;\ell), & \nonumber \\
\sum_{z' \in [1,h(\ell)]} \wll(z',z) \thC(\hd(k);z';\ell) = \thC_+(k;z;\ell), & \nonumber \\
-M_\ell \cdot \eC(k) \leq \thC_-(k;z;\ell) \leq M_\ell \cdot \eC(k), & \nonumber \\
-M_\ell \cdot \eC(k) \leq \thC_+(k;z;\ell) \leq M_\ell \cdot \eC(k), & \nonumber \\
k \in[\widetilde{\kC} + 1, \mC], z \in[1,\phid], \ell \in [0,\llmax - 1], &
\end{align}

\begin{align}
\sum_{z' \in [1,h(\ell)]} \wll(z',z) \thT(i - 1;z';\ell) = \thT_-(i;z;\ell), & i \in[2,\tT], & \nonumber \\
\sum_{z' \in [1,h(\ell)]} \wll(z',z) \thT(i + 1;z';\ell) = \thT_+(i;z;\ell), & i \in [1,\tT-1], & \nonumber \\
-M_\ell \cdot \eT(i) \leq \thT_-(i;z;\ell) \leq M_\ell \cdot \eT(i), & & \nonumber \\
-M_\ell \cdot \eT(i+1) \leq \thT_+(i;z;\ell) \leq M_\ell \cdot \eT(i+1), &  i \in[1,\tT],  & \nonumber \\
& z \in[1,\phid], \ell \in [0,\llmax - 1], & 
\end{align}

\begin{align}
\sum_{z' \in [1,h(\ell)]} \wll(z',z) \thF(i - 1;z';\ell) = \thF_-(i;z;\ell), & i \in[2,\tF], & \nonumber \\
\sum_{z' \in [1,h(\ell)]} \wll(z',z) \thF(i + 1;z';\ell) = \thF_+(i;z;\ell), & i \in [1,\tF-1], & \nonumber \\
-M_\ell \cdot \eF(i) \leq \thF_-(i;z;\ell) \leq M_\ell \cdot \eF(i), & & \nonumber \\
-M_\ell \cdot \eF(i+1) \leq \thF_+(i;z;\ell) \leq M_\ell \cdot \eF(i+1), &  i \in[1,\tF],  & \nonumber \\
& z \in[1,\phid], \ell \in [0,\llmax - 1], & 
\end{align}

\begin{align}
\sum_{z' \in [1,h(\ell)]} \wll(z',z) \thT(i ;z';\ell) - M_\ell \cdot (1 - \chiT(i,k) + \eT(i)) \leq \thCTT(k;z;\ell) & \nonumber \\
\leq \sum_{z' \in [1,h(\ell)]} \wll(z',z) \thT(i ;z';\ell) + M_\ell \cdot (1 - \chiT(i,k) + \eT(i)), i \in[1,\tT], \nonumber \\
\sum_{z' \in [1,h(\ell)]} \wll(z',z) \thT(i ;z';\ell) - M_\ell \cdot (1 - \chiT(i,k) + \eT(i + 1)) \leq \thTCT(k;z;\ell) & \nonumber \\
\leq \sum_{z' \in [1,h(\ell)]} \wll(z',z) \thT(i ;z';\ell) + M_\ell \cdot (1 - \chiT(i,k) + \eT(i + 1), i \in[1,\tT], \nonumber \\
-M_\ell \cdot \dclrT(k) \leq \thCTT(k;z;\ell) \leq M_\ell \cdot \dclrT(k), & \nonumber \\
-M_\ell \cdot \dclrT(k) \leq \thTCT(k;z;\ell) \leq M_\ell \cdot \dclrT(k), & \nonumber \\
k \in[1,\kC], z \in[1,\phid], \ell \in[0,\llmax - 1], & 
\end{align}

\begin{align}
\sum_{z' \in [1,h(\ell)]} \wll(z',z) \thC(\tail(k) ;z';\ell) - M_\ell \cdot (1 - \chiT(i,k) + \eT(i)) \leq \thCTC(i;z;\ell) & \nonumber \\
\leq \sum_{z' \in [1,h(\ell)]} \wll(z',z) \thC(\tail(k) ;z';\ell) + M_\ell \cdot (1 - \chiT(i,k) + \eT(i)), k \in[1,\kC], \nonumber \\
\sum_{z' \in [1,h(\ell)]} \wll(z',z) \thC(\hd(k) ;z';\ell) - M_\ell \cdot (1 - \chiT(i,k) + \eT(i + 1)) \leq \thTCC(i;z;\ell) & \nonumber \\
\leq \sum_{z' \in [1,h(\ell)]} \wll(z',z) \thC(\hd(k) ;z';\ell) + M_\ell \cdot (1 - \chiT(i,k) + \eT(i + 1), k \in[1,\kC], \nonumber \\
-M_\ell \cdot (1-\eT(i)) \leq \thCTC(i;z;\ell) \leq M_\ell \cdot (1-\eT(i)), & \nonumber \\
-M_\ell \cdot \vT(i) \leq \thCTC(i;z;\ell) \leq M_\ell \cdot \vT(i), & \nonumber \\
-M_\ell \cdot (1-\eT(i+1)) \leq \thTCC(i;z;\ell) \leq M_\ell \cdot (1-\eT(i+1)), & \nonumber \\
-M_\ell \cdot \vT(i) \leq \thTCC(i;z;\ell) \leq M_\ell \cdot \vT(i), & \nonumber \\
i \in[1,\tT], z \in[1,\phid], \ell \in[0,\llmax - 1], & 
\end{align}

\begin{align}
\sum_{z' \in [1,h(\ell)]} \wll(z',z) \thF(i ;z';\ell) - M_\ell \cdot (1 - \chiF(i,c) + \eF(i)) \leq \thCFF(c;z;\ell) & \nonumber \\
\leq \sum_{z' \in [1,h(\ell)]} \wll(z',z) \thF(i ;z';\ell) + M_\ell \cdot (1 - \chiF(i,c) + \eF(i)), i \in[1,\tF], \nonumber \\
-M_\ell \cdot \chiF(c) \leq \thCFF(c;z;\ell) \leq M_\ell \cdot \chiF(c), & \nonumber \\
c \in[1,\widetilde{\tC}], z \in[1,\phid], \ell \in[0,\llmax - 1], & 
\end{align}

\begin{align}
\sum_{z' \in [1,h(\ell)]} \wll(z',z) \thC(c ;z';\ell) - M_\ell \cdot (1 - \chiF(i,c) + \eF(i)) \leq \thCFC(i;z;\ell) & \nonumber \\
\leq \sum_{z' \in [1,h(\ell)]} \wll(z',z) \thC(c ;z';\ell) + M_\ell \cdot (1 - \chiF(i,c) + \eF(i)), c \in[1,\widetilde{\tC}], \nonumber \\
-M_\ell \cdot \sum_{c\in[1,\widetilde{\tC}]} \chiF(i,c) \leq \thCFC(i;z;\ell) \leq M_\ell \cdot \sum_{c\in[1,\widetilde{\tC}]} \chiF(i,c), & \nonumber \\
-M_\ell \cdot  (1-\eF(i)) \leq \thCFC(i;z;\ell) \leq M_\ell \cdot  (1-\eF(i)) , & \nonumber \\
i \in[1,\tF], z \in[1,\phid], \ell \in[0,\llmax - 1], & 
\end{align}

\begin{align}
\sum_{z' \in [1,h(\ell)]} \wll(z',z) \thF(j ;z';\ell) - M_\ell \cdot (1 - \chiF(j,\widetilde{\tC} + i) + \eF(j)) \leq \thTFF(i;z;\ell) & \nonumber \\
\leq \sum_{z' \in [1,h(\ell)]} \wll(z',z) \thF(j ;z';\ell) + M_\ell \cdot (1 - \chiF(j,\widetilde{\tC} + i) + \eF(j)), j \in [1,\tF], \nonumber \\
-M_\ell \cdot \chiF(\widetilde{\tC} + i) \leq \thTFF(i;z;\ell) \leq M_\ell \cdot \chiF(\widetilde{\tC} + i), & \nonumber \\
i \in[1,\tT], z \in[1,\phid], \ell \in[0,\llmax - 1], & 
\end{align}

\begin{align}
\sum_{z' \in [1,h(\ell)]} \wll(z',z) \thT(j ;z';\ell) - M_\ell \cdot (1 - \chiF(i,\widetilde{\tC} + j) + \eF(i)) \leq \thTFT(i;z;\ell) & \nonumber \\
\leq \sum_{z' \in [1,h(\ell)]} \wll(z',z) \thT(j ;z';\ell) + M_\ell \cdot (1 - \chiF(i,\widetilde{\tC} + j) + \eF(i)), j \in [1,\tT], \nonumber \\
-M_\ell \cdot \sum_{j\in[1,\tT]} \chiF(i, \widetilde{\tC} + j) \leq \thTFT(i;z;\ell) \leq M_\ell \cdot \sum_{j\in[1,\tT]} \chiF(i, \widetilde{\tC} + j), & \nonumber \\
-M_\ell \cdot  (1-\eF(i)) \leq \thTFT(i;z;\ell) \leq M_\ell \cdot  (1-\eF(i)) , & \nonumber \\
i \in[1,\tF], z \in[1,\phid], \ell \in[0,\llmax - 1], & 
\end{align}

\begin{align}
\sum_{i\in[1,\tX], \textrm{X} \in \{ \textrm{C,T,F} \} } \sum_{z \in [1,\phid]} \wftr(z,p) \thX(i;z;\llmax) = \tau_{\mathrm{ftr}}(p), & \nonumber \\
-M_{\llmax} \delta^\tau_{\mathrm{ftr}} (p) \le \thftr(p) - \kappa \tau_{\mathrm{ftr}}(p) \le M_{\llmax} \delta^\tau_{\mathrm{ftr}} (p), & \nonumber \\
-M_{\llmax} (1 - \delta^\tau_{\mathrm{ftr}} (p)) \le \thftr(p) - \tau_{\mathrm{ftr}}(p) \le M_{\llmax} (1 - \delta^\tau_{\mathrm{ftr}} (p)), & \nonumber \\
-M_{\llmax} \delta^\tau_{\mathrm{ftr}} (p) \le \tau_{\mathrm{ftr}}(p) \le M_{\llmax} (1 - \delta^\tau_{\mathrm{ftr}} (p)), & & p \in [1, \pcnv],
\end{align}

\end{document}